\documentclass[journal]{IEEEtran}

\usepackage{cite}
\usepackage{amsmath}
\usepackage[T1]{fontenc}
\usepackage[latin9]{inputenc}
\usepackage{units}
\usepackage{textcomp}
\usepackage{stmaryrd}
\usepackage[pdftex]{graphicx}
\usepackage[caption=false,font=footnotesize]{subfig}
\usepackage[english]{babel}
\usepackage{stfloats}
\usepackage{fixltx2e}
\usepackage{array}
\usepackage{color}
\usepackage{booktabs}
\usepackage{amssymb}
\usepackage{multirow}
\usepackage{makecell}

\makeatother

\begin{document}

\captionsetup[figure]{labelfont={bf},labelformat={default},labelsep=period,name={Fig.}}

\title{Non-Convex Tensor Low-Rank Approximation for Infrared Small Target Detection}

\author{Ting~Liu,
        Jungang~Yang,
        Boyang~Li, Chao~Xiao, Yang~Sun, Yingqian~Wang, Wei~An 

\thanks{This work was supported in part by the National Natural Science Foundation of China under
	Grant 61972435, Grant 61401474, and Grant 61921001. (Corresponding author: Jungang Yang) }

\thanks{T.~Liu, J.~Yang, B.~Li, C.~Xiao, Y.~Sun, Y.~Wang, and W.~An are with the College of Electronic Science, National University of Defense Technology, Changsha, China. (e-mail: liuting@nudt.edu.cn, yangjungang@nudt.edu.cn).}}

\markboth{Journal of \LaTeX\ Class Files,~Vol.~14, No.~8, August~2015}%
{Shell \MakeLowercase{\textit{et al.}}: Bare Demo of IEEEtran.cls for IEEE Journals}
\maketitle

\begin{abstract}
Infrared small target detection is an important fundamental task in the infrared system. Therefore, many infrared small target detection methods have been proposed, in which the low-rank model has been used as a powerful tool. However, most low-rank-based methods assign the same weights for different singular values, which will lead to inaccurate background estimation. Considering that different singular values have different importance and should be treated discriminatively, in this paper, we propose a non-convex tensor low-rank approximation (NTLA) method for infrared small target detection. In our method, NTLA regularization adaptively assigns different weights to different singular values for accurate background estimation. Based on the proposed NTLA, we propose asymmetric spatial-temporal total variation (ASTTV) regularization to achieve more accurate background estimation in complex scenes. Compared with the traditional total variation approach, ASTTV exploits different smoothness intensities for spatial and temporal regularization. We design an efficient algorithm to find the optimal solution of our method. Compared with some state-of-the-art methods, the proposed method achieves an improvement in terms of various evaluation metrics. Extensive experimental results in various complex scenes demonstrate that our method has strong robustness and low false-alarm rate. Code is available at https://github.com/LiuTing20a/ASTTV-NTLA.

\end{abstract}
 
\begin{IEEEkeywords}
Non-convex tensor low-rank, asymmetric spatial-temporal total variation, infrared small target detection.
\end{IEEEkeywords}

\section{Introduction}

\IEEEPARstart{I}{nfrared} small target detection is an important technique in many military and civilian fields, such as buried landmine detection, night navigation, precision guided weapons and missiles \cite{thanh2008infrared, bourennane2010improvement, bai2015infrared}. However, due to the long imaging distance of infrared detection systems, the targets usually lack texture features or fixed shape. In addition, the target has low signal-to-clutter ratio (SCR) because it is always immersed in complex noises and strong clutters scenes. In summary, infrared small target detection is an important and challenging problem.

In the past decades, many scholars have devoted themselves to the research of infrared small target detection and proposed different methods for this task. These methods can be classified into single-frame and sequential infrared small target detection. Single-frame infrared small target detection can be divided into three categories according to different assumptions. The first category supposes that the background changes slowly in the infrared image and there is a high correlation between adjacent pixels. Based on this assumption, many background suppression (BS)-based methods\cite{hadhoud1988two,rivest1996detection,deshpande1999max} were proposed. These methods use filter to suppress the background noise and clutter, and then use the intensity threshold to extract small targets. The BS-based methods obtain satisfactory computation efficiency. However, they achieve relatively poor detection performance with high false alarm rates under discontinuous backgrounds, clutter and noise. Considering that the small target is more visually salient than its surrounding background, human visual system (HVS)-based \cite{kim2009small,chen2013local,han2014robust} methods have been proposed. For example, Chen et al.\cite{chen2013local} introduced a novel local contrast method (LCM). To improve the detection performance, a series of improved LCM (ILCM)\cite{han2014robust} methods have been proposed, such as multiscale relative LCM \cite{han2018infrared}, weighted strengthened local contrast measure (WSLCM) \cite{han2020infrared} and multiscale tri-layer local contrast measure (TLLCM) \cite{han2019local}. However, in some cases of highlighted background or ground background, clutter and target may be similar on the saliency map, which will degrade the performance of target detection.

The second category uses the nonlocal self-correlation between background patches in infrared images to construct low-rank sparse decomposition (LRSD) model, which is a branch of the popular low-rank representation (LRR) in recent years. By constructing local patches, Gao et al. \cite{gao2013infrared} first designed a novel infrared patch-image model (IPI). Then the existing small target detection is transformed into a robust principal component analysis (RPCA) problem \cite{candes2011robust}. Subsequently, many LRSD methods have been proposed. However, due to the limitation of nuclear norm minimization (NNM), it will lead to over-shrinkage problem. To handle the above problems, Guo et al. \cite{guo2017small} designed a novel reweighted IPI (ReWIPI) method, in which weighted nuclear norm minimization (WNNM) was introduced to suppress sparse non-target pixels. However, it can only alleviate the over-shrinkage problem. In \cite{zhang2018infrared}, a non-convex rank approximation model (NRAM) was proposed, in which ${l_{2,1}}$ norm was used to constrain clutter. In recent years, considering the importance of model robustness, a series of multi-subspace structure methods have been proposed, such as low-rank and sparse representation (LRSR) \cite{he2015small}, stable multi-subspace learning methods (SMSL) \cite{wang2017infrared} and self-regularized weighted sparse (SRWS) \cite{zhang2021infrared}. Encouraged by the powerfulness of TV regularization, Wang et al.\cite{wang2017infrared1} introduced the TV regularization into the existing model (TV-PCP).

Compared with matrix domain, multi-directional tensor domain can exploit the inner relationship of the data from more views. Therefore, Dai et. al \cite{dai2017reweighted} developed a new reweighted infrared patch-tensor (RIPT) method. The RIPT model achieves infrared target detection from the perspective of the tensor domain, and can achieve relatively good performance. However, due to the limitation of sum of nuclear norm (SNN), RIPT method achieves less competitive performance in complex scenes. Considering the above problem, Sun et al.\cite{sun2018infrared} and Zhang et al. \cite{zhang2019infrared1} exploited different tensor nuclear norms to improve the detection results of the RIPT model. Guan et al.\cite{guan2020infrared} combined local contrast with nonconvex tensor rank surrogate (NTRS) to infrared small target detection. In\cite{rawat2020infrared }, non-convex triple tensor factorisation was used for target detection. Further, Kong et al. \cite{kong2021infrared} proposed a new nonconvex tensor fibered rank approximation model. The third category is deep learning based method. Recently, deep learning based methods have attracted extensive attention due to its powerful feature learning ability. It is widely used in infrared small target detection \cite{fan2018dim,ryu2018small,li2021dense,dai2021asymmetric}. Although they achieved improved performance, the main challenge of deep learning is that infrared small target lacks shape features and remarkable texture, which makes feature learning difficult.

Although the above single-frame methods have achieved good results, it is very necessary to consider the temporal information because single-frame image lacks sufficient information to detect the small target. Therefore, many sequential small target detection methods based on spatial-temporal information have been proposed. For example, Sun et al.\cite{sun2019infrared2} introduced spatial-temporal TV regularization and weighted tensor nuclear norm into the existing IPT model (STTVWNIPT). Liu et al.\cite{liu2020small} achieved infrared video detecion via a novel spatial-temporal tensor model (IVSTTM). Further, inspired by \cite{wang2017infrared}, Sun et al.\cite{sun2020infrared} introduced the multiple subspace learning strategy into the existing IPT model (MSLSTIPT) to improve its robustness in complex scenes. Considering the importance of spatial-temporal information, this paper mainly focuses on infrared sequential small target detection. Although the existing methods have achieved relatively good results, there are still several problems to be solved.

First, due to the limitation of nuclear norm minimization (NNM), IVSTTM method will lead to an over-shrinkage problem. To solve this problem, the weighted nuclear norm minimization (WNNM) is introduced \cite{sun2019infrared2}. In fact, the introduction of the WNNM method can only slightly alleviate the problem of over-shrinkage. Further, Sun et al. \cite{sun2019infrared, sun2020infrared} introduced weighted Schatten $ p $-norm minimization (WSNM) to obtain more accurate background estimation. In summary, the first problem is how to obtain accurate background estimation in complex background. Second, noise is important interference in real scene, which may lead to false alarm in target detection. In recent years, TV regularization is widely exploited in infrared small target detection. However, the classic TV only considers the spatial information, and its computational complexity is high. Although STTV exploits spatial-temporal information, but it treats spatial and temporal information equally, which will degrade the detection performance.

To solve the above problems, we propose a non-convex tensor low-rank approximation method, which combines NTLA and ASTTV regularization. Firstly, considering the importance of accurate background estimation, the NTLA regularization is introduced. Compared with the $ {l_1} $ norm, NTLA regularization can more closely approximate to the $ {l_0} $ norm through Laplace function (see Fig. 1). Meanwhile, it can adaptively assign weights to each singular value. Therefore, it is helpful to obtain more accurate background estimation. Additionally, considering that target is temporally consistant among successive frames and spatially smooth in local area. We exploit the ASTTV to thoroughly describe background feature, which helps obtain more accurate background estimation in complex scenes. Compared with STTV regularization, ASTTV regularization assigns different smoothness strength to adjacent frames. Therefore, the ASTTV regularization constraint on the background helps to better utilize the spatial-temporal information and detect the target more flexibly. In addition, considering the heavy noise in real scenes, we introduce Frobenius norm into the model. Finally, asymmetric spatial-temporal total variation regularized non-convex tensor low-rank approximation is proposed, named the ASTTV-NTLA model. Fig. 2 shows the framework of the ASTTV-NTLA method. The main contributions of our method is summarized as follows.

(1) We propose a non-convex tensor low-rank approximation method for sequential infrared small target detection. Different from existing low-rank methods, NTLA regularization adaptively assigns different weights to all singular values through Laplace function, which helps to obtain an accurate background estimation.

(2) To capture both spatial and temporal information, we introduce ASSTV regularization into the LRSD model. Furthermore, the ASTTV constraint on the background helps to preserve the details of the image and remove noise. Therefore, it can achieve better performance in complex background scenes.

(3) We integrate NTLA regularization, ASTTV regularization and Frobenius norm for infrared small target detection, and develop an algorithm to solve the ASTTV-NTLA model. The experimental results show that the proposed method can achieve promising detection performance in various scenes.

The rest of sections this paper is organized as follows. Section II describes research work in related fields. We summarize some notations and preliminaries in Section III. In Section IV, the ASTTV-NTLA is proposed, and its optimization procedure is designed. Section V shows the experimental results and performance evaluation along with discussion and analyses. Finally, conclusion is given in Section VI. 

\section{Notations and preliminaries}
In this paper, we use lowercase letters (e.g., x), boldface lowercase letters (e.g., $\boldsymbol{x}$) and boldface capital letters (e.g., $\boldsymbol{X}$) represent scalars, vectors and matrices, respectively. Considering that tensors are multi-index arrays, we use Euler script (e.g., $\mathcal{X}$) to represent them. Readers can refer to \cite{hu2016moving,zhang2014novel,yuan2016tensor,hu2016twist,chen2017iterative} for more details about TNN and t-SVD.

\begin{figure}
	\vspace{-0.2cm}
	\centering\includegraphics[width=5cm]{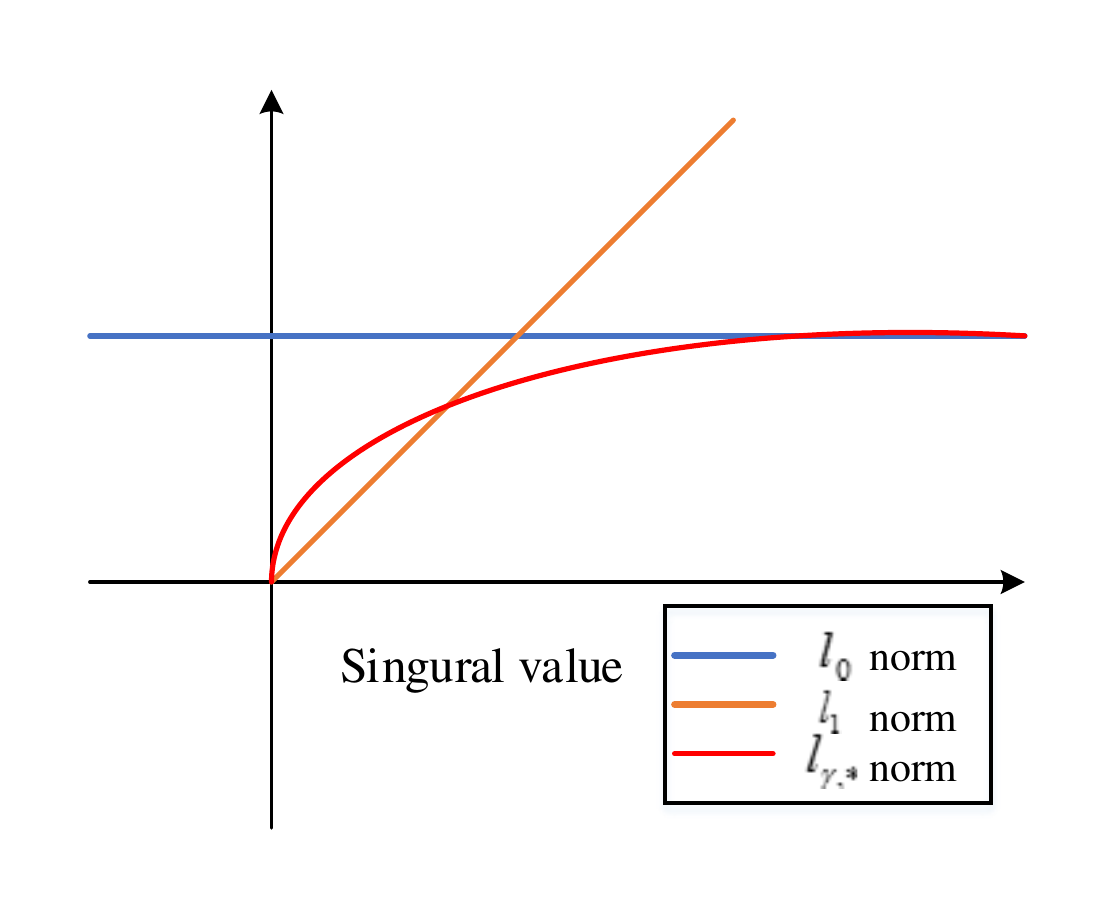}
	\caption{ Comparison of the ${l_0}$ norm, ${l_1}$ norm, and ${l_{\gamma , * }}$ norm for singular value}
	\vspace{-0.2cm}
\end{figure}

\subsection{Adaptive Thresholding Using Laplace Function}
The existing methods generally use TNN to capture the low-rank features of background. Moreover, the operation step of t-SVD is to calculate matrix SVDs of the frontal slices in the Fourier domain. The t-SVD \cite{chen2017iterative} operation on the tensor $ \mathcal{X} \in {\mathbb{R}^{{n_1} \times {n_2} \times {n_3}}} $ is expressed as follows:
\begin{equation}
\mathcal{X} = \mathcal{U} * \Sigma * {\mathcal{V}^{\rm T}},
\end{equation}
where $\mathcal{U}$ and $\mathcal{V}$ represent orthogonal tensors, and $\Sigma$ represents the f-diagonal tensor. The size of $\mathcal{U}$,  $\mathcal{V}$ and $\Sigma$  are $ {n_1} \times {n_1} \times {n_3} $, $ {n_2} \times {n_2} \times {n_3} $ and $ {n_1} \times {n_2} \times {n_3} $,  respectively. $*$ represents the t-product. Small singular values corresponds to other sparse disturbances or noise, which can be removed by setting appropriate thresholds. Then, we can use the remaining larger singular values to reconstruct a low-rank tensor. First, we need to solve the Fourier transform for the third dimension of $ \mathcal{X} $. Then, we assume that $ \bar{X} $ is the result of the Fourier transform along the third dimension. Now, the multirank is a vector whose $ k^{th} $ component gives the rank of the $ k^{th} $ frontal slice, such as $ rank(\mathcal{X}) = (rank({\bar X_1}),rank({\bar X_2}), \ldots ,rank({\bar X_{{n_3}}})) $, where the $ k^{th} $ frontal slice is denoted as $ X^k $ \cite{martin2013order,kilmer2013third}. The sum of the singular values of all the frontal slices (i.e., TNN) is expressed as
\begin{equation}
{\left\|\mathcal{X} \right\|_ *}= \frac{1}{{{n_3}}}\sum\limits_{k = 1}^{{n_3}} {\left\| {{{\bar X}_k}} \right\|}.   
\end{equation}
It is well known that TNN is a surrogate for tensor multirank \cite{lu2016tensor}. The main shortcoming of TNN is that different singular values have the same importance. However, singular values of natural images have clear physical meaning and should be treated differently. To handle the above problems, Xu et al. \cite{xu2019laplace} introduced the Laplace function in TNN, which can automatically assign weights according to the importance of singular values. It is defined as follows:
\begin{equation}
\begin{array}{l}
{\left\|\mathcal{X} \right\|_{\gamma  , * }} = \sum\limits_{k = 1}^{{n_3}} {\sum\limits_i^{\min ({n_1},{n_2})} {\phi \left( {{\sigma _i}({{\bar X}_k})} \right)} } \\
{\kern 1pt} {\kern 1pt} {\kern 1pt} {\kern 1pt} {\kern 1pt} {\kern 1pt} {\kern 1pt} {\kern 1pt} {\kern 1pt} {\kern 1pt} {\kern 1pt} {\kern 1pt} {\kern 1pt} {\kern 1pt} {\kern 1pt} {\kern 1pt} {\kern 1pt} {\kern 1pt} {\kern 1pt} {\kern 1pt} {\kern 1pt} {\kern 1pt} {\kern 1pt} {\kern 1pt} {\kern 1pt} {\kern 1pt} {\kern 1pt} {\kern 1pt}  = \sum\limits_{k = 1}^{{n_3}} {\sum\limits_i^{\min ({n_1},{n_2})} {\phi \left( {1 - {e^{ - {\sigma _i}({{\bar X}_k})/\varepsilon }}} \right)} }, 
\end{array}
\end{equation}
where $ \varepsilon $ is a positive constant and ${\sigma _i}( \cdot ) $ represents the $ i^{th} $ singular value. Laplacian function is represented by $\phi \left( x \right) = 1 - {e^{ - x/\varepsilon }} $. Compared with the ${l_1} $ norm, the Laplace function can better approximate the ${l_0} $ norm (see Fig.1). For the following optimization problems:
 \begin{equation}
\mathop {\arg \min }\limits_\mathcal{Z} {\left\|\mathcal{Z} \right\|_{\gamma , * }} + \frac{\eta }{2}\left\| {\mathcal{Z} - \mathcal{Q}} \right\|_F^2,
\end{equation}
the global optimal solution is $\mathcal{Q} =\mathcal{U} * \Sigma  * {\mathcal{V}^{\rm{H}}} $.
Then, the adaptive singular value threshold processing method is adopted for $\mathcal{Z} $, which is expressed as follows
\begin{equation}
\bar{\mathcal{Z}}  = \mathcal{U} * {\mathcal{D}_{\frac{{\nabla \phi }}{\beta }}} *{\mathcal{V}^{\rm{H}}},
\end{equation}
where $ {\mathcal{D}_{\frac{{\nabla \phi }}{\beta }}} \in {\mathbb{R}^{{n_1} \times {n_2} \times {n_3}}} $ denotes f-diagonal tensor. In Fourier domain, each frontal slice of  $ {\mathcal{D}_{\frac{{\nabla \phi }}{\beta }}}$ is $ {\bar {\mathcal{D}}_{\frac{{\nabla \phi }}{\beta }}} \in {\mathbb{R}^{{n_1} \times {n_2} \times {n_3}}} = {\left( {\bar S\left( {i,j,k} \right) - \frac{{\nabla \phi \left( {\sigma _i^{k,l}} \right)}}{\beta }} \right)_ + } $. The gradient of $ \phi $ in $ {\sigma _i^{k,l}} $ is $ \nabla \phi \left( {\sigma _i^{k,l}} \right) = \frac{1}{\varepsilon }\exp \left( { - \frac{{\sigma _i^{k,l}}}{\varepsilon }} \right)$. And the $ i^{th}$ singular value of the $ k^{th}$  frontal slice of $ \Sigma $ at the $ l^{th}$  previous iteration. $ \mathbf{Algorithm 1} $ briefly describes each iterative solution of the optimization problem in Eq. (4). 

\begin{figure*}[htbp]
	\centering
	\includegraphics[width=14cm]{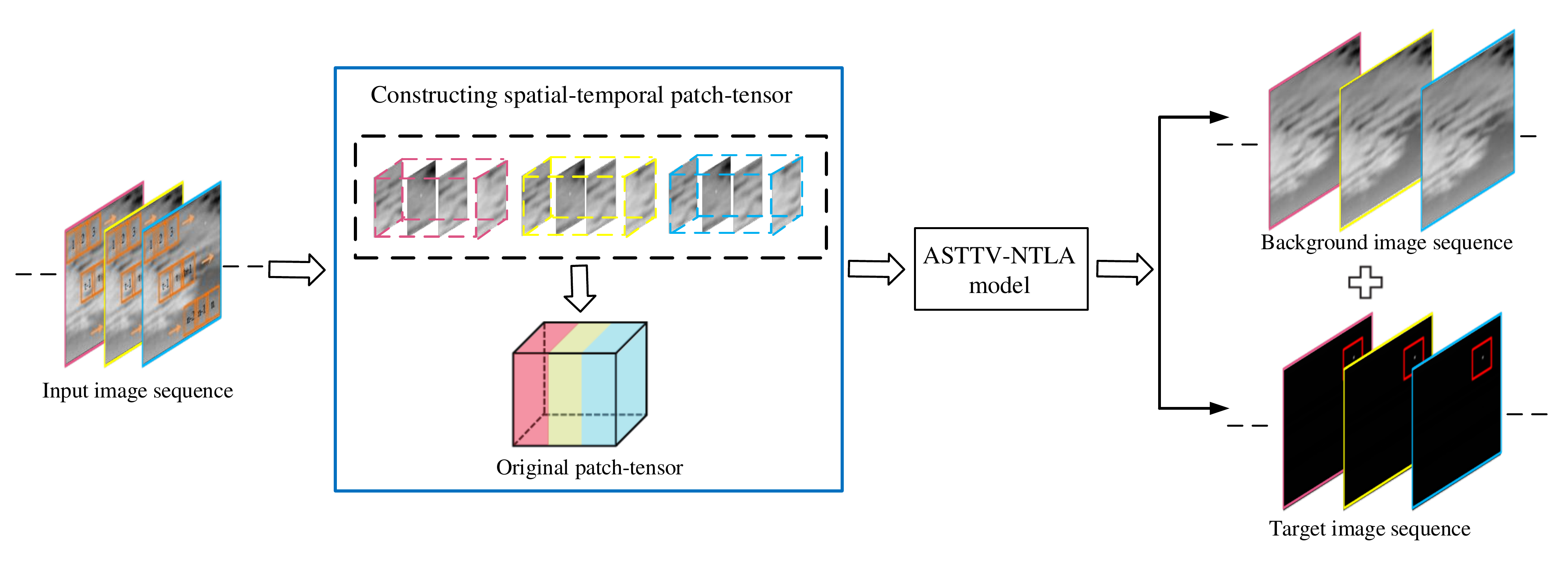}
	\caption{Overall framework of the proposed method.} \label{figDepth}
\end{figure*}

\subsection{Asymmetric spatial-temporal total variation regularization}
In recent years, TV regularization is widely exploited in infrared small target detection, such as \cite{wang2017infrared1}, \cite{sun2019infrared3,fang2020infrared} because of its good performance in preserving the spatial piecewise smoothness, edge structure and spatial sparsity of the images. However, existing methods are based on matrix framework and can only describe the spatial continuity of small targets, but ignore their temporal continuity. In addition, it is time-consuming to calculate SVD and TV regularization. Since target is temporally consistant among successive frames and spatially smooth in local area. Considering the importance of spatial-temporal information, Sun et al.\cite{sun2019infrared} introduced STTV into the existing IPT model. The remarkable performance of STTV-WNIPT demonstrate the effectiveness of simultaneously using spatial-temporal information. To model spatial and temporal continuity, we propose an asymmetric spatial-temporal total variation (ASTTV) regularization approach in the tensor framework\cite{sun2018novel}. The temporal coherence and the spatial-temporal smoothness of small targets are explored. There are two reasons for choosing the ASTTV regularization term. First, imposing ASTTV constraints to the background helps preserve the details of the image and remove noise \cite{tom2020simultaneous}. Second, compared with STTV-WNIPT method, the ASTTV-NTLA method introduces the parameter ${\delta } $ to give different weights to temporal TV and spatial TV. Therefore, ASTTV regularization can detect targets more flexibly. The formulation of ASTTV regularization can be expressed as follows:

\begin{equation}
	{\left\|\mathcal{X} \right\|_{ASTTV}} = {\left\| {{D_h}\mathcal{X}} \right\|_1} + {\left\| {{D_v}\mathcal{X} } \right\|_1} + \delta {\left\| {{D_z}\mathcal{X}} \right\|_1},
\end{equation}
where ${D_h}$, ${D_v}$ and ${D_z}$ represent the horizontal, vertical and temporal difference operators, respectively. ${\delta } $ denotes a positive constant, which is used to control the temporal dimension contribution. The ASTTV in Eq. (6) encourages both spatial and temporal smoothness. The three operators of ASTTV regularization are defined as:
\begin{equation}
	{D_h}\mathcal{X}\left( {i,j,k} \right) = \mathcal{X}\left( {i + 1,j,k} \right) - \mathcal{X}\left( {i,j,k} \right)
\end{equation}
\begin{equation}
	{D_v}\mathcal{X}\left( {i,j,k} \right) = \mathcal{X}\left( {i,j + 1,k} \right) - X\left( {i,j,k} \right)
\end{equation}
\begin{equation}
	{D_z}\mathcal{X}\left( {i,j,k} \right) = \mathcal{X}\left( {i,j ,k+1} \right) - X\left( {i,j,k} \right).
\end{equation}

\begin{tabular}{lcl}
	\toprule
	$\mathbf{Algorithm 1}$: ADMM for solving the Eq. (4)\\
	\midrule
	$\mathbf{Input}$: $\mathcal{X} \in {\mathbb{R}^{{n_1} \times {n_2} \times {n_3}}}$, $\mathcal{Q}$, $\eta $, $ \varepsilon $ \\
	$\mathbf{Output}$ : $ {{\bar {\mathcal{Z}}}^{k + 1}} $, $ {Z^{k + 1}} $\\
	$\mathbf{Step1}$: Compute $ \bar {\mathcal{Q}} = fft\left( {\mathcal{Q},\left[ {} \right],3} \right) $  \\
	$\mathbf{Step2}$ Compute each frontal slice of $ {{\bar {\mathcal{Z}}}^{k + 1}} $ by \\
	$ \mathbf{for} $ $ l = 1, \cdots ,\left[ {{{{n_3} + 1} \mathord{\left/
				{\vphantom {{{n_3} + 1} 2}} \right.
				\kern-\nulldelimiterspace} 2}} \right] $ do \\
	\quad 1. $ \left[ {{{\bar {\mathcal{U}}}^l},{{\bar {\mathcal{S}}}^l},{{\bar {\mathcal{V}}}^l}} \right] = {\rm{SVD}}\left( {{{\bar {\mathcal{Q}}}^l}} \right) $ ; \\
	\quad 2. Compute $ \bar D_{\frac{{\nabla \phi }}{\eta }}^l $ by \\
	\qquad\qquad $ \bar D_{\frac{{\nabla \phi }}{\eta }}^l = {\left( {{{\bar S}^l}\left( {i,j,k} \right) - \frac{{\nabla \phi \left( {\sigma _i^{k,l}} \right)}}{\eta }} \right)_ + } $  \\
	\quad 3. $ {\left( {{{\bar Z}^{k + 1}}} \right)^l} = {{\bar U}^l} * \bar D_{\frac{{\nabla \phi }}{\eta }}^l * {\left( {{{\bar V}^l}} \right)^H} $ ; \\
	$ \mathbf{end\:for} $	  \\
	\quad $ \mathbf{for} $ $ l = \left[ {{{{n_3} + 1} \mathord{\left/
				{\vphantom {{{n_3} + 1} 2}} \right.
				\kern-\nulldelimiterspace} 2}} \right], \cdots ,1 $ do \\
	\qquad	$ {\left( {{{\bar{\mathcal{Z}}}^{k + 1}}} \right)^l} = conj\left( {{{\left( {{{\bar{\mathcal{Z}}}^{k + 1}}} \right)}^{\left( {{n_3} - l + 2} \right)}}} \right); $\\
	\quad $ \mathbf{end\:for} $\\
	$\mathbf{Step3}$: Compute $ \left( {{\mathcal{Z}^{k + 1}}} \right) = ifft\left( {{{\bar {\mathcal{Z}}}^{k + 1}},\left[ {} \right],3} \right) $\\
	\bottomrule
\end{tabular}

\section{Proposed model}
\subsection{Spatial-temporal Infrared Patch Tensor Model}
Given an infrared image, it can be modeled linearly as:
\begin{equation}
f_{D}=f_{B}+f_{T}+f_{N},
\end{equation}	
where $f_{D}$, $f_{T}$, $f_{B}$ and $f_{N}$ denote the input image, target image, background image and noise image, respectively. We use the approach in \cite{sun2019infrared2} to generate 3D tensor. As shown in Fig. 2, each frame image is segmented into patches by sliding window from the top left to the bottom right, and a 3D patch-tensor is formed by stacking all the image patches from consecutive $ L $ frames into a 3D tensor. Similar to Eq. (10), the original tensor is divided into three parts as follows:
\begin{equation}
\mathcal{D}=\mathcal{B}+\mathcal{T}+\mathcal{N},
\end{equation}
where $\mathcal{D}$, $\mathcal{T}$, $\mathcal{B}$, $\mathcal{N} \in {\mathbb{R}^{{n_1} \times {n_2} \times {n_3}}}$ are the patch-tensor forms corresponding to  $f_{D}$, $f_{T}$, $f_{B}$ and $f_{N}$, respectively. The height and width of the sliding window are represented by ${n_1}$ and ${n_3}$, respectively. The number of patches is represented by ${n_2}$. Compared with matrix-based methods, constructing tensor data has two advantages. Firstly, in the tensor domain, we can exploit the inner relationship of data from more views. Secondly, the target detection performance is further improved by combining temporal information. 

\begin{figure*}[htbp]
	\vspace{-0.2cm}
	\centering
	
	\subfloat[Original image]{
		\includegraphics[width=2.8cm]{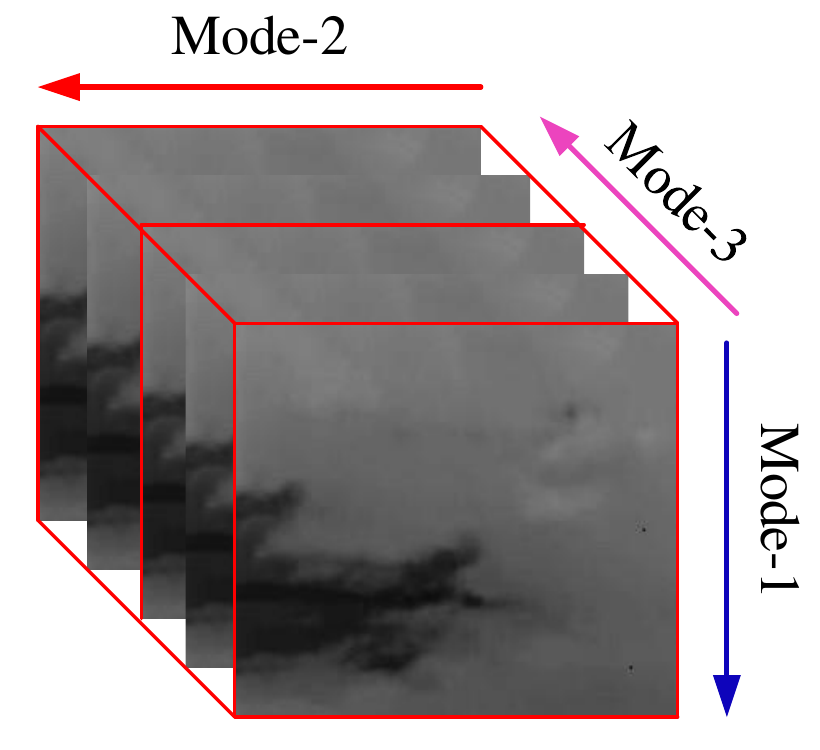}}\subfloat[Mode-1 unfolding]{
		\includegraphics[width=3.7cm]{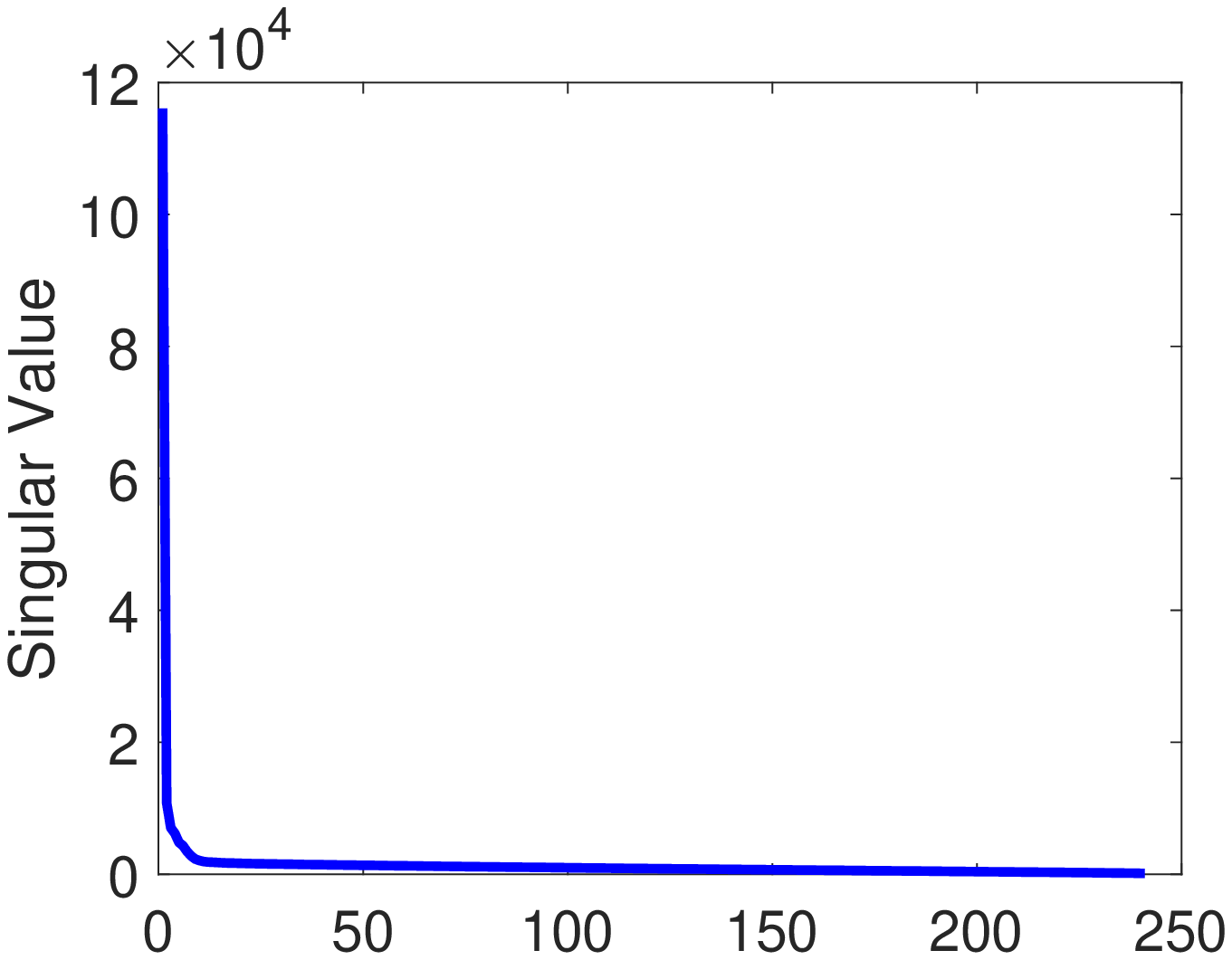}}\subfloat[Mode-2 unfolding]{
		\includegraphics[width=3.7cm]{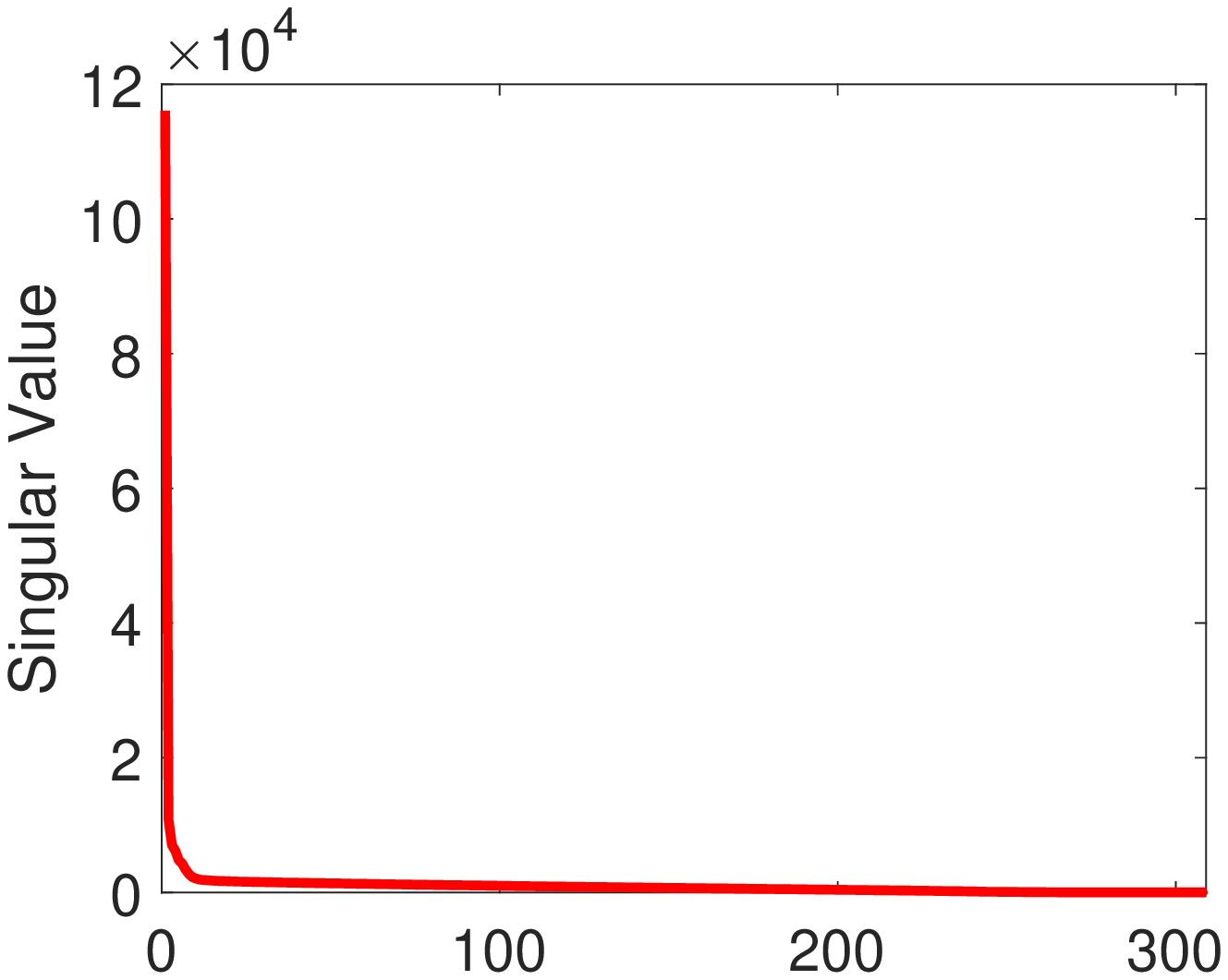}}\subfloat[Mode-3 unfolding ]{
		\includegraphics[width=3.7cm]{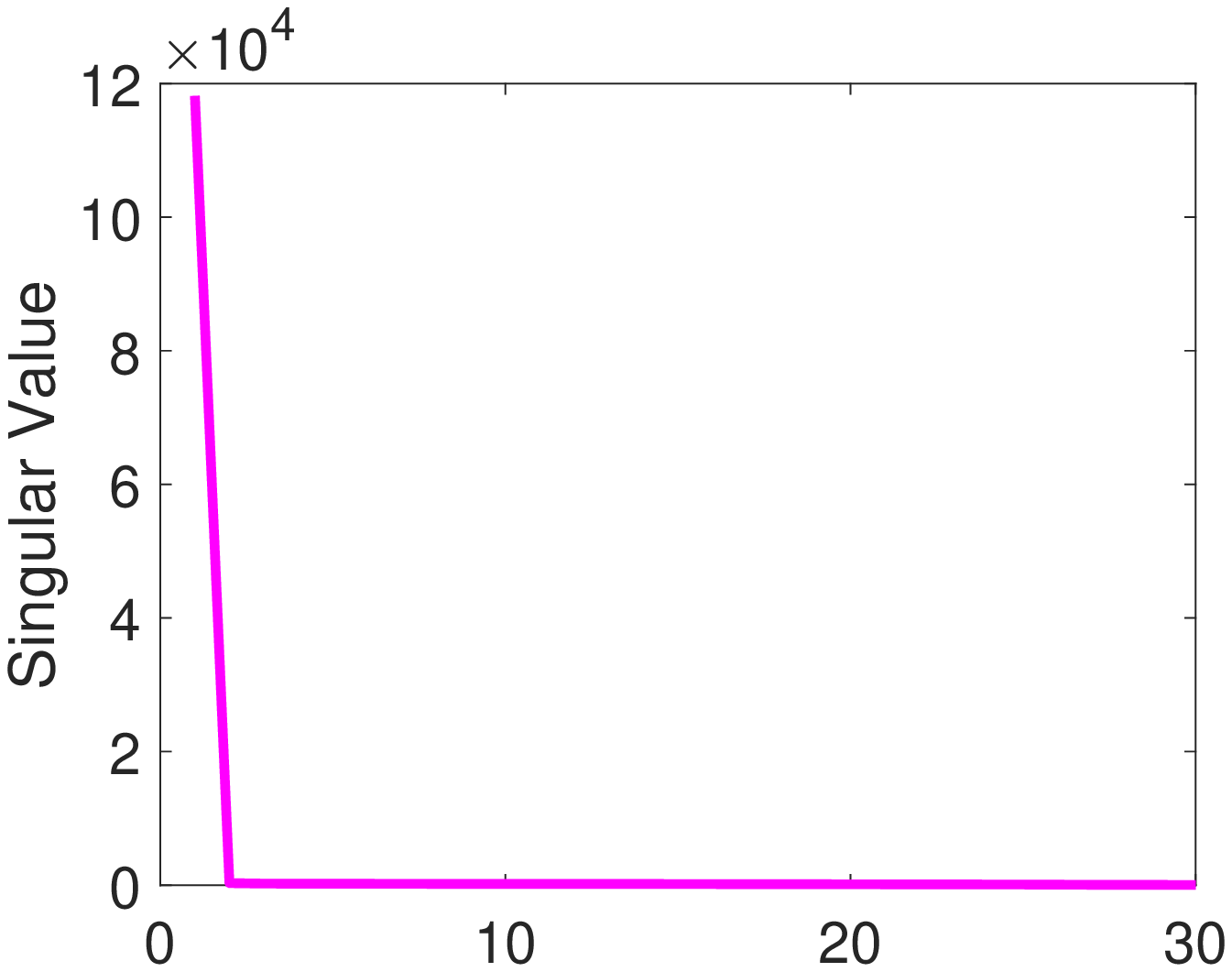}}
	\caption{Singular values of unfolding matrices.}
	\vspace{-0.2cm}
\end{figure*}

It is generally considered that the relative motion between the imaging sensor and the target is slow \cite{rawat2020infrared}. Therefore, the background of different frames is usually assumed to change slowly on the whole sequence images, which indicates high correlations among the adjacent frames in sequence \cite{gao2013infrared}, \cite{gao2018infrared}. It can be seen from Fig. 3 (b)-(d) that the singular values of all the unfolding matrices decrease to zero rapidly when ${n_3}=6$, which indicates background tensor  $ \mathcal{B} $ can be considered as a low-rank tensor.

\subsection{The proposed ASTTV-NTLA model}
By integrating the ASTTV regularization and non-convex tensor rank surrogate into a unified framework, we propose a novel model as follows: 
\begin{equation}
\begin{array}{l}
	\mathcal{B},\mathcal{T},\mathcal{N} = \mathop {\arg \min }\limits_{\mathcal{B},\mathcal{T},\mathcal{N}} {\left\| \mathcal{B} \right\|_{\gamma , * }} + {\lambda _{tv}}{\left\| \mathcal{B} \right\|_{ASTTV}}\\
	{\kern 1pt} {\kern 1pt} {\kern 1pt} {\kern 1pt} {\kern 1pt} {\kern 1pt} {\kern 1pt} {\kern 1pt} {\kern 1pt} {\kern 1pt} {\kern 1pt} {\kern 1pt} {\kern 1pt} {\kern 1pt} {\kern 1pt} {\kern 1pt} {\kern 1pt} {\kern 1pt} {\kern 1pt} {\kern 1pt} {\kern 1pt} {\kern 1pt} {\kern 1pt} {\kern 1pt} {\kern 1pt} {\kern 1pt} {\kern 1pt} {\kern 1pt} {\kern 1pt} {\kern 1pt} {\kern 1pt} {\kern 1pt} {\kern 1pt} {\kern 1pt} {\kern 1pt} {\kern 1pt} {\kern 1pt} {\kern 1pt} {\kern 1pt} {\kern 1pt} {\kern 1pt} {\kern 1pt} {\kern 1pt} {\kern 1pt} {\kern 1pt} {\kern 1pt} {\kern 1pt} {\kern 1pt} {\kern 1pt} {\kern 1pt} {\kern 1pt} {\kern 1pt} {\kern 1pt} {\kern 1pt} {\kern 1pt} {\kern 1pt} {\kern 1pt} {\kern 1pt} {\kern 1pt} {\kern 1pt} {\kern 1pt} {\kern 1pt} {\kern 1pt} {\kern 1pt} {\kern 1pt} {\kern 1pt}  + {\lambda _s}{\left\|\mathcal{T} \right\|_1} + {\lambda _3}\left\| \mathcal{N} \right\|_F^2\\
	{\kern 1pt} {\kern 1pt} {\kern 1pt} {\kern 1pt} {\kern 1pt} {\kern 1pt} {\kern 1pt} {\kern 1pt} {\kern 1pt} {\kern 1pt} {\kern 1pt} {\kern 1pt} {\kern 1pt} {\kern 1pt} {\kern 1pt} {\kern 1pt} {\kern 1pt} {\kern 1pt} {\kern 1pt} {\kern 1pt} {\kern 1pt} {\kern 1pt} {\kern 1pt} {\kern 1pt} {\kern 1pt} {\kern 1pt} {\kern 1pt} {\kern 1pt} {\kern 1pt} {\kern 1pt} {\kern 1pt} {\kern 1pt} {\kern 1pt} {\kern 1pt} {\kern 1pt} {\kern 1pt} {\kern 1pt} {\kern 1pt} s.t.{\kern 1pt} {\kern 1pt} {\kern 1pt} {\kern 1pt} \mathcal{D} = \mathcal{B} + \mathcal{T} + \mathcal{N},
	\end{array}
\end{equation}
where $ {\left\|  \cdot  \right\|_{\gamma , * }}$ is the Laplace function based TNN surrogate,
and ${\lambda _{tv}}$, ${\lambda _s}$, ${\lambda _3}$ denote the positive regularization parameters for ASTTV term, target and noise component, respectively. The first term ${\left\|\mathcal{B} \right\|_{\gamma , * }}$ is used to separate the background from the whole infrared image. The second term ${\left\| \mathcal{B} \right\|_{ASTTV}}$ is used to fully capture spatial-temporal information. It helps preserve local details and remove noise. The third term ${\left\| \mathcal{T} \right\|_1} $ is used to find the sparse target. It can be seen from \cite{sun2019infrared, wang2017hyperspectral} that Frobenius norm has a better effect in noise suppression. Therefore, we further introduce Frobenius norm term $ \left\| \mathcal{N} \right\|_F^2$, which is used to model Gaussian noise and can enhance the performance of the proposed ASTTV-NTLA method in some heavy Gaussian noise situations. Then we can use Eq. (6) to rewrite Eq. (13) as below:
\begin{equation}
\begin{array}{l}
	\mathcal{B},\mathcal{T},\mathcal{N} = \mathop {\arg \min }\limits_{\mathcal{B},\mathcal{T},\mathcal{N}} {\left\| \mathcal{B} \right\|_{\gamma , * }}{\kern 1pt} {\kern 1pt} {\kern 1pt}  + {\lambda _s}{\left\|\mathcal{T} \right\|_1} + {\lambda _3}\left\| \mathcal{N} \right\|_F^2\\
	{\kern 1pt} {\kern 1pt} {\kern 1pt} {\kern 1pt} {\kern 1pt} {\kern 1pt} {\kern 1pt} {\kern 1pt} {\kern 1pt} {\kern 1pt} {\kern 1pt} {\kern 1pt} {\kern 1pt} {\kern 1pt} {\kern 1pt} {\kern 1pt} {\kern 1pt} {\kern 1pt} {\kern 1pt} {\kern 1pt} {\kern 1pt} {\kern 1pt} {\kern 1pt} {\kern 1pt} {\kern 1pt} {\kern 1pt} {\kern 1pt} {\kern 1pt} {\kern 1pt} {\kern 1pt} {\kern 1pt} {\kern 1pt} {\kern 1pt} {\kern 1pt} {\kern 1pt} {\kern 1pt} {\kern 1pt} {\kern 1pt} {\kern 1pt} {\kern 1pt} {\kern 1pt}  + {\lambda _{tv}}\left( {{{\left\| {{D_h}(\mathcal{B})} \right\|}_1} + {{\left\| {{D_v}(\mathcal{B})} \right\|}_1} + \delta {{\left\| {{D_z}(\mathcal{B})} \right\|}_1}} \right){\kern 1pt} {\kern 1pt} {\kern 1pt} {\kern 1pt} {\kern 1pt} {\kern 1pt} {\kern 1pt} {\kern 1pt} {\kern 1pt} {\kern 1pt} {\kern 1pt} {\kern 1pt} {\kern 1pt} {\kern 1pt} {\kern 1pt} {\kern 1pt} {\kern 1pt} {\kern 1pt} \\
	{\kern 1pt} {\kern 1pt} {\kern 1pt} {\kern 1pt} {\kern 1pt} {\kern 1pt} {\kern 1pt} {\kern 1pt} {\kern 1pt} {\kern 1pt} {\kern 1pt} {\kern 1pt} {\kern 1pt} {\kern 1pt} {\kern 1pt} {\kern 1pt} {\kern 1pt} {\kern 1pt} {\kern 1pt} {\kern 1pt} {\kern 1pt} {\kern 1pt} {\kern 1pt} {\kern 1pt} {\kern 1pt} {\kern 1pt} {\kern 1pt} {\kern 1pt} {\kern 1pt} {\kern 1pt} {\kern 1pt} {\kern 1pt} {\kern 1pt} {\kern 1pt} {\kern 1pt} {\kern 1pt} {\kern 1pt} {\kern 1pt} s.t.{\kern 1pt} {\kern 1pt} {\kern 1pt} {\kern 1pt} \mathcal{D} = \mathcal{B} + \mathcal{T} + \mathcal{N},
	\end{array}
\end{equation}

In order to fully capture the spatial and temporal information, the proposed model incorporates NTLA and ASTTV regularization. The main reasons is that the NTLA regularization ensures automatic weight assignment to the singular values, which helps more accurate background evaluation. Furthermore, ASTTV is more flexible as compared with STTV for target detection. 

\subsection{Optimization Procedure}
The optimization Eq. (13) can be solved effectively by using the ADMM\cite{boyd2011distributed} approach. By introducing four auxiliary variables $\mathcal{Z}$, ${V_1} = {D_h}\mathcal{B}$, ${V_2} = {D_v}\mathcal{B}$, ${V_3} = {D_z}\mathcal{B}$, Eq. (13) is rewritten as:
\begin{equation}
\begin{array}{l}
\mathcal{B},\mathcal{T},\mathcal{N} = \mathop {\arg \min }\limits_{\mathcal{B},\mathcal{T},\mathcal{N}} {\left\|\mathcal{Z} \right\|_{\gamma , * }}{\kern 1pt} {\kern 1pt} {\kern 1pt}  + {\lambda _s}{\left\| \mathcal{T} \right\|_1} + {\lambda _3}\left\|\mathcal{N} \right\|_F^2\\
{\kern 1pt} {\kern 1pt} {\kern 1pt} {\kern 1pt} {\kern 1pt} {\kern 1pt} {\kern 1pt} {\kern 1pt} {\kern 1pt} {\kern 1pt} {\kern 1pt} {\kern 1pt} {\kern 1pt} {\kern 1pt} {\kern 1pt} {\kern 1pt} {\kern 1pt} {\kern 1pt} {\kern 1pt} {\kern 1pt} {\kern 1pt} {\kern 1pt} {\kern 1pt} {\kern 1pt} {\kern 1pt} {\kern 1pt} {\kern 1pt} {\kern 1pt} {\kern 1pt} {\kern 1pt} {\kern 1pt} {\kern 1pt} {\kern 1pt} {\kern 1pt} {\kern 1pt} {\kern 1pt} {\kern 1pt} {\kern 1pt} {\kern 1pt} {\kern 1pt} {\kern 1pt}  + {\lambda _{tv}}\left( {{{\left\| {{V_1}} \right\|}_1} + {{\left\| {{V_2}} \right\|}_1} + \delta {{\left\| {{V_3}} \right\|}_1}} \right){\kern 1pt} {\kern 1pt} {\kern 1pt} {\kern 1pt} {\kern 1pt} {\kern 1pt} {\kern 1pt} {\kern 1pt} {\kern 1pt} {\kern 1pt} {\kern 1pt} {\kern 1pt} {\kern 1pt} {\kern 1pt} {\kern 1pt} {\kern 1pt} {\kern 1pt} {\kern 1pt} \\
{\kern 1pt} {\kern 1pt} {\kern 1pt} {\kern 1pt} {\kern 1pt} {\kern 1pt} {\kern 1pt} {\kern 1pt} {\kern 1pt} {\kern 1pt} {\kern 1pt} {\kern 1pt} {\kern 1pt} {\kern 1pt} {\kern 1pt} {\kern 1pt} {\kern 1pt} {\kern 1pt} {\kern 1pt} {\kern 1pt} {\kern 1pt} {\kern 1pt} {\kern 1pt} {\kern 1pt} {\kern 1pt} {\kern 1pt} {\kern 1pt} {\kern 1pt} {\kern 1pt} {\kern 1pt} {\kern 1pt} {\kern 1pt} {\kern 1pt} {\kern 1pt} {\kern 1pt} {\kern 1pt} {\kern 1pt} {\kern 1pt} s.t.{\kern 1pt} {\kern 1pt} {\kern 1pt} {\kern 1pt} \mathcal{D} = \mathcal{B} + \mathcal{T} + \mathcal{N},\mathcal{Z} = \mathcal{B},{V_1} = {D_h}\mathcal{B},\\
{\kern 1pt} {\kern 1pt} {\kern 1pt} {\kern 1pt} {\kern 1pt} {\kern 1pt} {\kern 1pt} {\kern 1pt} {\kern 1pt} {\kern 1pt} {\kern 1pt} {\kern 1pt} {\kern 1pt} {\kern 1pt} {\kern 1pt} {\kern 1pt} {\kern 1pt} {\kern 1pt} {\kern 1pt} {\kern 1pt} {\kern 1pt} {\kern 1pt} {\kern 1pt} {\kern 1pt} {\kern 1pt} {\kern 1pt} {\kern 1pt} {\kern 1pt} {\kern 1pt} {\kern 1pt} {\kern 1pt} {\kern 1pt} {\kern 1pt} {\kern 1pt} {\kern 1pt} {\kern 1pt} {\kern 1pt} {\kern 1pt} {\kern 1pt} {\kern 1pt} {\kern 1pt} {\kern 1pt} {\kern 1pt} {\kern 1pt} {\kern 1pt} {\kern 1pt} {\kern 1pt} {\kern 1pt} {\kern 1pt} {\kern 1pt} {\kern 1pt} {\kern 1pt} {\kern 1pt} {V_2} = {D_v}\mathcal{B},{V_3} = {D_z}\mathcal{B}.
\end{array}
\end{equation}
We use the inexact augmented Lagrangian multiplier (IALM) \cite{lin2010augmented} approach to solve Eq. (14), which is expressed as follows:
\begin{equation}
\begin{array}{l}
{L_A}\left( {\mathcal{B},\mathcal{T},\mathcal{N},\mathcal{Z},V} \right)\\
{\kern 1pt} {\kern 1pt} {\kern 1pt} {\kern 1pt}  = {\left\|\mathcal{Z} \right\|_{\gamma , * }}{\kern 1pt} {\kern 1pt} {\kern 1pt}  + {\lambda _s}{\left\| \mathcal{T} \right\|_1} + {\lambda _{tv}}\left( {{{\left\| {{V_1}} \right\|}_1} + {{\left\| {{V_2}} \right\|}_1} + \delta {{\left\| {{V_3}} \right\|}_1}} \right){\kern 1pt} {\kern 1pt} {\kern 1pt} \\
{\kern 1pt} {\kern 1pt} {\kern 1pt} {\kern 1pt}  + \left\langle {{y_1},\mathcal{D} - \mathcal{B} - \mathcal{T} -\mathcal{N}} \right\rangle {\kern 1pt} {\kern 1pt} {\kern 1pt} {\kern 1pt}  + \left\langle {{y_2},\mathcal{Z} - \mathcal{B}} \right\rangle {\kern 1pt} {\kern 1pt} {\kern 1pt} {\kern 1pt} \\
{\kern 1pt} {\kern 1pt} {\kern 1pt} {\kern 1pt}  + \left\langle {{y_3},{V_1} - {D_h}\mathcal{B}} \right\rangle  + \left\langle {{y_4},{V_2} - {D_v}\mathcal{B}} \right\rangle {\kern 1pt} {\kern 1pt} {\kern 1pt}  + \left\langle {{y_5},{V_3} - {D_z}\mathcal{B}} \right\rangle {\kern 1pt} {\kern 1pt} \\
{\kern 1pt} {\kern 1pt} {\kern 1pt} {\kern 1pt}  + \frac{\mu }{2}{\kern 1pt} {\kern 1pt} {\kern 1pt} (\left\| {\mathcal{D} - \mathcal{B} - \mathcal{T} -\mathcal{N}} \right\|_F^2 + \left\| {\mathcal{Z} - \mathcal{B}} \right\|_F^2 + \left\| {{V_1} - {D_h}B} \right\|_F^2{\kern 1pt} {\kern 1pt} {\kern 1pt} \\
{\kern 1pt} {\kern 1pt} {\kern 1pt} {\kern 1pt}  + \left\| {{V_2} - {D_v}\mathcal{B}} \right\|_F^2 + \left\| {{V_3} - {D_z}\mathcal{B}} \right\|_F^2){\kern 1pt} {\kern 1pt}  + {\lambda _3}\left\| \mathcal{N} \right\|_F^2,
\end{array}
\end{equation}
where $\mu$ represent a positive penalty scalar and ${{y_1}}$, ${{y_2}}$, ${{y_3}}$, ${{y_4}}$, ${{y_5}}$ represent the Lagrangian multiplier. Eq. (15) is decomposed into five optimization subproblems by ADMM algorithm, including $\mathcal{Z}$, $\mathcal{B}$, $\mathcal{T}$, ${V_1},{V_2},{V_3}$, $\mathcal{N}$. Since it is difficult to optimize all of these variables in Eq. (15) at the same time, we can alternately update the variables as:

  1) Updating $\mathcal{Z}$ with other variables being fixed:
  \begin{equation}
  {\mathcal{Z}^{k + 1}} = \mathop {\arg \min }\limits_\mathcal{Z} {\left\|\mathcal{Z} \right\|_{\gamma , * }} + \frac{{{\mu ^k}}}{2}\left\| {\mathcal{Z} - {\mathcal{B}^k} + \frac{{y_2^k}}{{{\mu ^k}}}} \right\|_F^2.
  \end{equation}
Let ${\mathcal{B}^k} - \frac{{y_2^k}}{{{\mu ^k}}} = \mathcal{U} * \mathcal{S}*{\mathcal{V}^{\rm{H}}}$, then the optimal solution can be obtained by Eq. (5). Therefore, the solution of Eq. (16) is
\begin{equation}
{\mathcal{Z}^{k + 1}} =\mathcal{U} * {\mathcal{D}_{\frac{{\nabla \phi }}{\beta }}} * {\mathcal{V}^{\rm{H}}},
\end{equation}
where  $ {{\bar{\mathcal{D}}}_{\frac{{\nabla \phi }}{\beta }}} = {\left( {\bar S\left( {i,j,k} \right) - \frac{{\nabla \phi \left( {\sigma _i^{k,l}} \right)}}{\beta }} \right)_ + }. $ 
The detailed solving process of Eq. (16) is shown in $\mathbf{Algorithm 1} $.

 2) Updating $\mathcal{B}$ with other variables being fixed:
 \begin{equation}
 \begin{array}{l}
 {\mathcal{B}^{k + 1}} = \frac{{{\mu ^k}}}{2}(\left\| {\mathcal{D} -\mathcal{B} - {\mathcal{T}^k} - {\mathcal{N}^k} + \frac{{y_1^k}}{{{\mu ^k}}}} \right\|_F^2\\
 {\kern 1pt} {\kern 1pt} {\kern 1pt} {\kern 1pt} {\kern 1pt} {\kern 1pt} {\kern 1pt} {\kern 1pt} {\kern 1pt} {\kern 1pt} {\kern 1pt} {\kern 1pt} {\kern 1pt} {\kern 1pt} {\kern 1pt} {\kern 1pt} {\kern 1pt} {\kern 1pt} {\kern 1pt} {\kern 1pt} {\kern 1pt} {\kern 1pt} {\kern 1pt} {\kern 1pt} {\kern 1pt} {\kern 1pt} {\kern 1pt}  + \left\| {{\mathcal{Z}^{k + 1}} - \mathcal{B} + \frac{{y_2^k}}{{{\mu ^k}}}} \right\|_F^2 + \left\| {V_1^k - {D_h}\mathcal{B} + \frac{{y_3^k}}{{{\mu ^k}}}} \right\|_F^2\\
 {\kern 1pt} {\kern 1pt} {\kern 1pt} {\kern 1pt} {\kern 1pt} {\kern 1pt} {\kern 1pt} {\kern 1pt} {\kern 1pt} {\kern 1pt} {\kern 1pt} {\kern 1pt} {\kern 1pt} {\kern 1pt} {\kern 1pt} {\kern 1pt} {\kern 1pt} {\kern 1pt} {\kern 1pt} {\kern 1pt} {\kern 1pt} {\kern 1pt} {\kern 1pt} {\kern 1pt} {\kern 1pt} {\kern 1pt} {\kern 1pt}  + \left\| {V_2^k - {D_v}\mathcal{B} + \frac{{y_4^k}}{{{\mu ^k}}}} \right\|_F^2 + \left\| {V_3^k - {D_z}\mathcal{B} + \frac{{y_5^k}}{{{\mu ^k}}}} \right\|_F^2).
 \end{array}
 \end{equation}
 The solution to Eq. (18) is equivalent to the following liner system of equations:
 \begin{equation}
\left( {2{\rm I}{\rm{ + }}\Delta } \right){\mathcal{B}^{k + 1}} = {L^k} + {\theta _1} + {\theta _2} + {\theta _3},
 \end{equation}
where $ \Delta  = D_h^{\rm{T}}{D_h} + D_v^{\rm{T}}{D_v} + D_z^{\rm{T}}{D_z}$, $ {L^k} =\mathcal{D} - {\mathcal{T}^k} - {\mathcal{N}^k} + \frac{{y_1^k}}{{{\mu ^k}}} + \mathcal{Z} + \frac{{y_2^k}}{{{\mu ^k}}}$, $ {\theta _1}={\kern 1pt} {\kern 1pt} {\kern 1pt} {\kern 1pt} D_h^{\rm{T}}{\kern 1pt} \left( {V_1^k + \frac{{y_3^k}}{{{\mu ^k}}}} \right){\kern 1pt} {\kern 1pt} {\kern 1pt} {\kern 1pt},$ ${\kern 1pt} {\theta _2} = {\kern 1pt} {\kern 1pt} {\kern 1pt} {\kern 1pt} D_h^{\rm{T}}{\kern 1pt} \left( {V_2^k + \frac{{y_4^k}}{{{\mu ^k}}}} \right){\kern 1pt} {\kern 1pt} {\kern 1pt} {\kern 1pt} {\kern 1pt} {\kern 1pt} {\kern 1pt} {\kern 1pt} {\kern 1pt} $, $ {\theta _3} = {\kern 1pt} {\kern 1pt} {\kern 1pt} {\kern 1pt} D_h^{\rm{T}}{\kern 1pt} \left( {V_3^k + \frac{{y_5^k}}{{{\mu ^k}}}} \right){\kern 1pt} {\kern 1pt} {\kern 1pt} {\kern 1pt} {\kern 1pt} {\kern 1pt} {\kern 1pt} {\kern 1pt} {\kern 1pt} $, and T is the matrix transpose. By considering ${{D_h}\mathcal{B}}$, $ {{D_v}\mathcal{B}}$, and ${{D_z}\mathcal{B}}$ as convolutions along two spatial directions and one temporal direction, the closed form solution of Eq. (19) is obtained by nFFT as follows:
\begin{equation}
{\mathcal{B}^{k + 1}} = {\mathcal{F}^{ - 1}}\left( {\frac{{\mathcal{F}\left( {{L^k} + {\theta _1} + {\theta _2} + {\theta _3}} \right)}}{{2 + \sum\nolimits_{i \in \left\{ {h,v,z} \right\}} {\mathcal{F}{{\left( {{D_i}} \right)}^{\rm{H}}}\mathcal{F}\left( {{D_i}} \right)} }}} \right),
\end{equation}
where H, $ \mathcal{F}$ and $ {\mathcal{F}^{ - 1}} $ denote the complex conjugate, the fast nFFT operator and the inverse nFFT operator, respectively.

3) Updating $\mathcal{T}$ with other variables being fixed:
\begin{equation}
{\mathcal{T}^{k + 1}} = \mathop {\arg \min }\limits_T {\lambda _s}{\left\| \mathcal{T} \right\|_1} + \frac{{{\mu ^k}}}{2}\left\| {\mathcal{D} -\mathcal B^{k + 1} - \mathcal{T} - \mathcal N^{k} + \frac{{y_1^k}}{{{\mu ^k}}}} \right\|_F^2.
\end{equation}
The similar element-wise shrinkage operation approach in \cite{beck2009fast} is used to solve Eq. (21):
\begin{equation}
\begin{array}{l}
{\mathcal{T}^{k + 1}} = T{h_{{\lambda _1}\left( {{\mu ^k}} \right)}}^{ - 1}\left( {\mathcal{D} - \mathcal B^{k + 1}- \mathcal N^{k} + \frac{{y_1^k}}{{{\mu ^k}}}} \right),
\end{array}
\end{equation}
where $Th\left(  \cdot  \right) $ denotes the element-wise shrinkage operator.

4) Updating ${{V_1}},{{V_2}},{{V_3}}$ with other variables being fixed:	

\begin{equation}
	\left\{ \begin{array}{l}
	V_1^{k + 1} = \mathop {\arg \min }\limits_{{V_1}} {\lambda _{tv}}{\left\| {{V_1}} \right\|_1} + \frac{{{\mu ^k}}}{2}\left\| {V_1^k - {D_h}{B^{k + 1}} + \frac{{y_3^k}}{{{\mu ^k}}}} \right\|_F^2\\
	V_2^{k + 1} = \mathop {\arg \min }\limits_{{V_2}} {\lambda _{tv}}{\left\| {{V_2}} \right\|_1} + \frac{{{\mu ^k}}}{2}\left\| {V_2^k - {D_v}{B^{k + 1}} + \frac{{y_4^k}}{{{\mu ^k}}}} \right\|_F^2\\
	V_3^{k + 1} = \mathop {\arg \min }\limits_{{V_3}} \delta {\lambda _{tv}}{\left\| {{V_3}} \right\|_1} + \frac{{{\mu ^k}}}{2}\left\| {V_3^k - {D_z}{B^{k + 1}} + \frac{{y_5^k}}{{{\mu ^k}}}} \right\|_F^2.
	\end{array} \right.
\end{equation}

The above problem can also be solved by element-wise shrinkage operator:
\begin{equation}
	\left\{ \begin{array}{l}
	V_1^{k + 1} = T{h_{{\lambda _{tv\left( {{\mu ^k}} \right)}}^{ - 1}}}\left( {{D_h}{B^{k + 1}} - \frac{{y_3^k}}{{{\mu ^k}}}} \right)\\
	V_2^{k + 1} = T{h_{{\lambda _{tv\left( {{\mu ^k}} \right)}}^{ - 1}}}\left( {{D_v}{B^{k + 1}} - \frac{{y_4^k}}{{{\mu ^k}}}} \right)\\
	V_3^{k + 1} = T{h_{\delta {\lambda _{tv\left( {{\mu ^k}} \right)}}^{ - 1}}}\left( {{D_z}{B^{k + 1}} - \frac{{y_5^k}}{{{\mu ^k}}}} \right).
	\end{array} \right.
\end{equation}

5) Updating ${\mathcal{N}^{k + 1}}$ with other variables being fixed:
\begin{equation}
\begin{array}{l}
{\mathcal{N}^{k + 1}} = \mathop {\arg \min }\limits_\mathcal{N} {\lambda _3}\left\| \mathcal{N} \right\|_F^2\\
{\kern 1pt} {\kern 1pt} {\kern 1pt} {\kern 1pt} {\kern 1pt} {\kern 1pt} {\kern 1pt} {\kern 1pt} {\kern 1pt} {\kern 1pt} {\kern 1pt} {\kern 1pt} {\kern 1pt} {\kern 1pt} {\kern 1pt} {\kern 1pt} {\kern 1pt} {\kern 1pt} {\kern 1pt} {\kern 1pt} {\kern 1pt} {\kern 1pt} {\kern 1pt} {\kern 1pt} {\kern 1pt} {\kern 1pt} {\kern 1pt} {\kern 1pt} {\kern 1pt} {\kern 1pt}  + \frac{{{\mu ^k}}}{2}\left\| {\mathcal{D} - {\mathcal{B}^{k + 1}} - {\mathcal{T}^{k + 1}} - \mathcal{N} + \frac{{y_1^k}}{{{\mu ^k}}}} \right\|_F^2.
\end{array}
\end{equation}	
The solution of Eq.(25) is expressed as follows:
\begin{equation}
{\mathcal{N}^{k + 1}} = \frac{{{\mu ^k}\left( {\mathcal{D} - {\mathcal{B}^{k + 1}} - {\mathcal{T}^{k + 1}}} \right) + y_1^k}}{{{\mu ^k} + 2{\lambda _3}}}.
\end{equation}

6) Updating multipliers ${y_1},{y_2},{y_3},{y_4},{y_5}$ with other variables being fixed:
\begin{equation}
\left\{ \begin{array}{l}
y_1^{k + 1} = y_1^k + {\mu ^k}\left( {\mathcal{D} - {\mathcal{B}^{k + 1}} - {\mathcal{T}^{k + 1}} - {\mathcal{N}^{k + 1}}} \right)\\
y_2^{k + 1} = y_2^k + {\mu ^k}\left( {{\mathcal{Z}^{k + 1}} - {\mathcal{B}^{k + 1}}} \right)\\
y_3^{k + 1} = y_3^k + {\mu ^k}\left( {V_1^{k + 1} - {D_h}{\mathcal{B}^{k + 1}}} \right)\\	y_4^{k + 1} = y_4^k + {\mu ^k}\left( {V_2^{k + 1} - {D_v}{\mathcal{B}^{k + 1}}} \right)\\
y_5^{k + 1} = y_5^k + {\mu ^k}\left( {V_3^{k + 1} - {D_z}{\mathcal{B}^{k + 1}}} \right).
\end{array} \right.
\end{equation}

7) Updating $ {\mu ^{k + 1}}$ by $ {\mu ^{k + 1}} = \min \left( {\rho {\mu ^k},{\mu _{\max }}} \right) $.

Finally, the proposed ASTTV-NTLA method is summarized in $\mathbf{Algorithm 2}$. 

\begin{tabular}{lcl}
	\toprule
	$\mathbf{Algorithm 2}$:  ASTTV-NTLA Algorithm   \\
	\midrule
	$\mathbf{Input}$: infrared image sequence $ {d_1},\cdots ,{d_P} \in {\mathbb{R}^{{n_1} \times {n_2}}} $,\\number of frames L, parameters $ {\lambda _s,\lambda _{tv},\lambda _3,\mu > 0}$  \\
	$\mathbf{Initialize}$: Transform the image sequence into the \\original tensor $ \mathcal{D},{\mathcal{B}^0} = {\mathcal{T}^0} = {\mathcal{N}^0}= V_i^0 = 0,i = 1,2,3$,\\ $y_i^0 = 0,i = 1, \cdots ,5$,	${\mu _0} = 1e - 2$, ${\mu _{\max }} = 1e7$, $k = 0$,\\
	$\rho  = 1.5$, $\zeta  = 1e - 6$. \\
	$\mathbf{While:}$ not converged do \\
	$\mathbf{1:}$ Update ${\mathcal{Z}^{k + 1}}$ by $\mathbf{Algorithm 1}$ \\
	$\mathbf{2:}$ Update ${\mathcal{B}^{k + 1}}$ by Eq.(20) \\
	$\mathbf{3:}$ Update ${\mathcal{T}^{k + 1}}$ by Eq.(22) \\
	$\mathbf{4:}$ Update $ V_1^{k + 1},V_2^{k + 1},V_3^{k + 1} $ by Eq.(24) \\
	$\mathbf{5:}$ Update ${\mathcal{N}^{k + 1}}$ by Eq.(26) \\
	$\mathbf{6:}$ Update multipliers $ y_i^{k + 1},i = 1, \cdots ,5 $ by Eq.(27)\\
	$\mathbf{7:}$ Update $ {\mu ^{k + 1}}$ by \\
	\qquad $ {\mu ^{k + 1}} = \min \left( {\rho {\mu ^k},{\mu _{\max }}} \right) $\\
	$\mathbf{8:}$ Check the convergence conditions \\
	\qquad$\frac{{\left\| {\mathcal{D} - {\mathcal{B}^{k + 1}} - {\mathcal{T}^{k + 1}} - {\mathcal{N}^{k + 1}}} \right\|_F^2}}{{\left\|\mathcal{D} \right\|_F^2}} \le \zeta $\\
	$\mathbf{9:}$ Update $ k = k + 1 $  \\
	$\mathbf{end\:While} $ \\
	$\mathbf{Output:} $ $ {\mathcal{B}^{k + 1}},{\mathcal{T}^{k + 1}},{\mathcal{N}^{k + 1}} $ \\
	\bottomrule
\end{tabular}

\begin{figure}[htb]
	\vspace{-0.2cm}
	\centering
	
	\includegraphics[width=2.6cm,height=2cm]{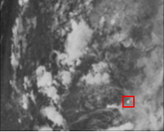}
	\includegraphics[width=2.6cm,height=2cm]{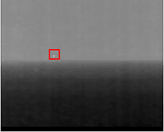}
	\includegraphics[width=2.6cm,height=2cm]{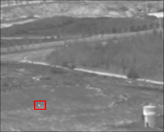}
	\quad
	\subfloat[Scene 1]{
		\includegraphics[width=2.6cm,height=2cm]{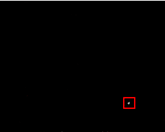}}\subfloat[Scene 2]{
		\includegraphics[width=2.6cm,height=2cm]{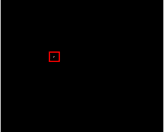}}\subfloat[Scene 3]{
		\includegraphics[width=2.6cm,height=2cm]{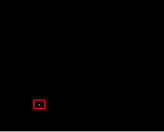}}
	\caption{Single target scenes and results. Row 1: the original images. Row 2:
		the corresponding target detection results.}
	\vspace{-0.2cm}
\end{figure}

\begin{figure}[htb]
	\vspace{-0.2cm}
	\centering
	
	\includegraphics[width=2.6cm,height=2cm]{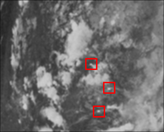}
	\includegraphics[width=2.6cm,height=2cm]{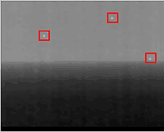}
	\includegraphics[width=2.6cm,height=2cm]{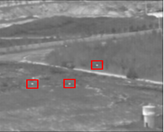}
	\quad
	\subfloat[Scene 1]{
		\includegraphics[width=2.6cm,height=2cm]{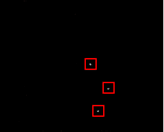}}\subfloat[Scene 2]{
		\includegraphics[width=2.6cm,height=2cm]{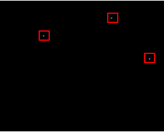}}\subfloat[Scene 3]{
		\includegraphics[width=2.6cm,height=2cm]{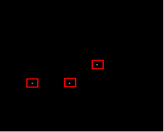}}
	\caption{ Multiple target scenes and results. Row 1: the original images. Row 2:
		the corresponding target detection results.}
	\vspace{-0.2cm}
\end{figure}

\subsection{Target Detection procedure}
The flow chart of the proposed ASTTV-NTLA method is shown in Fig. 2. Next, we will describe each step in detail.

1) Patch-tensor construction. By stacking $ {{n_3}} $ adjacent frames in chronological order, the original infrared image sequence $ {d_1},{d_2}, \cdots ,{d_P} \in {\mathbb{R}^{{n_1} \times {n_2}}} $ transform into several patch-tensors $ \mathcal{D} \in {\mathbb{R}^{{n_1} \times {n_2} \times {n_3}}} $ 

2) Background and target separation. The original patch-tensor $\mathcal{D} $ is decomposed into three parts by $\mathbf{ Algorithm\:2} $, which are target patch-tenor $\mathcal{T} $, noise patch-tensor $\mathcal{N} $ and background patch-tenor $\mathcal{B} $. 

3) Image reconstruction. The target image $ {f_{\rm{T}}} $ and background image $ {f_{\rm{B}}} $ can be reconstructed by simple inverse operation.

4) Target detection. Due to the high pixel value of the true target in the reconstructed target image, we exploit the adaptive threshold segmentation approach in \cite{gao2013infrared} to extract small target.

\begin{equation}
{t_{th}} = \max \left( {{v_{\min }},\mu  + k\sigma } \right),
\end{equation}
where $ \sigma $ and $ \mu $ are the standard deviation and mean of the $ {f_{\rm{T}}} $ , respectively. $ k $ is a constant determined experimentally. $ {v_{\min }} = 0.85 $ is an adaptive value. If $ {f_{\rm{T}}}\left( {x,y} \right) > {t_{th}} $, then the pixel at $ \left( {x,y} \right)$ is considered as the target.

 \begin{figure*}[htp]
 	\centering
 	
 		\includegraphics[width=16cm]{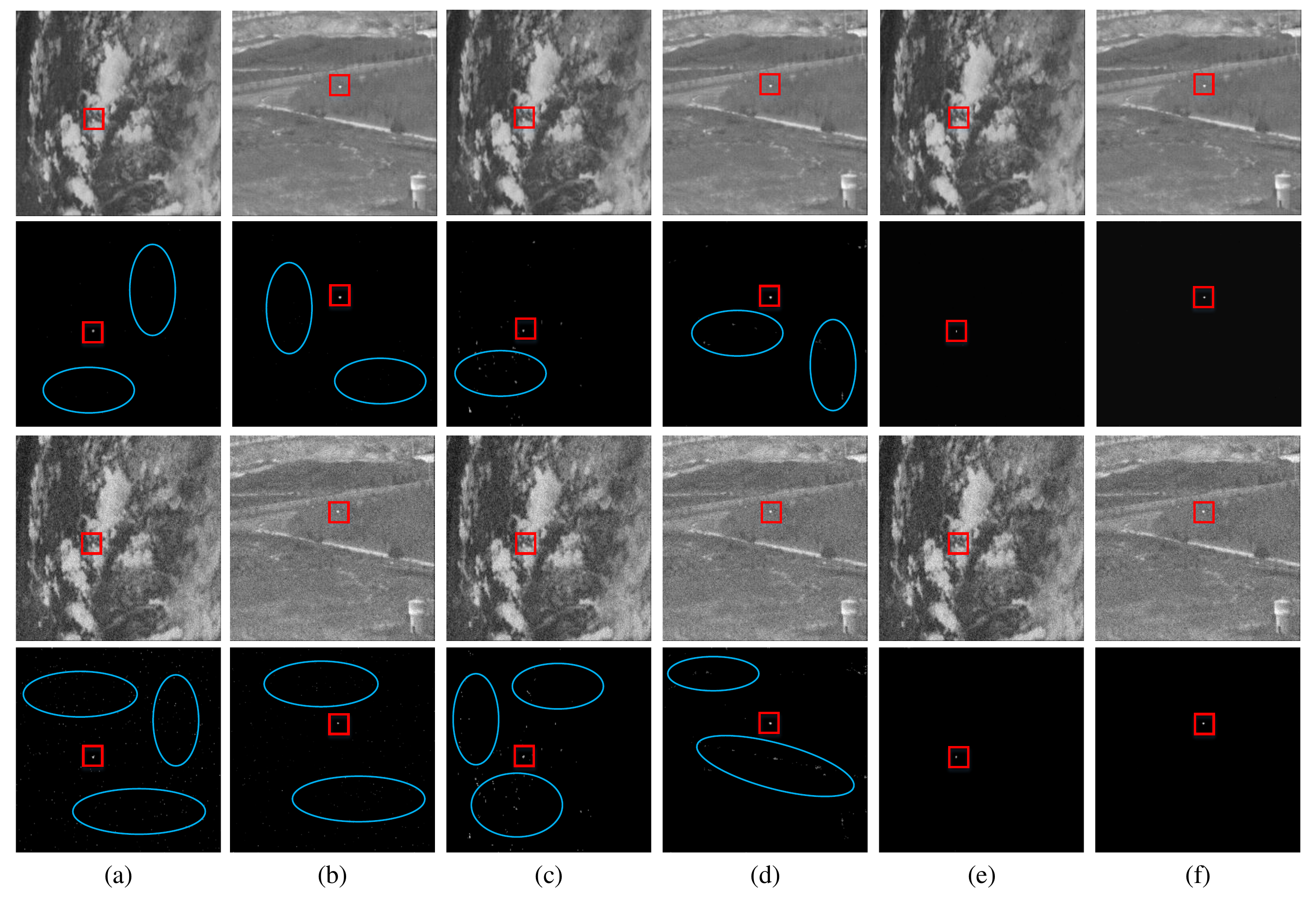}
 		\caption{Ablation experiments of each regularization of ASTTV-NTLA method. Row 1 and 3: the noisy images. Row 2 and 4: the corresponding results. Results in (a) and (b) are achieved by our model without Frobenius norm. Results in (c) and (d) are achieved  by our model without ASTTV regularization. Results in (e) and (f) are achieved by our model. The blue ellipses denote noise and background residuals.} \label{figDepth}
 
 \end{figure*}
 
\subsection{Complexity analyses}
In this subsection, we concisely analyze the computational complexity of the ASTTV-NTLA method. For the input image sequence $ {d_1},{d_2}, \cdots ,{d_p} \in {\mathbb{R}^{{n_1} \times {n_2}}}$, we can obtain $ s = {p \mathord{\left/	{\vphantom {p {{n_3}}}} \right.\kern-\nulldelimiterspace} {{n_3}}} $, and the dimension of each tensor is $ \mathcal{D} \in {\mathbb{R}^{{n_1} \times {n_2} \times {n_3}}} $. In ASTTV-NTLA algorithm, the main cost is to update $ \mathcal{Z} $ and $ \mathcal{B} $. Updating $ \mathcal{Z} $ requires performing FFT and $ \left[ {\frac{{{n_3} + 1}}{2}} \right] $ SVDs of $ {n_1} \times {n_2} $ matrices in each iteration by t-SVT, which cost $ \mathcal{O}\left( {ks{n_1}{n_2}{n_3}\left( {\log {n_3} + {{({n_2}\left[ {\frac{{{n_3} + 1}}{2}} \right])} \mathord{\left/{\vphantom {{({n_2}\left[ {\frac{{{n_3} + 1}}{2}} \right])} {{n_3}}}} \right.\kern-\nulldelimiterspace} {{n_3}}}} \right)} \right) $. Updating $ \mathcal{B} $ requires performing FFT operation, which cost $ \mathcal{O}\left( {ks{n_1}{n_2}{n_3}\log \left( {{n_3}} \right)} \right) $. The $ k $ denotes the iteration times. In summary, the computational complexity of each iteration is $ O\left( {ks{n_1}{n_2}{n_3}\left( {2\log {n_3} + {{({n_2}\left[ {\frac{{{n_3} + 1}}{2}} \right])} \mathord{\left/{\vphantom {{({n_2}\left[ {\frac{{{n_3} + 1}}{2}} \right])} {{n_3}}}} \right.\kern-\nulldelimiterspace} {{n_3}}}} \right)} \right) $. 

\subsection{Convergence analysis}
In this section, we analyze the convergence of the proposed ASTTV-NTLA method. Due to the existence of NTLA regularization, the solving process of Eq.(12) is actually a nonconvex optimization problem. In order to solve this problem, we use Theorem 3 in \cite{xie2016weighted}. In our method, we use an empirical convergence condition $ \frac{{\left\| {\mathcal{D} - {\mathcal{B}^{k + 1}} - {\mathcal{T}^{k + 1}} - {\mathcal{N}^{k + 1}}} \right\|_F^2}}{{\left\| \mathcal{D} \right\|_F^2}} \le \zeta  $ to analyze the convergence. Fig. 7 shows the convergence curve on Sequence 2. It can be seen from Fig. 7 that the value of the objective function converges to zero when $ k \ge 60 $. 

\begin{figure}[htb]
	\vspace{-0.2cm}
	\centering
	
	\includegraphics[width=6cm]{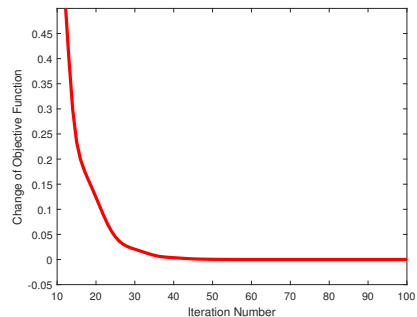}
	
	\caption{The change of objective function values versus the iteration number of ASTTV-NTLA.}
	\vspace{-0.2cm}
\end{figure}

\begin{figure*}[htbp]
	\vspace{-0.2cm}
	\centering
	
	\subfloat[Sequence 1]{
		\includegraphics[width=5cm]{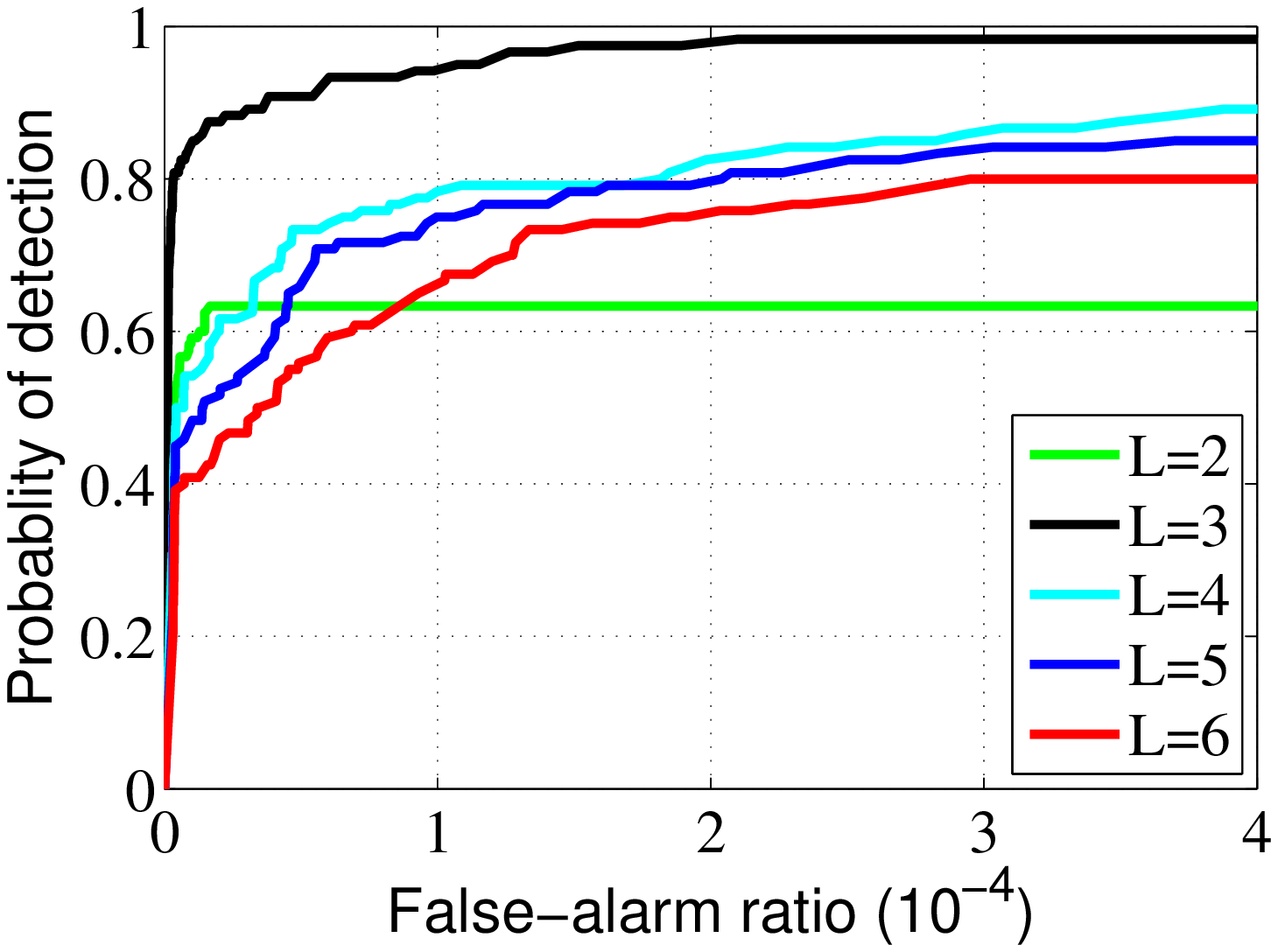}}\subfloat[Sequence 2]{
		\includegraphics[width=5cm]{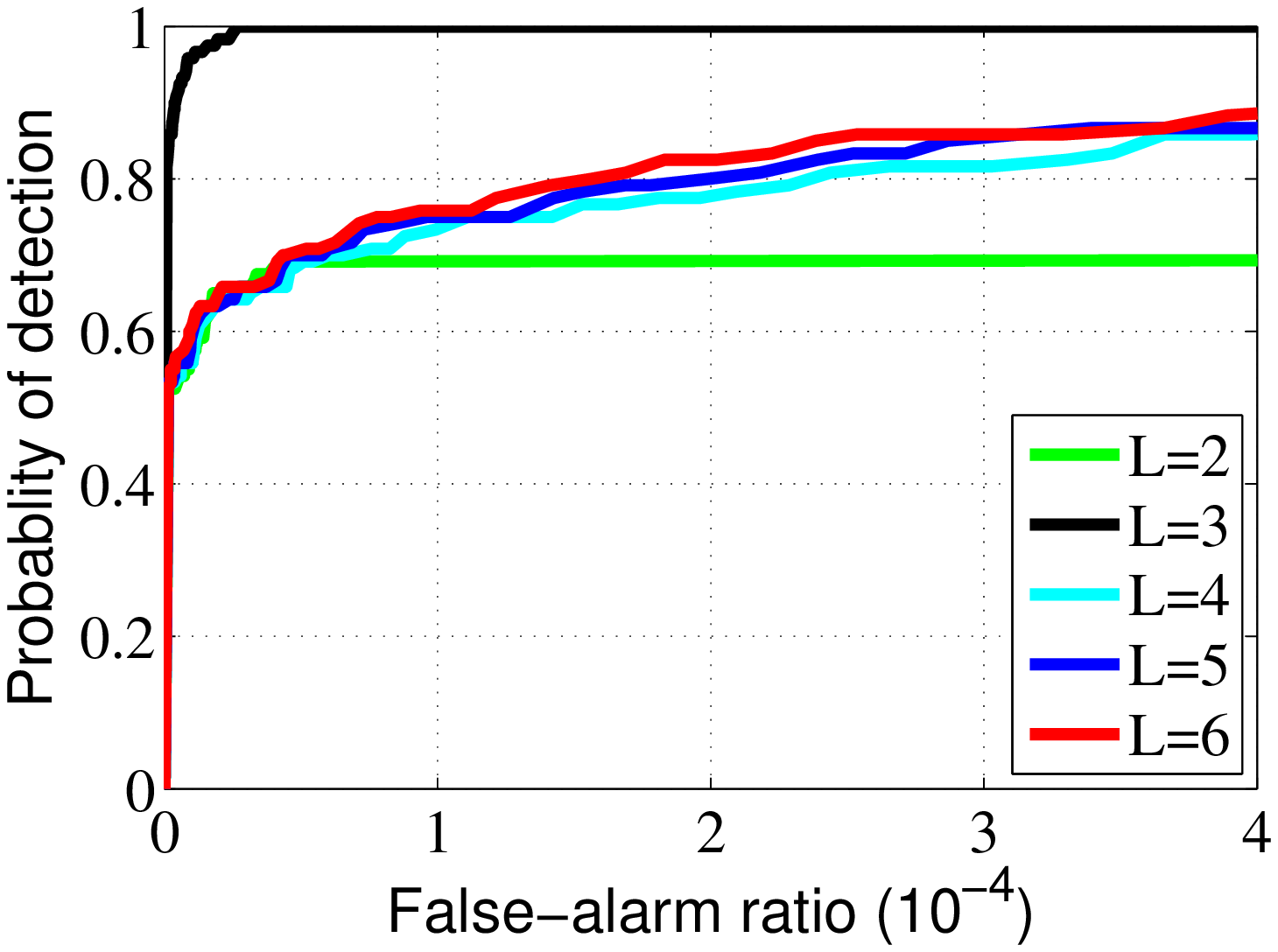}}\subfloat[Sequence 3]{
		\includegraphics[width=5cm]{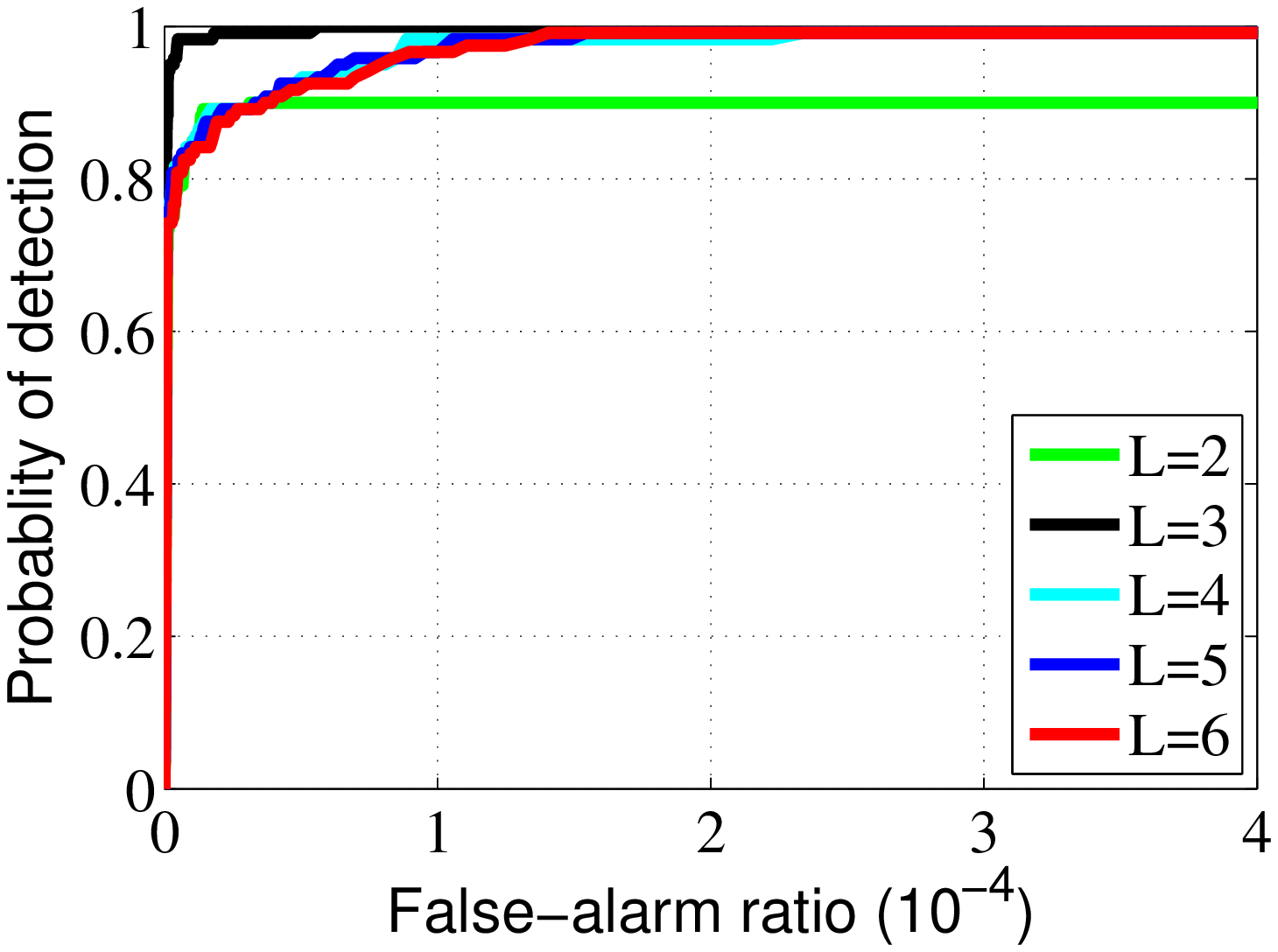}}
	\quad
	\subfloat[Sequence 4]{
		\includegraphics[width=5cm]{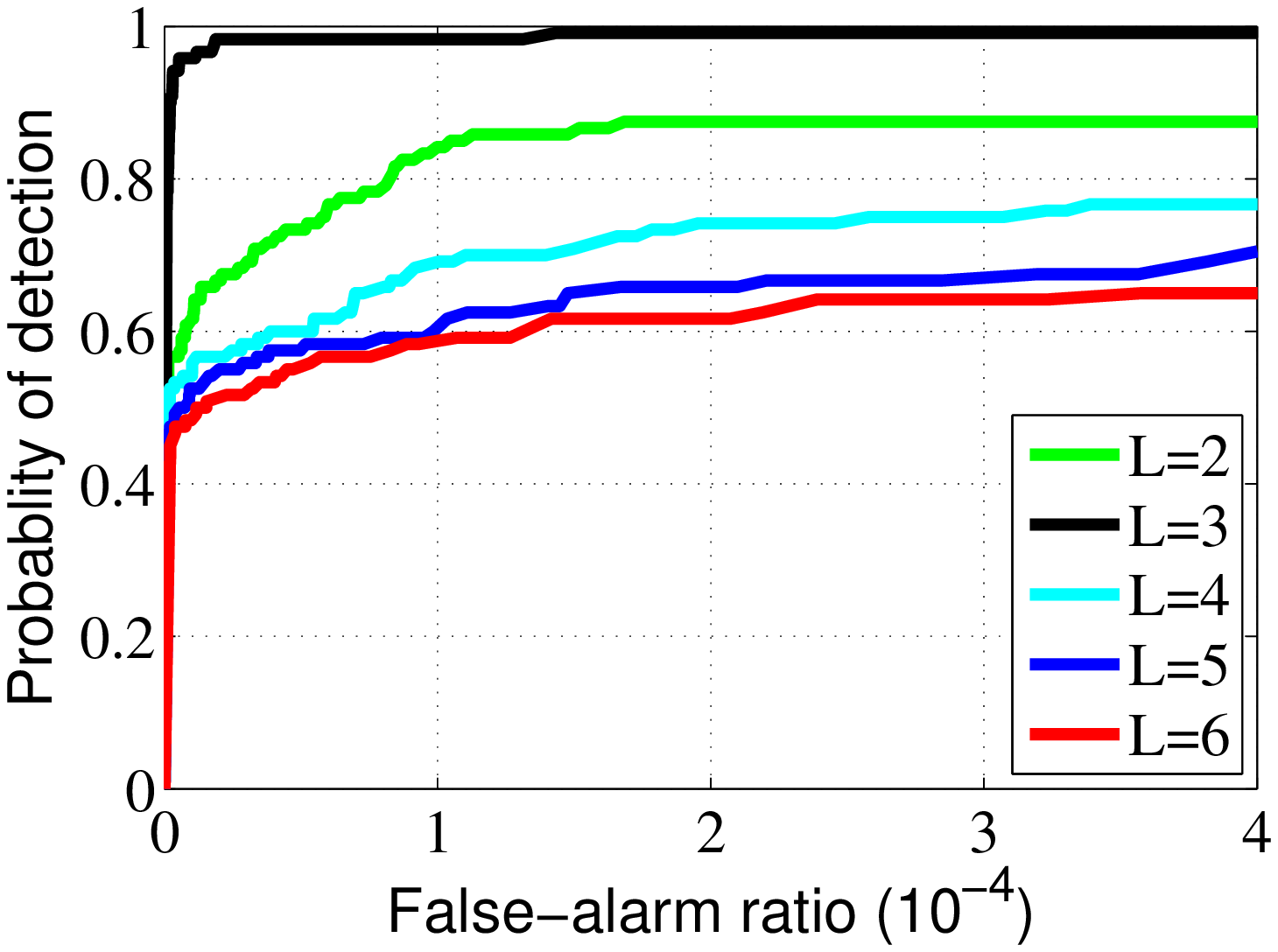}}\subfloat[Sequence 5]{
		\includegraphics[width=5cm]{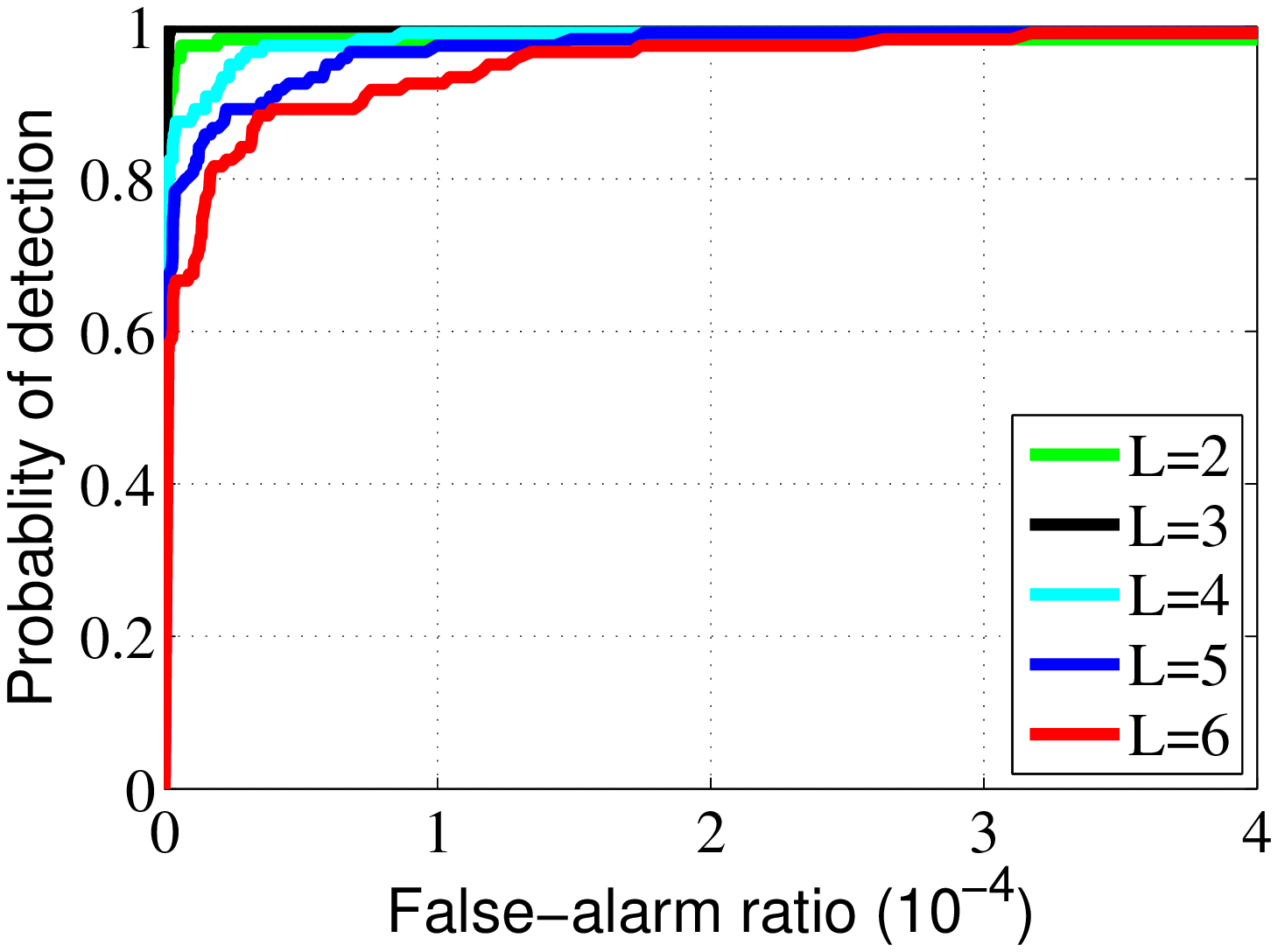}}\subfloat[Sequence 6]{
		\includegraphics[width=5cm]{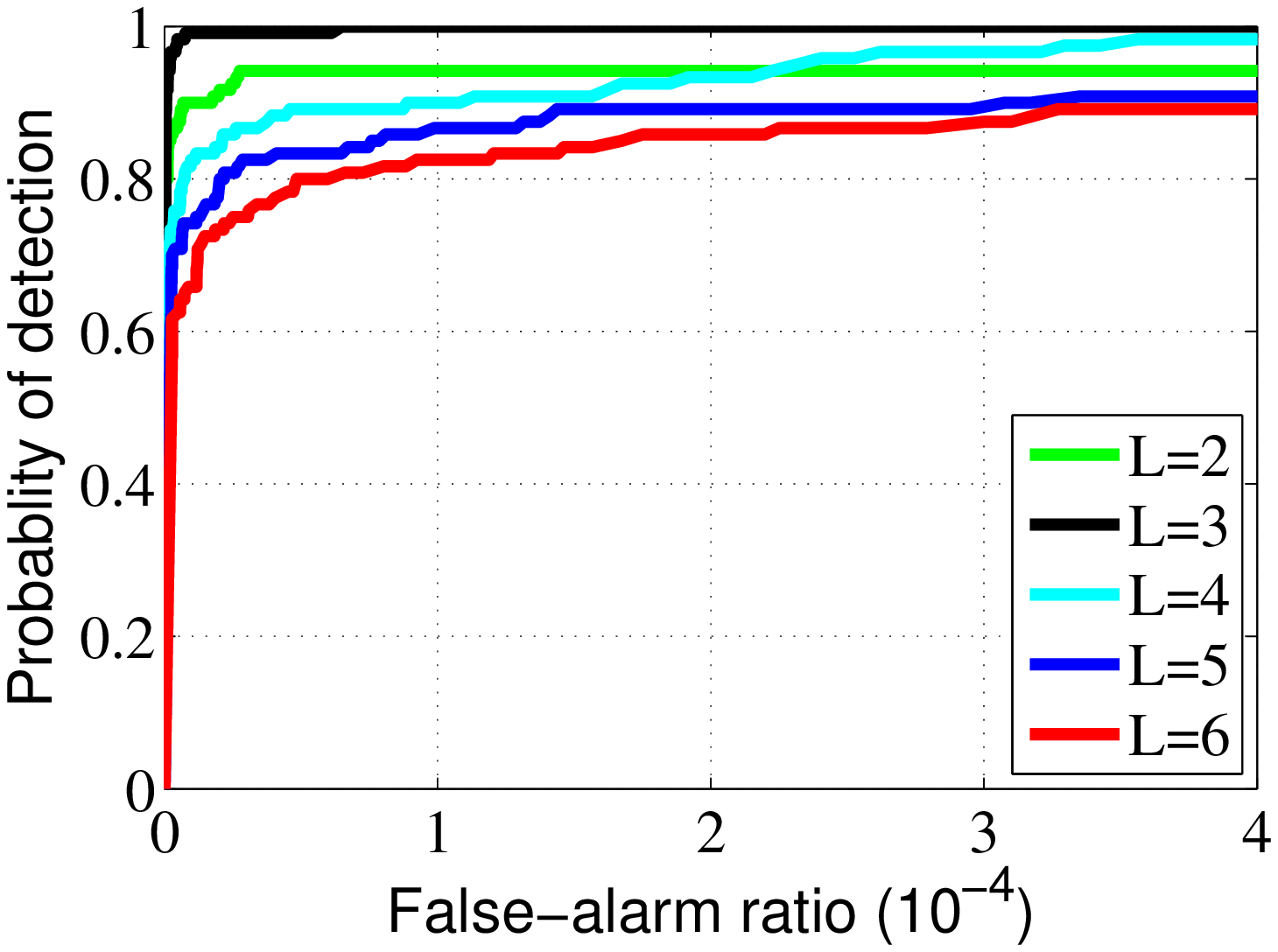}}
	\caption{ ROC curves with respect to different $ L  $.}
	\vspace{-0.2cm}
\end{figure*}

\section{Experimental results and analyses}
This section is mainly composed of the following parts. First, it introduces four relevant evaluation metrics and ten baseline methods, and then analyzes several important parameters in the ASTTV-NTLA method. Finally, the advantages of the ASTTV-NTLA method and the baseline methods are compared in various scenes.

\begin{figure}[htp]
	\vspace{-0.2cm}
	\centering
	\includegraphics[width=6cm]{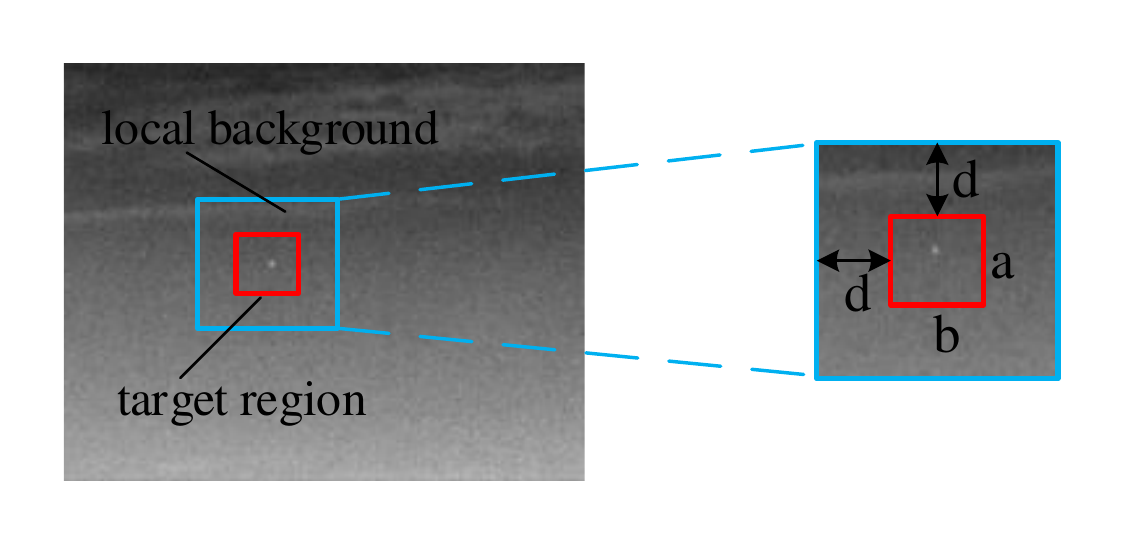}
	\caption{ The neighboring background area of target.}
	\vspace{-0.2cm}
\end{figure}

\begin{figure*}[htbp]
	\vspace{-0.2cm}
	\centering
	
	\subfloat[Sequence 1]{
		\includegraphics[width=5cm]{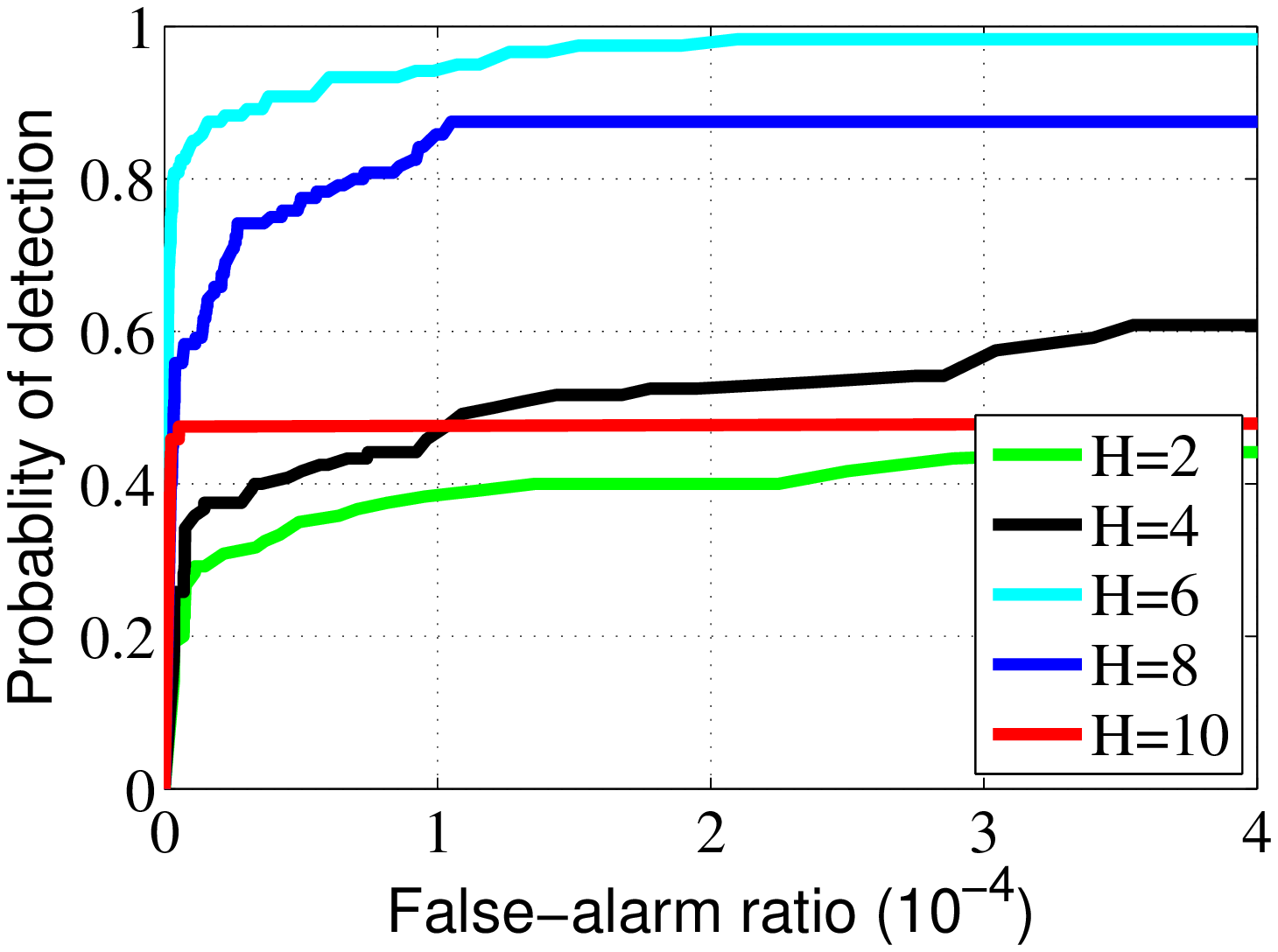}}\subfloat[Sequence 2]{
		\includegraphics[width=5cm]{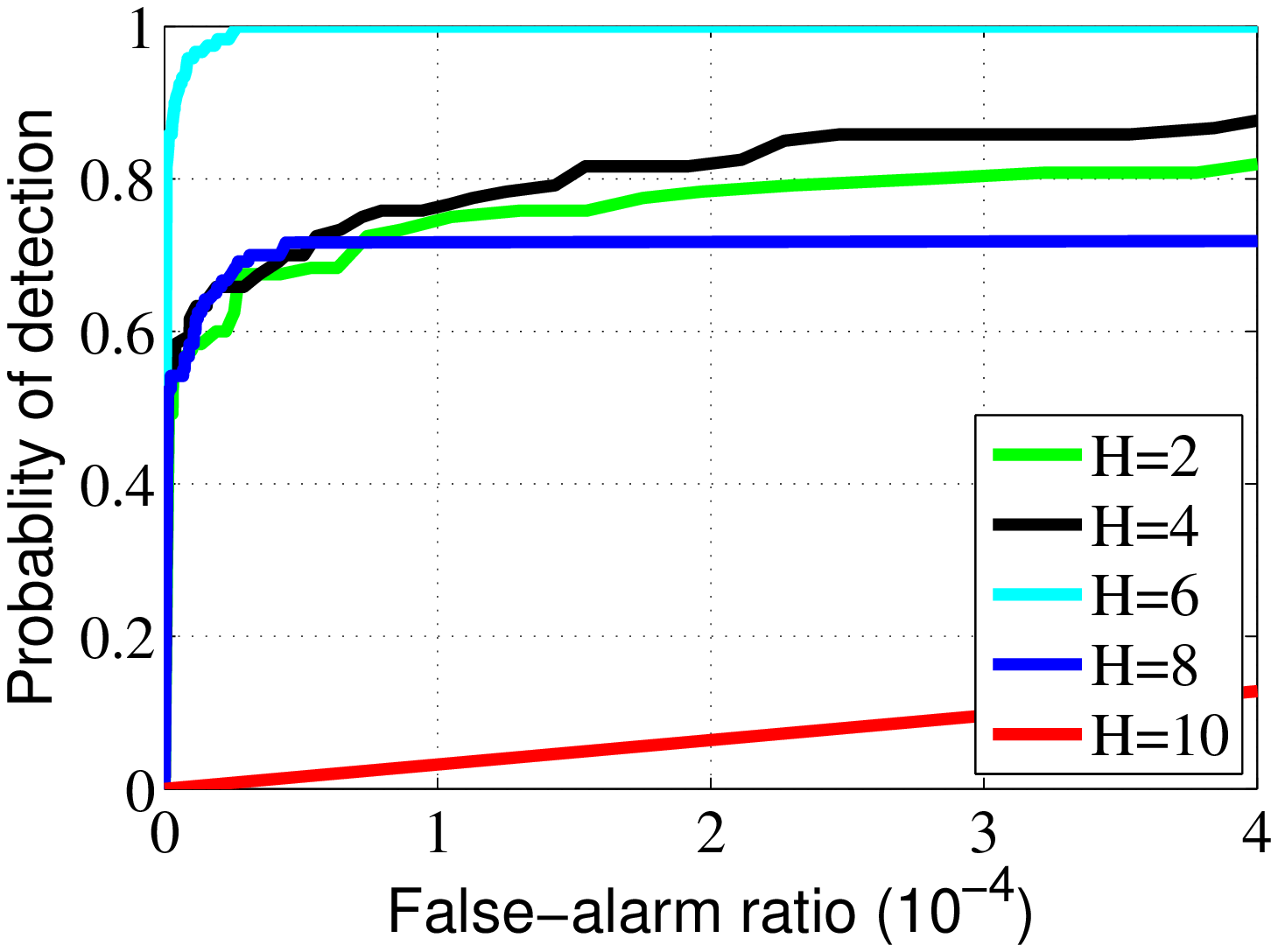}}\subfloat[Sequence 3]{
		\includegraphics[width=5cm]{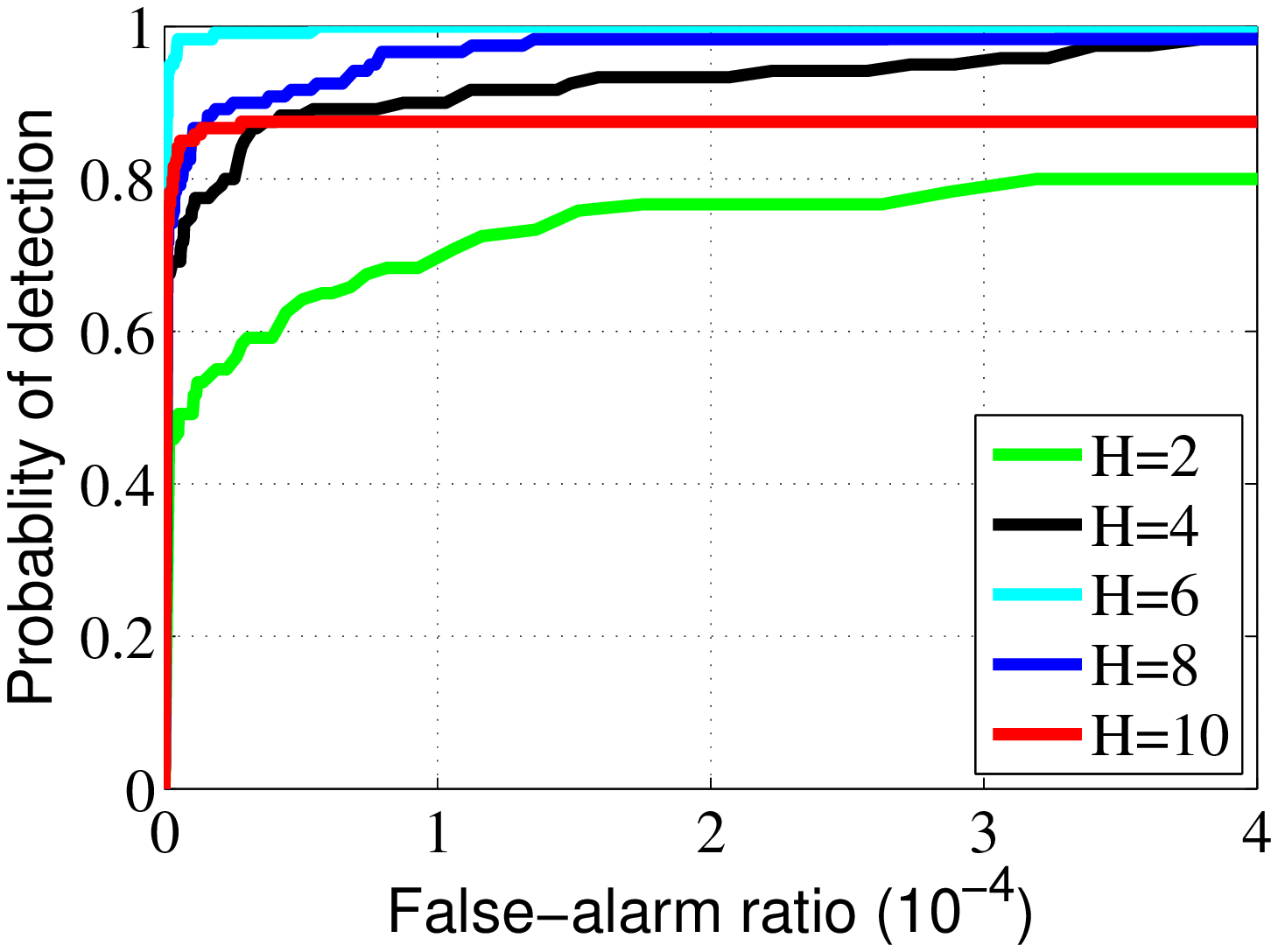}}
	\quad
	\subfloat[Sequence 4]{
		\includegraphics[width=5cm]{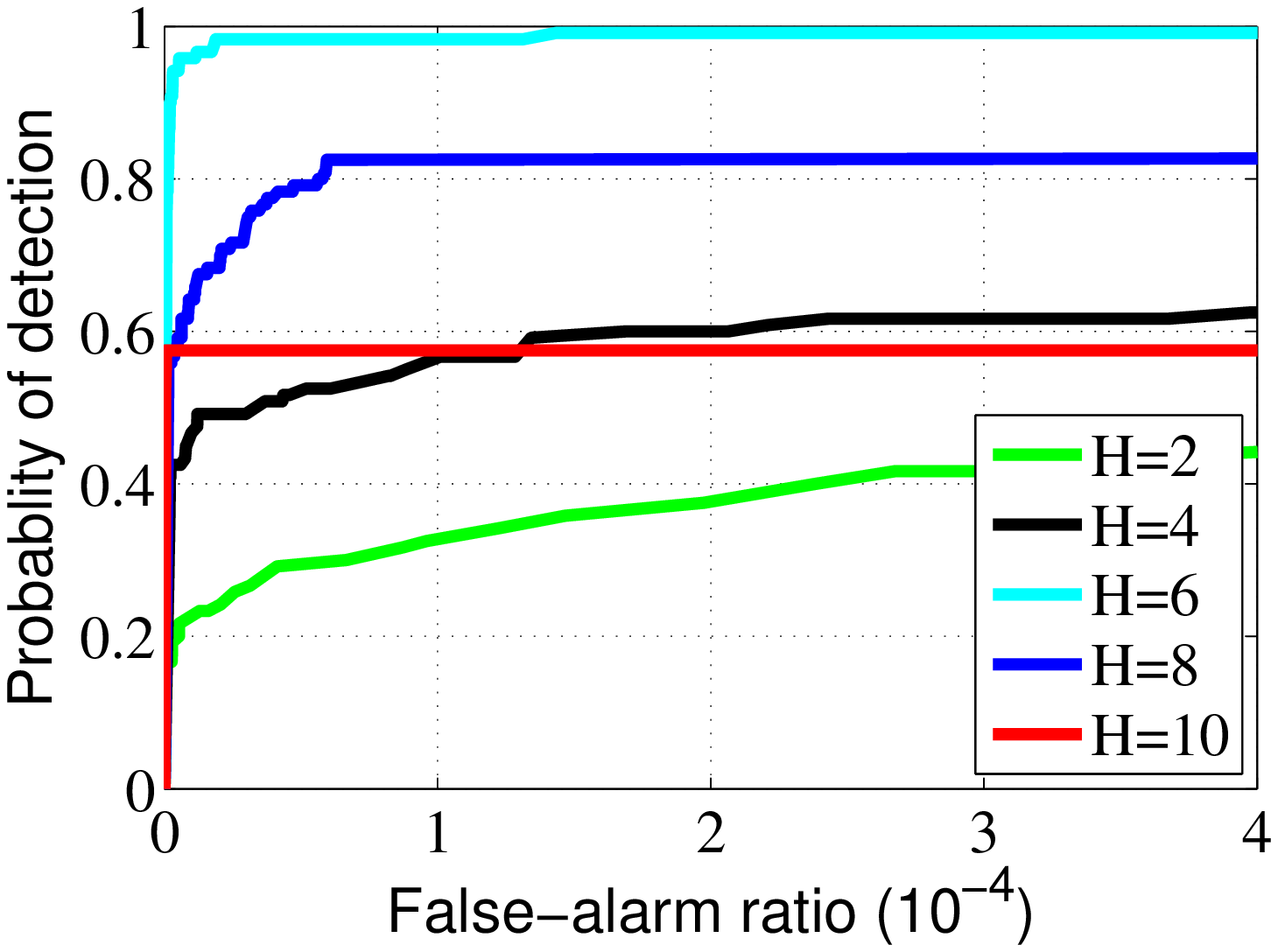}}\subfloat[Sequence 5]{
		\includegraphics[width=5cm]{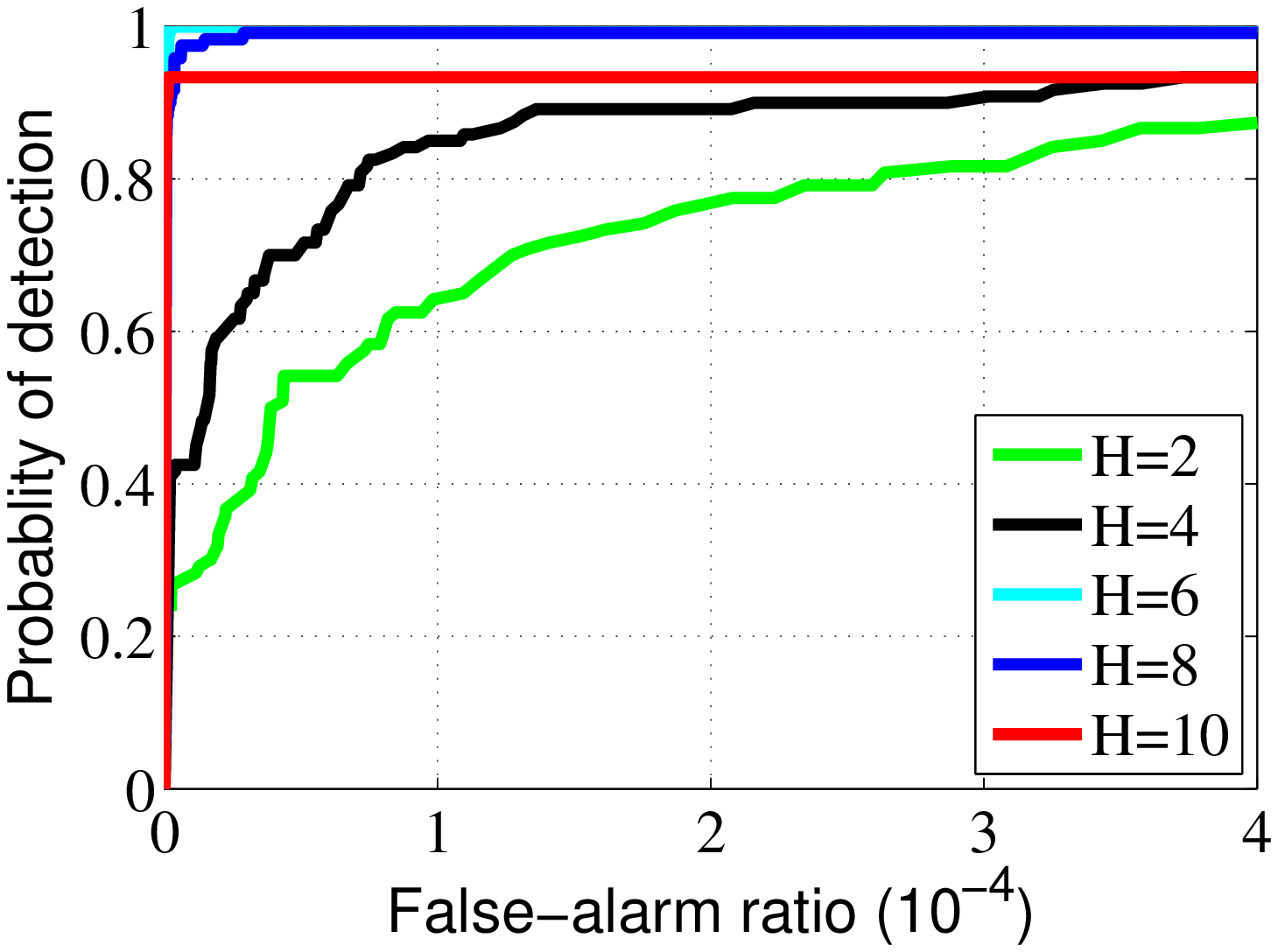}}\subfloat[Sequence 6]{
		\includegraphics[width=5cm]{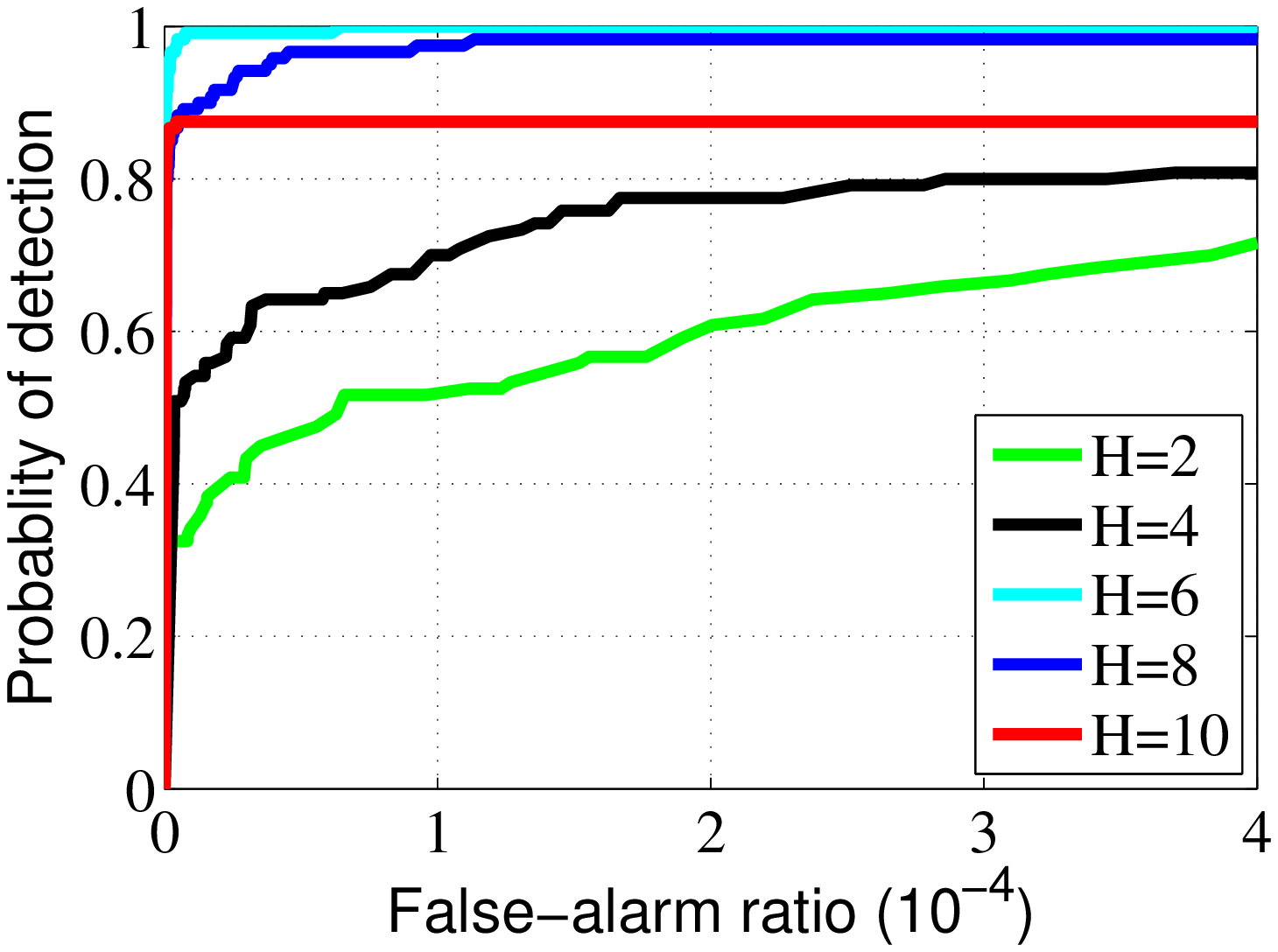}}
	\caption{ ROC curves with respect to different $ H $.}
	\vspace{-0.2cm}
\end{figure*}

\subsection{Evaluation Metrics and Baseline Methods}
To quantitatively validate the detection performance of the ASTTV-NTLA method, we use four evaluation metrics including the background suppression factor (BSF), local signal to noise ratio gain (LSNRG), contrast gain (CG) and signal to clutter ratio gain (SCRG). They are used to evaluate detection performance and background suppression ability. LSNRG is usually used to describe the local signal to noise ratio (LSNR) gain as follows
\begin{equation}
{\rm{LSNRG = }}\frac{{{\rm{LSN}}{{\rm{R}}_{{\rm{out}}}}}}{{{\rm{LSN}}{{\rm{R}}_{{\rm{in}}}}}},
\end{equation}
where ${{\rm{LSN}}{{\rm{R}}_{{\rm{out}}}}}$ and ${{\rm{LSN}}{{\rm{R}}_{{\rm{in}}}}}$ represent LSNR values before and after processing respectively. $ {\rm{LSNR = }}{{{P_{\rm{T}}}} \mathord{\left/
		{\vphantom {{{P_{\rm{T}}}} {{P_{\rm{B}}}}}} \right.
		\kern-\nulldelimiterspace} {{P_{\rm{B}}}}} $. The maximum pixel values of the neighborhood and the target are represented by ${{P_{\rm{B}}}}$ and  ${{P_{\rm{T}}}}$, respectively. Then, BSF compares the background suppression ability, which is expressed as:
\begin{equation}
{\rm{BSF = }}\frac{{{\sigma _{{\rm{in}}}}}}{{{\sigma _{{\rm{out}}}}}},
\end{equation}
where $ {{\sigma _{{\rm{out}}}}}$ and $ {{\sigma _{{\rm{in}}}}}$ are the standard variance of the neighboring background region of target image and original image, respectively. The SCRG denotes the signal-to-clutter ratio (SCR) before and after processing, which is defined as:
\begin{equation}
{\rm{SCRG = }}\frac{{{\rm{SC}}{{\rm{R}}_{{\rm{out}}}}}}{{{\rm{SC}}{{\rm{R}}_{{\rm{in}}}}}},
\end{equation}
where SCR uses the same expression as \cite{gao2012small}:
\begin{equation}
{\rm{SCR = }}\frac{{\left| {{\mu _{\rm{t}}}{\rm{ - }}{\mu _{\rm{b}}}} \right|}}{{{\sigma _{\rm{b}}}}},
\end{equation}
where $ {{\mu _{\rm{t}}}}$ , $ {{\mu _{\rm{b}}}}$ and $ {{\sigma _{\rm{b}}}}$ represent the average value of the target area, the pixel average value and standard deviation of the surrounding local neighborhood region, respectively. It can be seen from Fig. 9 that $d$ and $ a \times b $ denote the size of the width of the adjacent area and the target region, respectively. Meanwhile, $ (a + 2d) \times (b + 2d)$ represents the size of the neighboring area. In our paper, $ d =40 $, $ a=b=9 $. To compare the ability of gray difference between expanded target and background, we introduce CG evaluation metric\cite{gao2018infrared}, which is expressed as:
	
\begin{equation}
{\rm{CG = }}\frac{{{\rm{CO}}{{\rm{N}}_{{\rm{out}}}}}}{{{\rm{CO}}{{\rm{N}}_{{\rm{in}}}}}},
\end{equation}
where $ {{\rm{CO}}{{\rm{N}}_{{\rm{out}}}}}$ and $ {{\rm{CO}}{{\rm{N}}_{{\rm{in}}}}} $ denote the contrast (CON) of the target and original images, respectively, and CON is defined as:
\begin{equation}
{\rm{CON}} = \left| {{\mu _t} - {\mu _b}} \right|.
\end{equation}
Generally speaking, the higher values of the above four evaluation metrics indicate that the method has better background suppression ability. It is worth noting that LSNRG, BSF and SCRG are the evaluation metrics to describe the local neighborhood suppression ability, not the global suppression ability. In addition, detection probability $ {P_d} $ and false-alarm rate $ {F_a} $ are also important evaluation indicators, which are defined as \cite{gao2013infrared}:
\begin{equation}
{P_d} = \frac{{{\rm{number \:of\: true\: detections}}}}{{{\rm{number\: of\: actual\: targets}}}}
\end{equation}
\begin{equation}
{F_a} = \frac{{{\rm{number\:of \:false\: detections}}}}{{{\rm{number\: of\: image \: pixels}}}}.
\end{equation}
The above two indicators range between 0 and 1.

\begin{table*}[htbp]
	\vspace{-.05in}
	\scriptsize
	\centering
	\renewcommand\arraystretch{1.1}
	\caption{DETAILED PARAMETER SETTING FOR TESTED METHODS. }\label{ablation1}
	\begin{tabular}{lccccccc}
		\hline
		Methods &  Acronyms  & Parameter settings \\[0.05cm]
		\hline
		Top-Hat method  & Top-Hat & Structure size: $ 3 × 3 $, structure shape: square  \\
		Weighted strengthened local contrast measure & WSLCM & $ K = 9 $, gaussian filter size: $ 3 \times 3 $\\
		Non-Convex Rank Approximation Minimization & NRAM & \makecell[c]{  Sliding step: 10, $ \lambda  = \frac{1}{{\sqrt {\max \left( {M,N} \right)} }} $, patch Size: $ 50 \times 50 $, $ \gamma  = 0.002$, \\ $ C = {{\sqrt {\min \left( {M,N} \right)} } \mathord{\left/
					{\vphantom {{\sqrt {\min \left( {M,N} \right)} } {2.5}}} \right.
					\kern-\nulldelimiterspace} {2.5}} $, $ {\mu ^0} = 3\sqrt {\min \left( {M,N} \right)} $, $ \varepsilon  = 1e - 7 $}\\
		Total Variation Regularization and Principal Component Pursuit & TV-PCP & $ {\lambda _1} = 0.005 $,  $ {\lambda _2} = \frac{1}{{\sqrt {\max \left( {M,N} \right)} }} $, $ \beta  = 0.025 $, $ \gamma  = 1.5 $ \\
		Reweighted Infrared Patch-Tensor Model & RIPT & \makecell[c]{Sliding step: 10, $ \lambda  = \frac{L}{{\sqrt {\min \left( {{n_1},{n_2},{n_3}} \right)} }} $, patch size: $ 50 \times 50 $,  \\ $ h = 10 $, $ \varepsilon  = 1e - 7 $, $ L = 1 $} \\
		Partial Sum of the Tensor Nuclear Norm & PSTNN & Sliding step: 40, $ \lambda  = \frac{{0.6}}{{\sqrt {\max \left( {{n_1},{n_2}} \right) * {n_3}} }}$, patch size: $ 40 \times 40 $, $ \varepsilon  = 1e - 7 $  \\
		\makecell[lc] {Spatial-temporal Total Variation Regularization \\and weighted Tensor Nuclear Norm }& STTV-WNIPT & $ L = 3 $, $ H = 8 $,$ {\lambda _1} = 0.005 $, $ {\lambda _2} = \frac{H}{{\sqrt {\max \left( {M,N} \right) * L} }} $, $ {\lambda _3} = 100 $ \\
		Infrared Videos Based on Spatio-Temporal Tensor Model  & IVSTTM &  Sliding step: 15, patch size: $ 80 \times 80 $, spatial patch cube: ${m_s} = 9$ \\
		Multiple Subspace Learning and Spatial-temporal Patch-Tensor Model & MSLSTIPT & $ L = 6 $, $ p = 0.8 $, $ \lambda  = {1 \mathord{\left/
				{\vphantom {1 {\sqrt {{n_3}\max ({n_1},{n_2})} }}} \right.
				\kern-\nulldelimiterspace} {\sqrt {{n_3}\max ({n_1},{n_2})} }} $, patch size: $ 30 \times 30 $ \\
	   Nonconvex tensor fibered rank approximation & NTFRA & Sliding step: 40, patch size: $ 40 \times 40 $, $ \lambda  = {1 \mathord{\left/
				{\vphantom {1 {\sqrt {{n_3}\max ({n_1},{n_2})} }}} \right.
				\kern-\nulldelimiterspace} {\sqrt {{n_3}\max ({n_1},{n_2})} }} $, $\beta  = 0.01$ \\
		\makecell[lc]{Non-Convex Tensor Low-Rank Approximation } & ASTTV-NTLA & $ L = 3 $, $ H = 6 $,$ {\lambda _{tv}} = 0.005 $, $ {\lambda _s} = \frac{H}{{\sqrt {\max \left( {M,N} \right) * L} }} $, $ {\lambda _3} = 100 $\\  	
		
		\hline
	\end{tabular}
\end{table*}

In order to estimate the advantages of the ASTTV-NTLA method, we compare it with ten other methods in the above metrics and various scenes. These methods are mainly divided into the following categories: BS-based methods (Top-Hat\cite{rivest1996detection}), HVS-based methods (WSLCM\cite{han2020infrared}), and recently developed LRSD-based methods (NRAM\cite{zhang2018infrared}, TV-PCP \cite{wang2017infrared1}, RIPT \cite{dai2017reweighted}, PSTNN \cite{zhang2019infrared1}, STTV-WNIPT \cite{sun2019infrared}, IVSTTM \cite{liu2020small}, MSLSTIPT\cite{sun2020infrared}, NTFRA\cite{kong2021infrared}). Among the above methods, the single-frame method using only spatial information includes Top-Hat, WSLCM, NRAM, TV-PCP, RIPT, PSTNN and NTFRA. Because the proposed ASTTV-NTLA method uses spatial-temporal information, we compare our method with three methods using spatio-temporal information, namely STTV-WNIPT, IVSTTM, MSLSTIPT. Table I shows the detailed parameter settings of the comparison method in this paper. All experiments were conducted on a computer with a 16 GB RAM and an Inter Core i7-10870H CPU (2.20 GHz). Our method was implemented in MATLAB 2014a, and the codes are available at https://github.com/LiuTing20a/ASTTV-NTLA. 

\begin{figure*}[htbp]
	\vspace{-0.2cm}
	\centering
	
	\subfloat[Sequence 1]{
		\includegraphics[width=5cm]{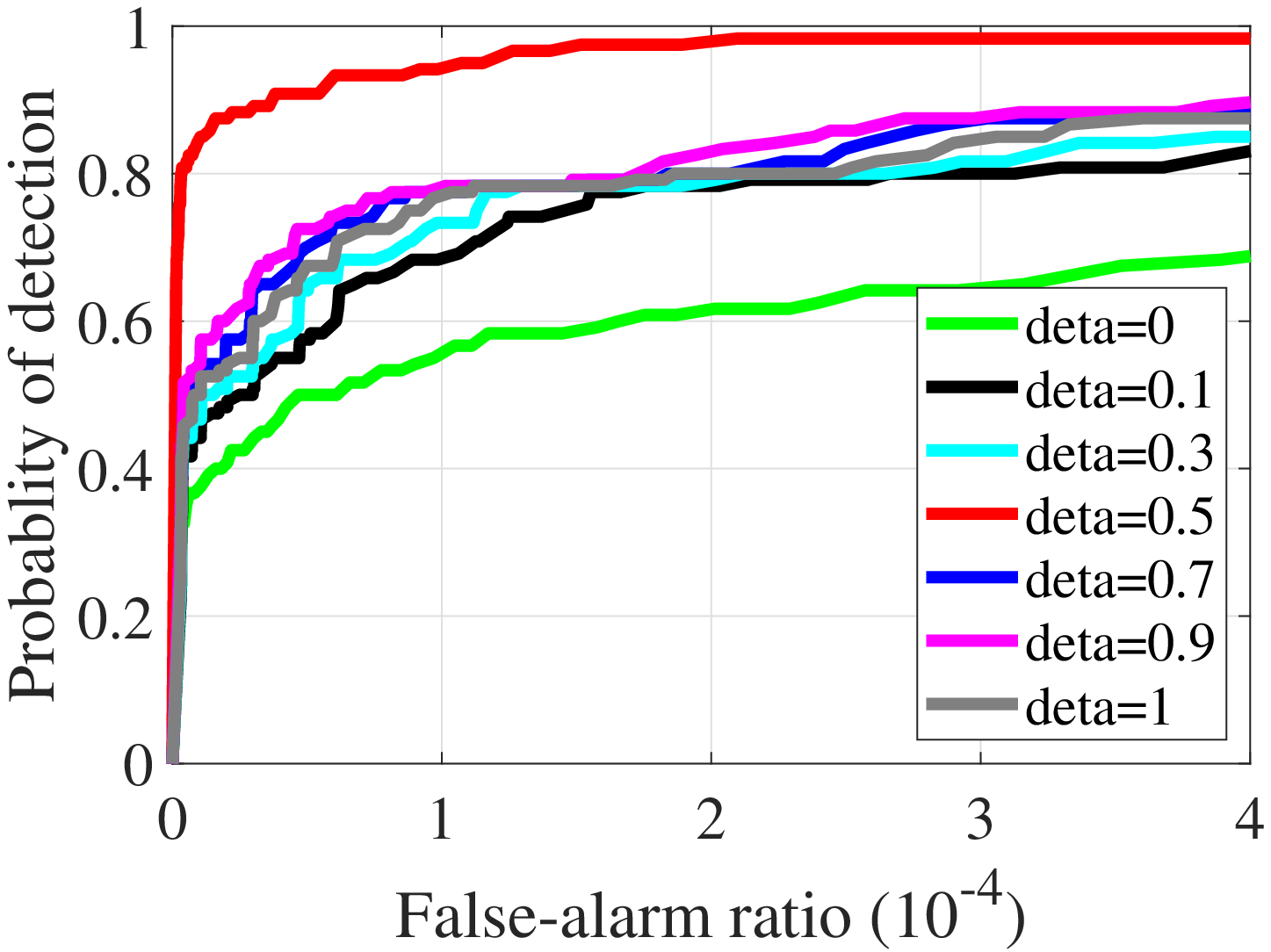}}\subfloat[Sequence 2]{
		\includegraphics[width=5cm]{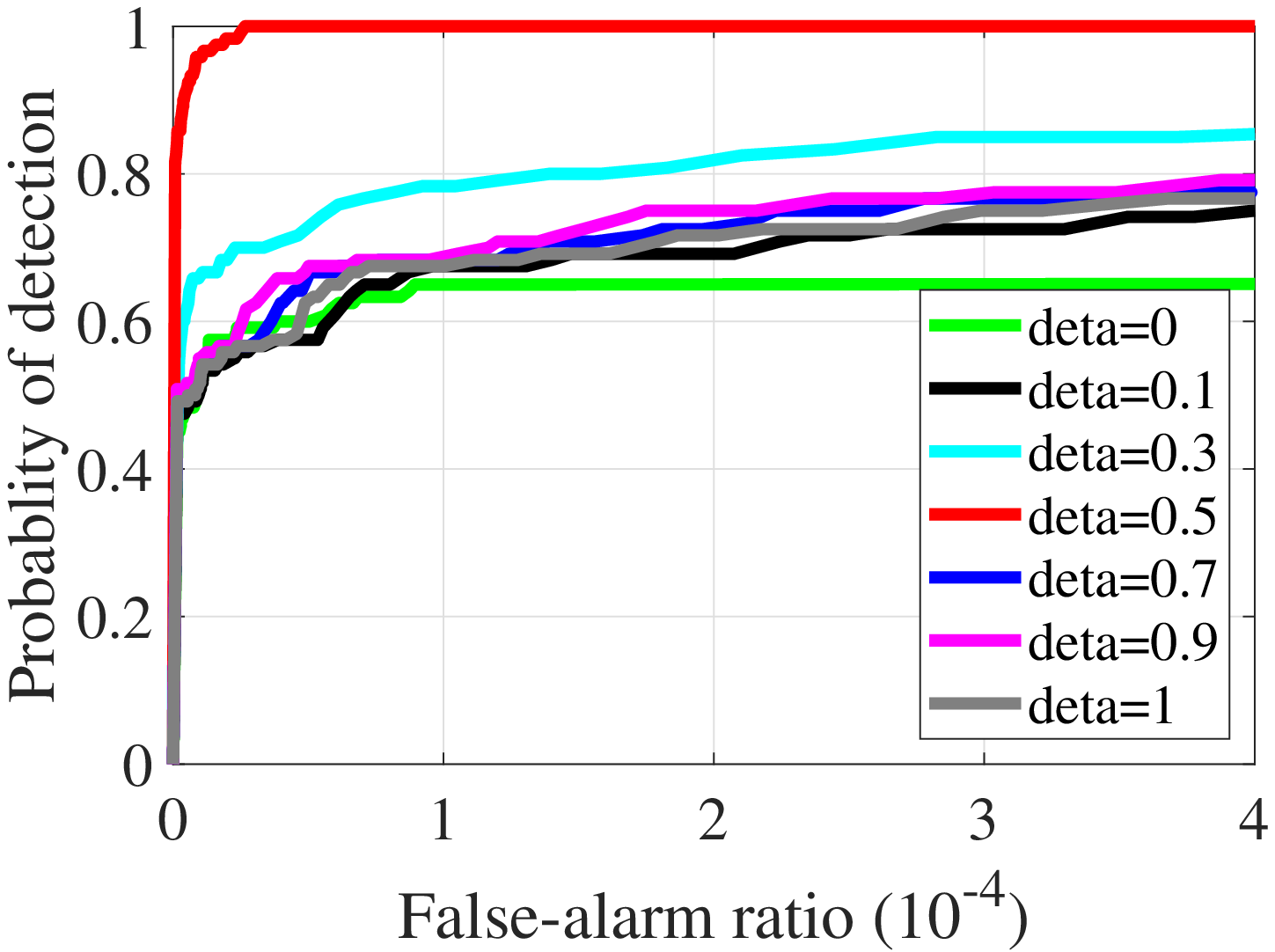}}\subfloat[Sequence 3]{
		\includegraphics[width=5cm]{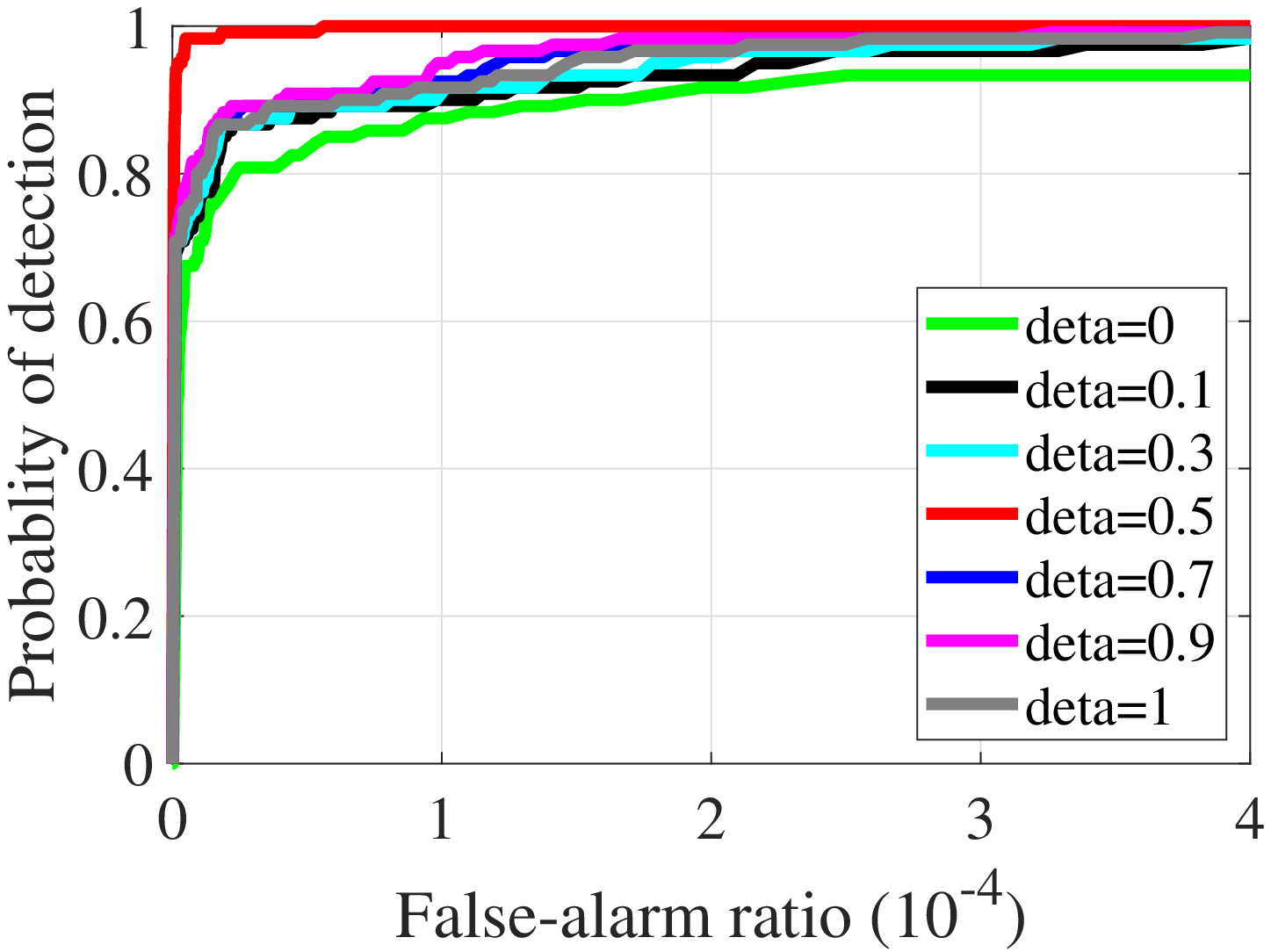}}
	\quad
	\subfloat[Sequence 4]{
		\includegraphics[width=5cm]{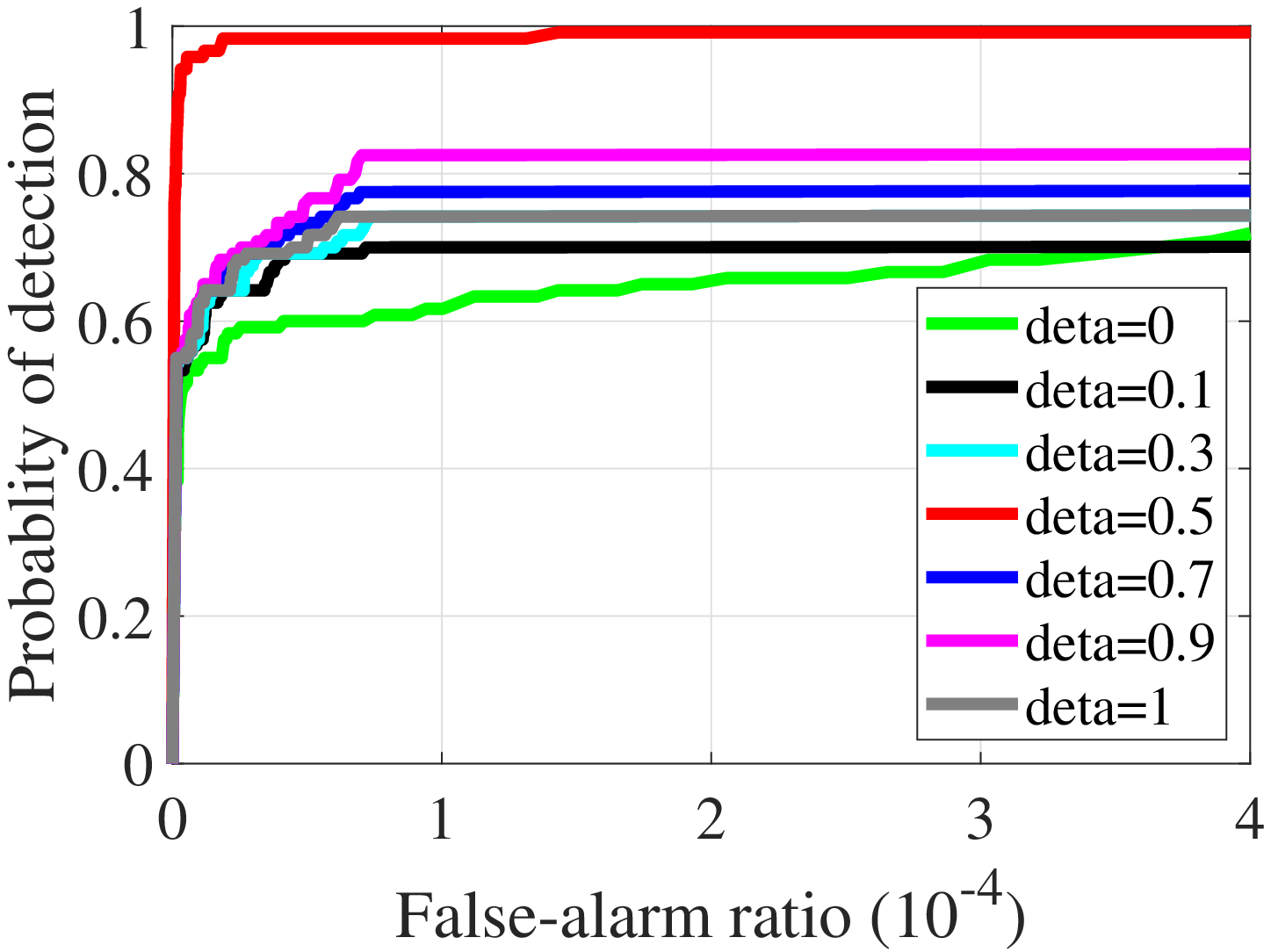}}\subfloat[Sequence 5]{
		\includegraphics[width=5cm]{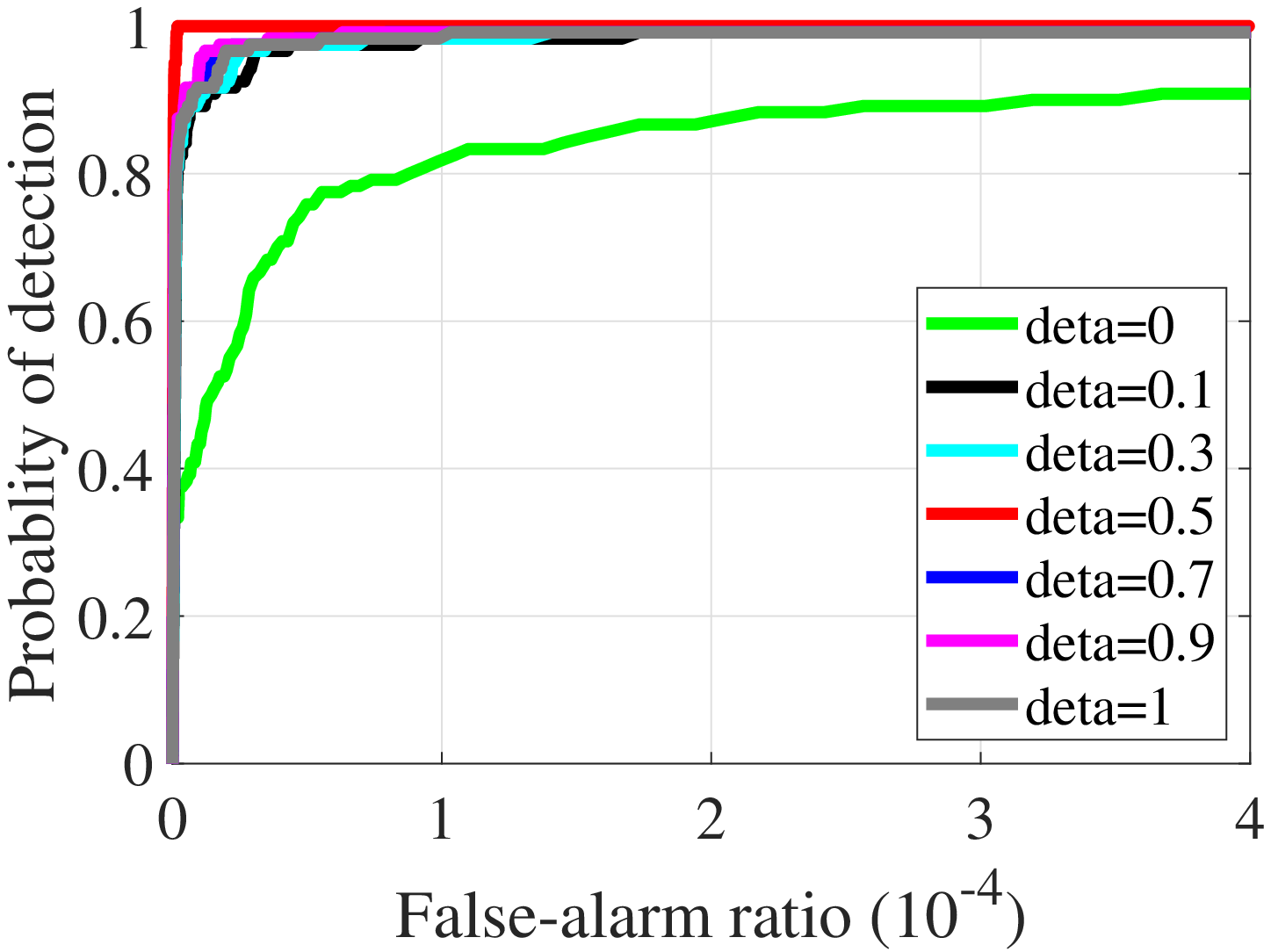}}\subfloat[Sequence 6]{
		\includegraphics[width=5cm]{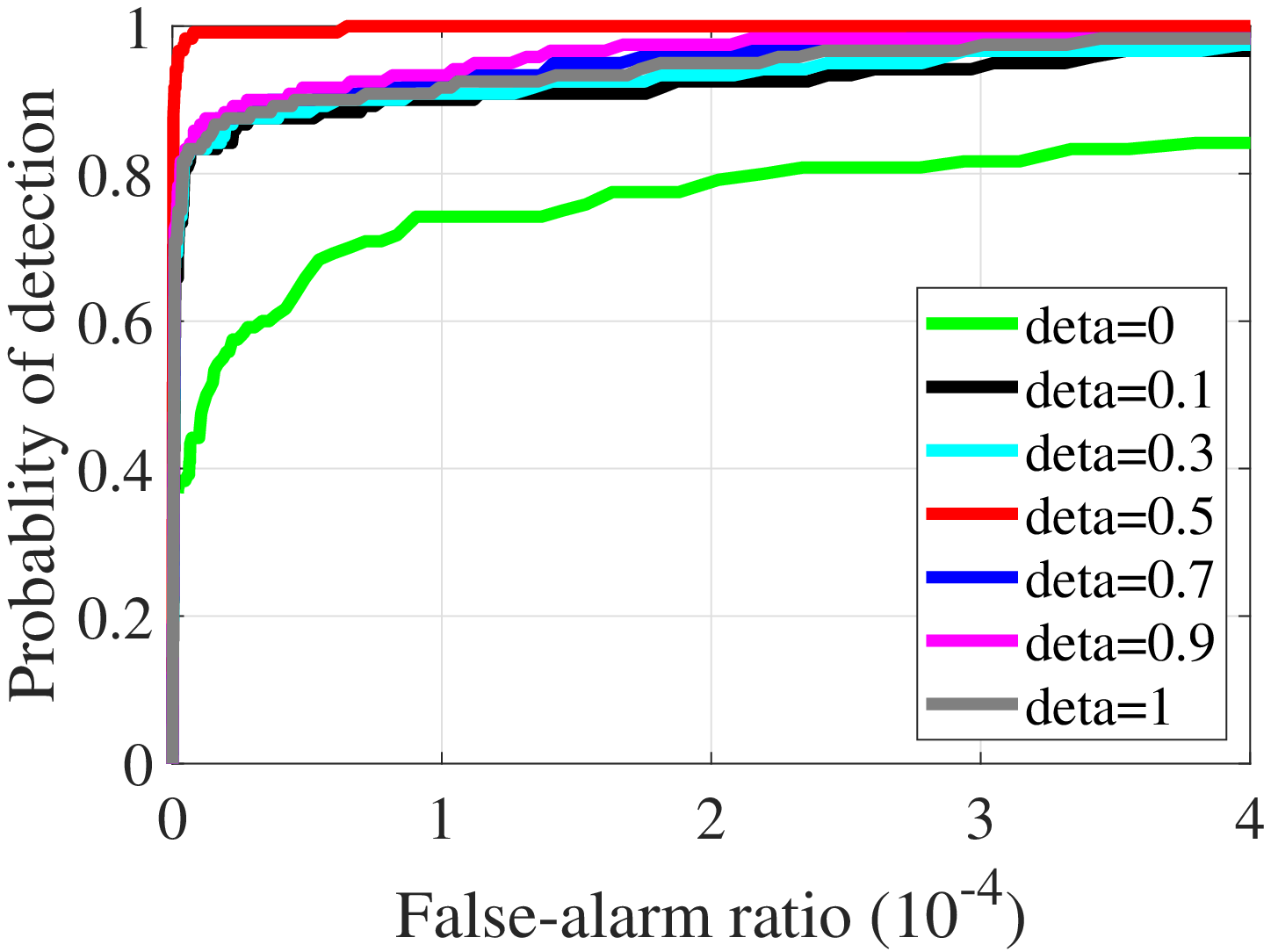}}
	\caption{ ROC curves with respect to different $ \delta $.}
	\vspace{-0.2cm}
\end{figure*}

\begin{table*}[t]
	\vspace{-.05in}
	\scriptsize
	\centering
	\renewcommand\arraystretch{1.1}
	\caption{CHARACTERISTICS OF the DATASET. }\label{ablation1}
	\begin{tabular}{lccccccc}
		\hline
		Sequence & Frames &   Image Size &  Target Size & Average SCR &  Target Descriptions  &  Background Descriptions  \\[0.05cm]
		\hline
		1   & 120 & $ 200 \times 250 $ & $ 3 \times 4 \sim 3 \times 6 $ & 4.79 & Fast-moving, tiny, regular shape & scenario with multilayer cloud, heavy noise \\
		2   & 120 & $ 200 \times 300 $ & $ 4 \times 4 \sim 5 \times 5 $ & 3.89 &Fast-moving, irregularly shaped aircraft & Fierce clouds and heavy noise\\
		3   & 120  & $ 200 \times 250 $ & $ 4 \times 3 \sim 4 \times 6 $ & 3.51 & Small and dim , quick motion & \makecell[c]{A blurred sealand background with  \\ a strong reflective artificial building}\\
		4    & 120 &  $ 250 \times 250 $ & $ 3 \times 3 \sim 5 \times 5 $ & 2.33 & Dim and slow-moving  airplane & Mountains with strong reflections \\
		5 & 120 & $ 250 \times 250 $ & $ 3 \times 3 \sim 3 \times 6 $ & 1.83 & Small and slow-moving airplane  & village, reflective road and roof\\
		6  & 120 & $ 250 \times 250 $ & $ 4 \times 3 \sim 5 \times 4 $ & 2.64 & Small and slow-moving airplane & Forest, reflective road and roof \\

		\hline
	\end{tabular}
\end{table*}

\subsection{Parameter Setting and Datasets}
In our model, we test on a synthetic data to determine the values of several important parameters. The regularized parameter $ {\lambda _{tv}}$ could balance the tradeoff between the non-convex tensor low-rank approximation and ASTTV regularization, and it is empirically set to 0.005 following\cite{sun2018novel}. In most cases, the range of $ \delta $ is [0,1]. we follow\cite{tom2020simultaneous} to set $ \delta=0.5 $. 
Following \cite{lu2019tensor}, we set $ {\lambda _s} = \frac{H}{{\sqrt {\max \left( {m,n} \right) \times L} }}$. $ H $ represents a tunning parameter.
According to \cite{wang2017hyperspectral}, we set $ {\lambda _3}=100 $. In the following experiments, we further analyzed the influence of the parameters $ L $ , $ H $ and $ \delta $ on the detection performance. Section IV-D explains more details.

As can be seen from Table II, we simulated six image sequences come from diverse scenes to demonstrate the stability and detection ability of our method. In our datasets, the sky background with some clouds in Sequences 1 and 2 are relatively simple. In Sequence 3, the small target is moving in a blurred sealand background. The main challenges of Sequence 3 are the bright artificial buildings in the background. In Sequence 4, an airplane is flying towards the mountains with strong reflections at the bottom of the mountains. The main challenges of Sequence 4 are the high reflection intensity, which easily leads to miss detection and false alarm. The main challenges of Sequences 5 and 6 are the reflective road, roof and forests in the background. Therefore, our datasets consists of both simple and complex background. The diverse datasets with various scenes can help to comprehensively evaluate the performance of each algorithm. To obtain synthetic data, we use the the approach in \cite{gao2013infrared} to add the targets on six real background data. Figs. 12-13 shows the 2D gray distribution of the representative image.
        
\subsection{Validation of the proposed ASTTV-NTLA method}
In this subsection, we validate the robustness of the ASTTV-NTLA method in various scenes.

1) Robustness to single targets scene: First, we test the ASTTV-NTLA method on three real single target scenes. The first and second row of Fig. 4 show the representative images and the related target images, respectively. For better visualization, we adopt red rectangular box to mark the target in the result image. The results in Fig. 4 demonstrate that each target is detected successfully and the background clutters are suppressed perfectly.

2) Robustness to multiple targets scene: In a variety of real scenes, the number of targets of interest is different. Therefore, we test the robustness of ASTTV-NTLA method in multi-targets scenario (actually 3). It is worth noting that we adopt the approach in \cite{gao2013infrared} to synthesize multi-targets scene. The experimental results in the second row of Fig. 5 show that the background noise and clutter are well suppressed.
                
3) Robustness to noisy scene: In real scenes, noise is another crucial factor that impacts the performance of background suppression and target detection. Therefore, we tested the ASTTV-NTLA method in different noise scenes. In our method, ASTTV regularization can not only make full use of spatial-temporal information, but also remove noise and preserve image details. The robustness of ASTTV-NTLA method to noisy scenes is tested on scenes with noise of $\sigma = 15 $ and $\sigma = 25 $. To validate the effectiveness of ASTTV regularization, we compare our model with the model without ASTTV. Fig. 6 (c)-(d) show the experimental results of the model without ASTTV regularization. Fig. 6 (e)-(f) show the experimental results of our model. As can be seen from the second and fourth rows of Fig. 6 (c)-(d), the detection results in non-smooth and non-uniform scenes have noise and background edge residue. It can be seen from Fig. 6 (e)-(f) that the proposed ASTTV-NTLA method can obtain good detection results in non-smooth and non-uniform scenes and remove noise at the same time. Therefore, introducing ASTTV regularization into the model can better solve the problem of small target detection in complex noisy scenes. Meanwhile, to demonstrate the effectiveness of Frobenius norm, we compare the performance of our model and the model without Frobenius norm. From the second row of Fig. 6 (a)-(b), it can be seen that in the noisy scene with $ \sigma=15 $, the result image has only a little noise residue. As can be seen from the fourth row of Fig. 6 (a)-(b), the detection results in heavy noisy scenes have many noise residue. As can be seen from Fig. 6 (e)-(f), the proposed ASTTV-NTLA method can obtain good detection results in heavy noisy scenes. It demonstrates that the introduction of Frobenius norm into the model is helpful to better suppress noise, especially in heavy noisy scenes. In summary, the experimental results in Fig. 6 show that the proposed ASTTV-NTLA method can suppress noise well, which validates the robustness of the proposed ASTTV-NTLA method to different noisy scenes.

\begin{figure*}[htb]
	\vspace{-0.2cm}
	\centering
	\includegraphics[width=18cm]{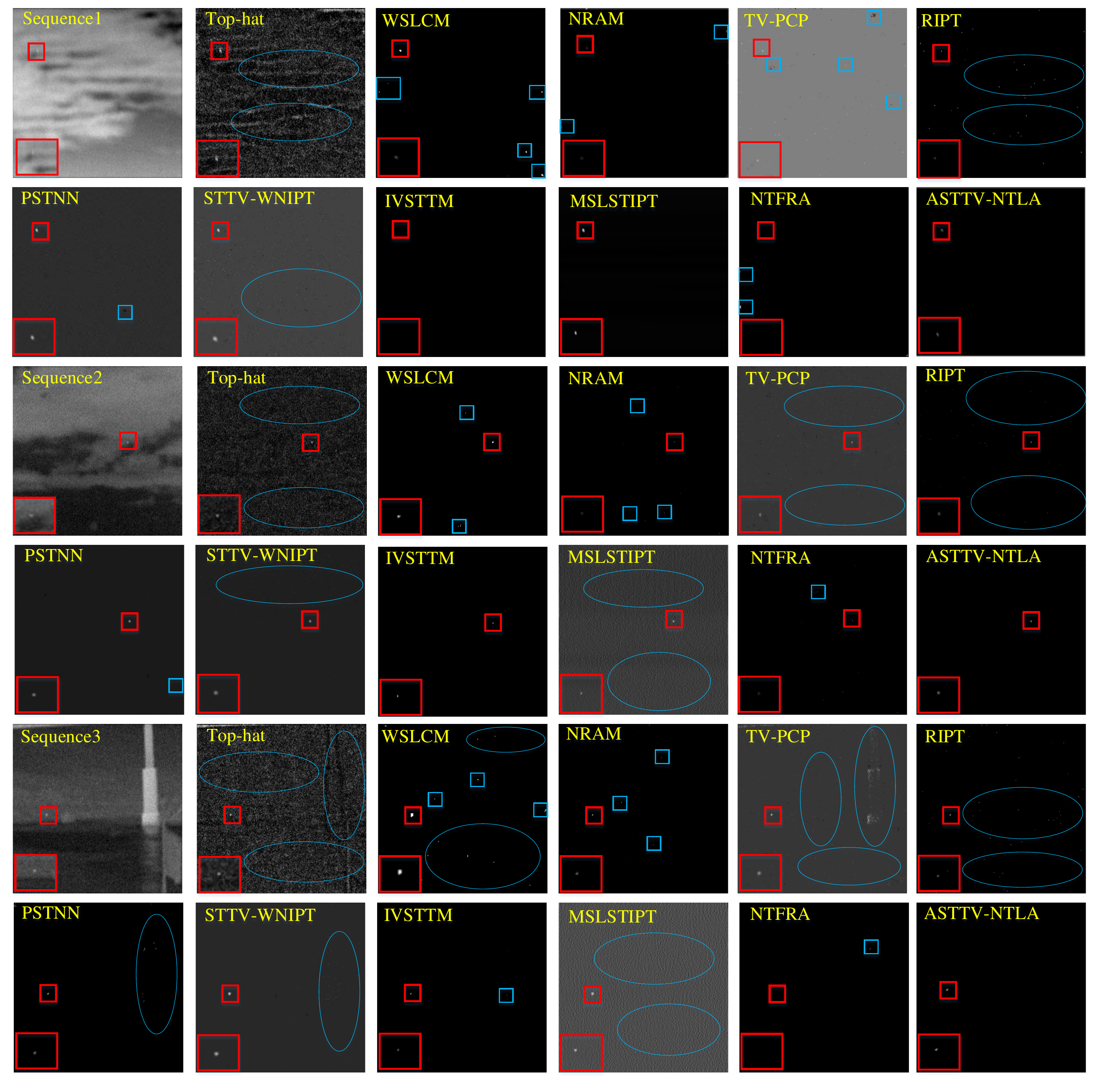}
	\caption{Comparative results achieved by different methods on Sequence 1-3 (without segmentation). The blue ellipses denote noise and background residuals. For more intuitive visualization, the demarcated target area by the red rectangle is zoomed in the left bottom corner.} \label{figDepth}
	\vspace{-0.2cm}
\end{figure*}
 
\subsection{Parameter Analysis}
In this subsection, we analyze the impact of number of frame $ L $, tunning parameter $ H $, and parameter $ \delta $ on the performance of the method.

1) Number of frames: We use the temporal information via introducing ASTTV regularization. Note that the $ L $ is an important parameter. We change $ L $ from 2 to 6 with a step of 1. Fig. 8 shows the relevant ROC curve. The experimental results in Fig. 8 show that $ L = 3 $ has the best performance. 
It is worth noting that if the $ L $ value is set smaller, the detection probability will be decrease. At the same time, Figs. 8 (d) and (f) show that an over-large $ L $ will decrease the detection probability. The main reason is that too large $ L $ leads to the failure of low-rank assumption. In the following experiments, we set  $ L=3 $ , which can maintain a balance between effectiveness and performance, so as to achieve good experimental results.

\begin{figure*}[htb]
	\vspace{-0.2cm}
	\centering	
		\includegraphics[width=18cm]{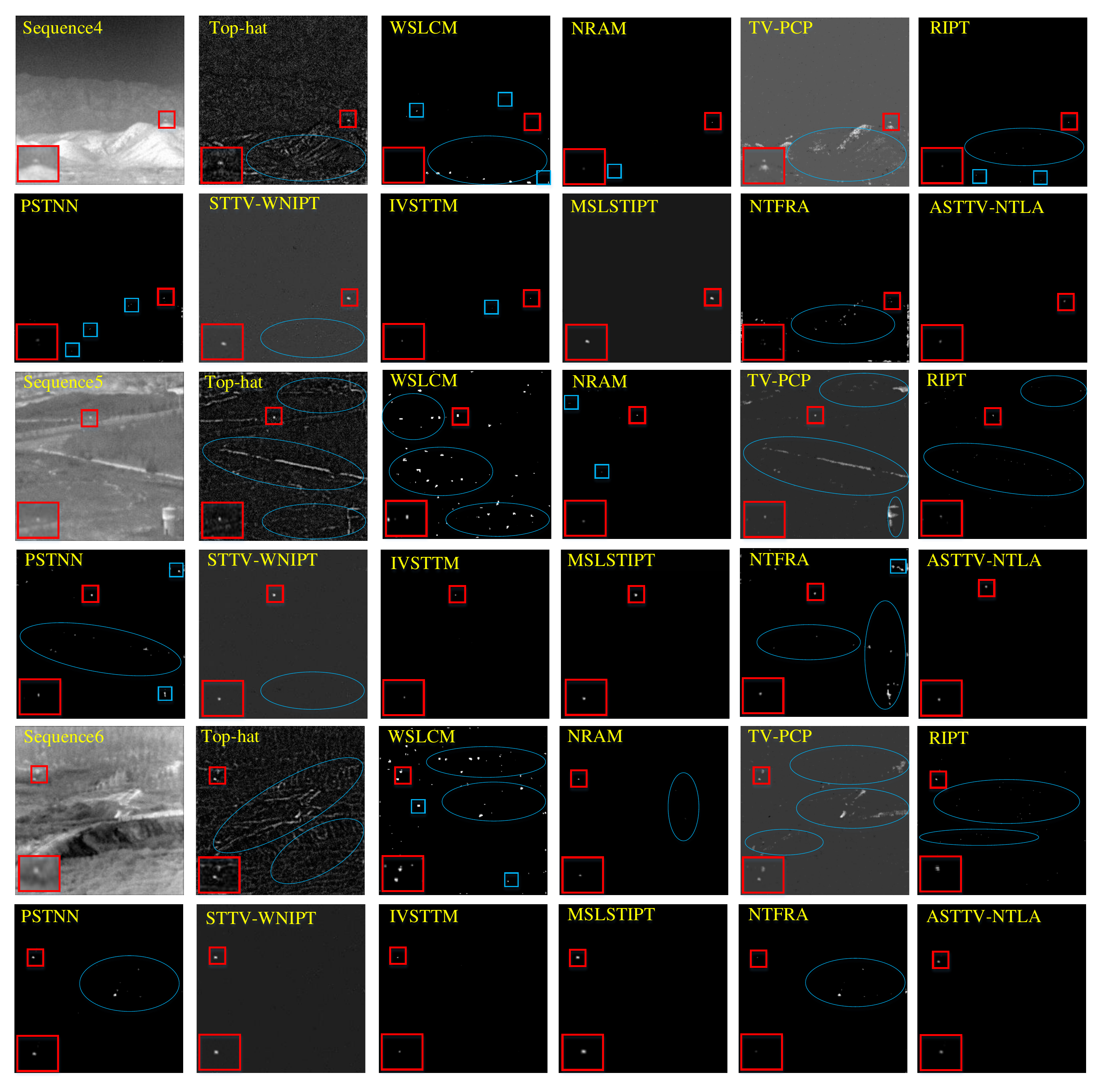}
		\caption{Comparative results achieved by different methods on Sequence 4-6 (without segmentation). The blue ellipses denote noise and background residuals. For more intuitive visualization, the target area demarcated by the red rectangle is zoomed in the left bottom corner.} \label{figDepth}
\vspace{-0.2cm}
\end{figure*}

2) Tunning parameter: $ H $ plays a key role in the optimization of the ASTTV-NTLA model. We change $ H $ from 2 to 10 with a step of 2, and Fig. 10 shows the relevant ROC curve. The ROC curves of $ H = 10 $ in Figs. 10 (a) and (b) demonstrate that an over-large $ H $ will decrease the detection probability. Meanwhile, from the results of $ H = 2 $ and $ H = 4 $ in Figs. 10 (a), (d) and (f), we can conclude that an over-small $ H $ will increase the false alarm rate. Therefore, in the following experiment, we set $ H = 6 $. 

3) Parameter $ \delta $: $\delta $ is a crucial parameter related to temporal information. It indicates that the temporal difference in the ASTTV regularization helps to improve the performance of the ASTTV-NTLA method. We vary $ \delta $ from 0 to 1 with a step of 0.2. Fig. 11 shows the relevant ROC curve. The conclusion that can be drawn from the ROC curves of $ \delta = 0 $ in Fig. 11 is that if there is no temporal information, the detection probability will decrease. $ \delta=1 $ is STTV regularization. As can be seen from Fig. 11, selecting the appropriate $ \delta $ value will get better detection performance. Therefore, in the following experiment, we set  $ \delta=0.5 $. 

\begin{figure*}[htb]
	\vspace{-0.2cm}
	\centering
	
	\subfloat[Sequence 1]{
		\includegraphics[width=5.5cm]{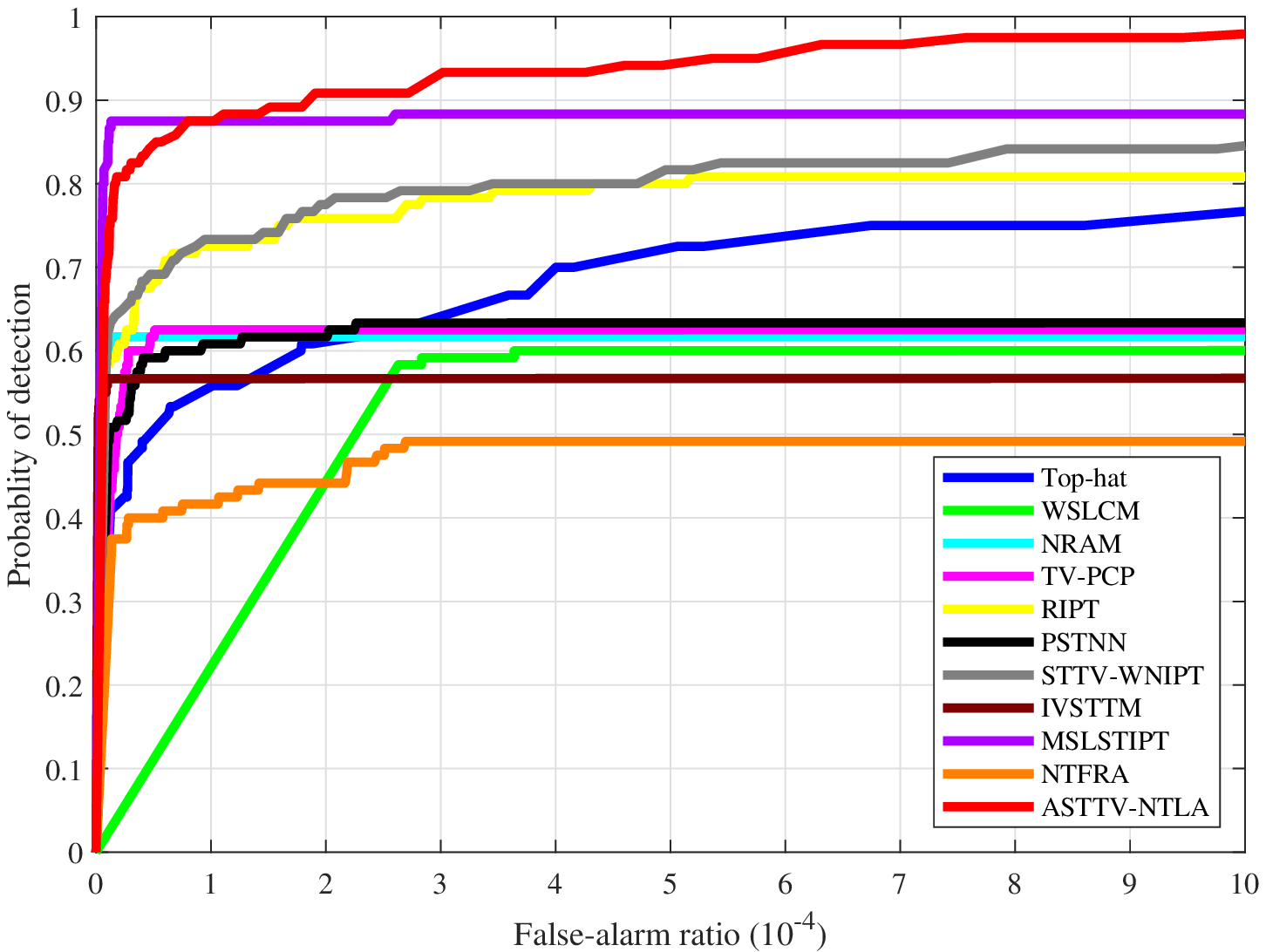}}\subfloat[Sequence 2]{
		\includegraphics[width=5.5cm]{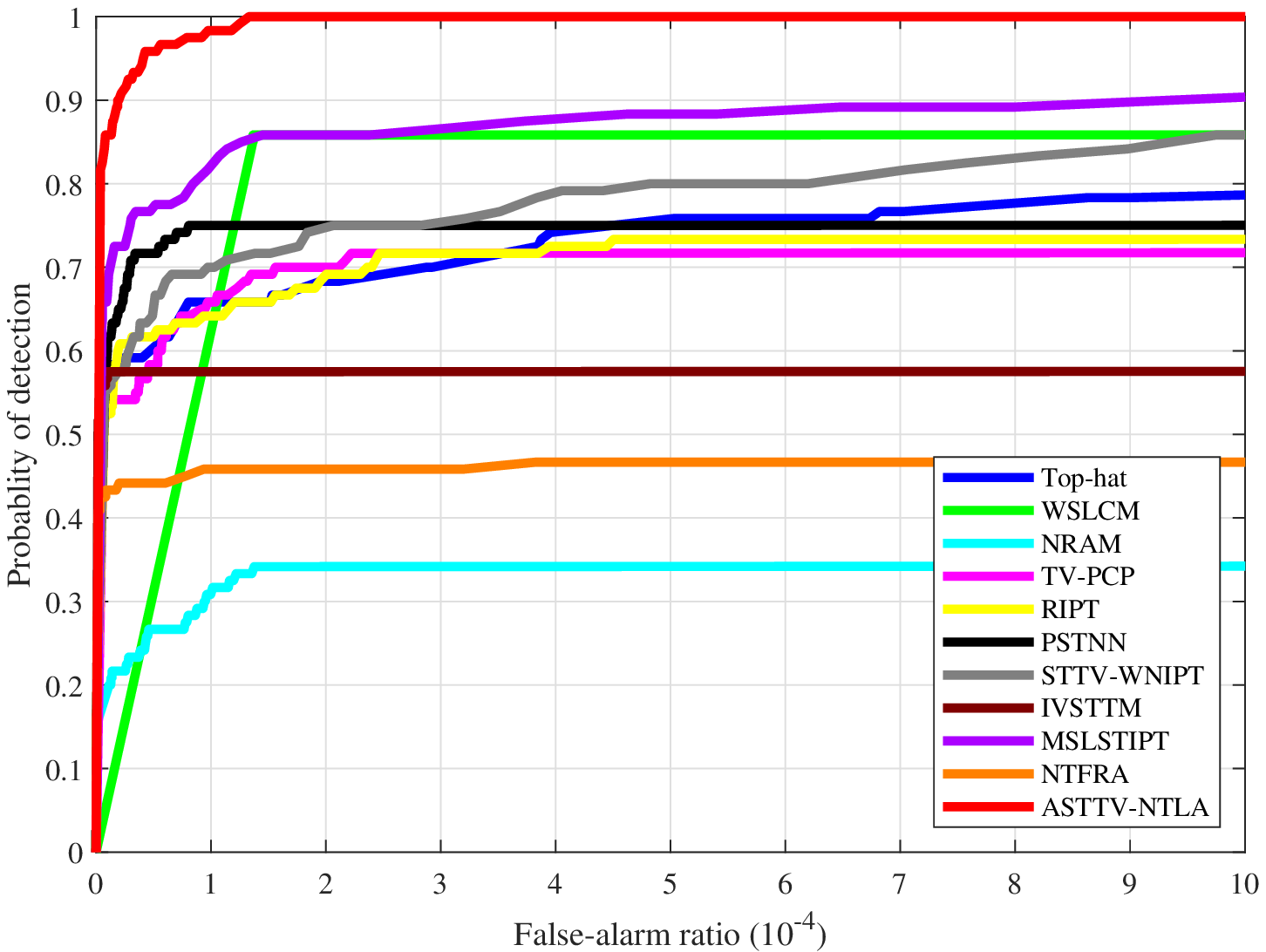}}\subfloat[Sequence 3]{
		\includegraphics[width=5.5cm]{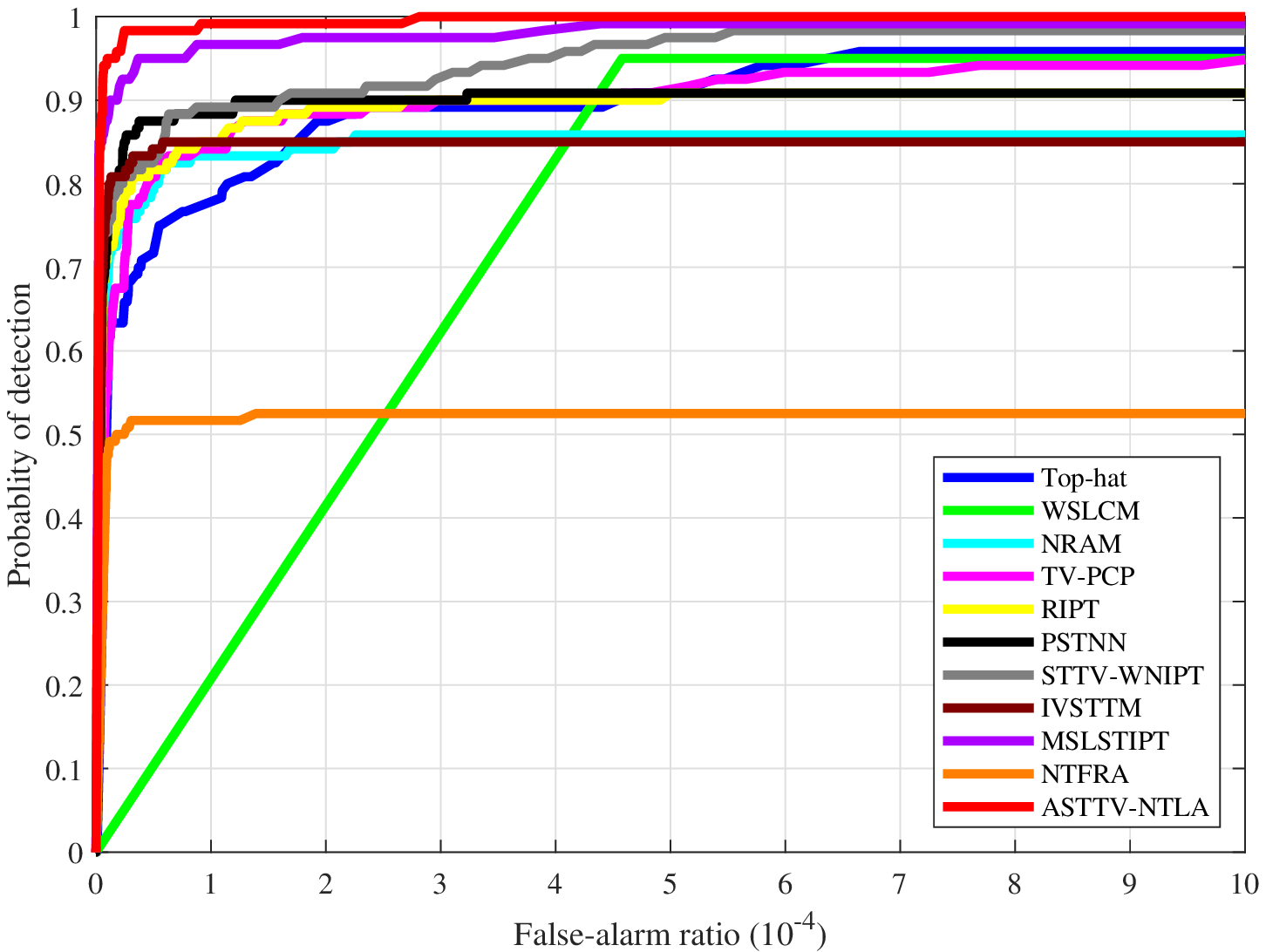}}
	\quad
	\subfloat[Sequence 4]{
		\includegraphics[width=5.5cm]{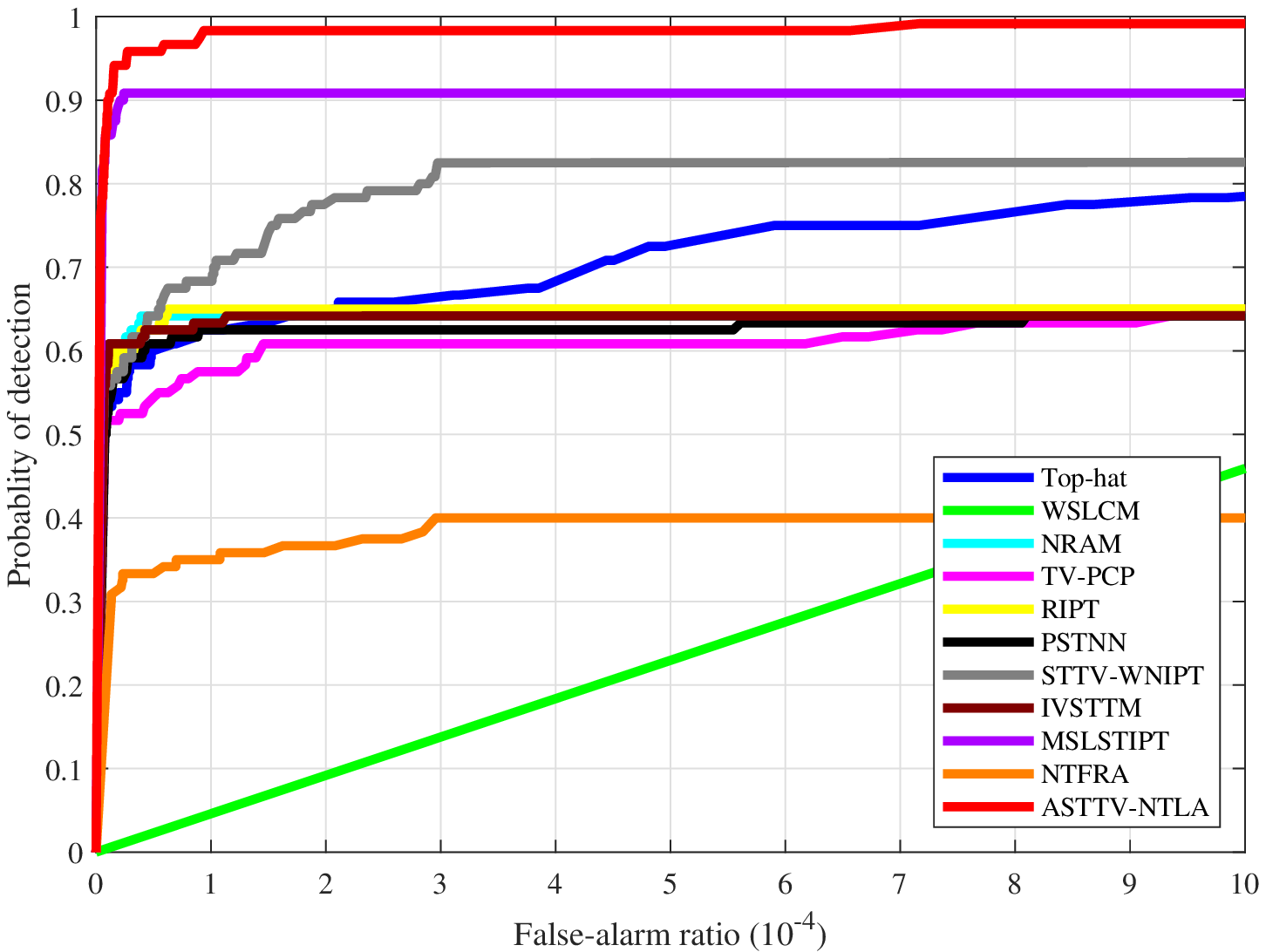}}\subfloat[Sequence 5]{
		\includegraphics[width=5.5cm]{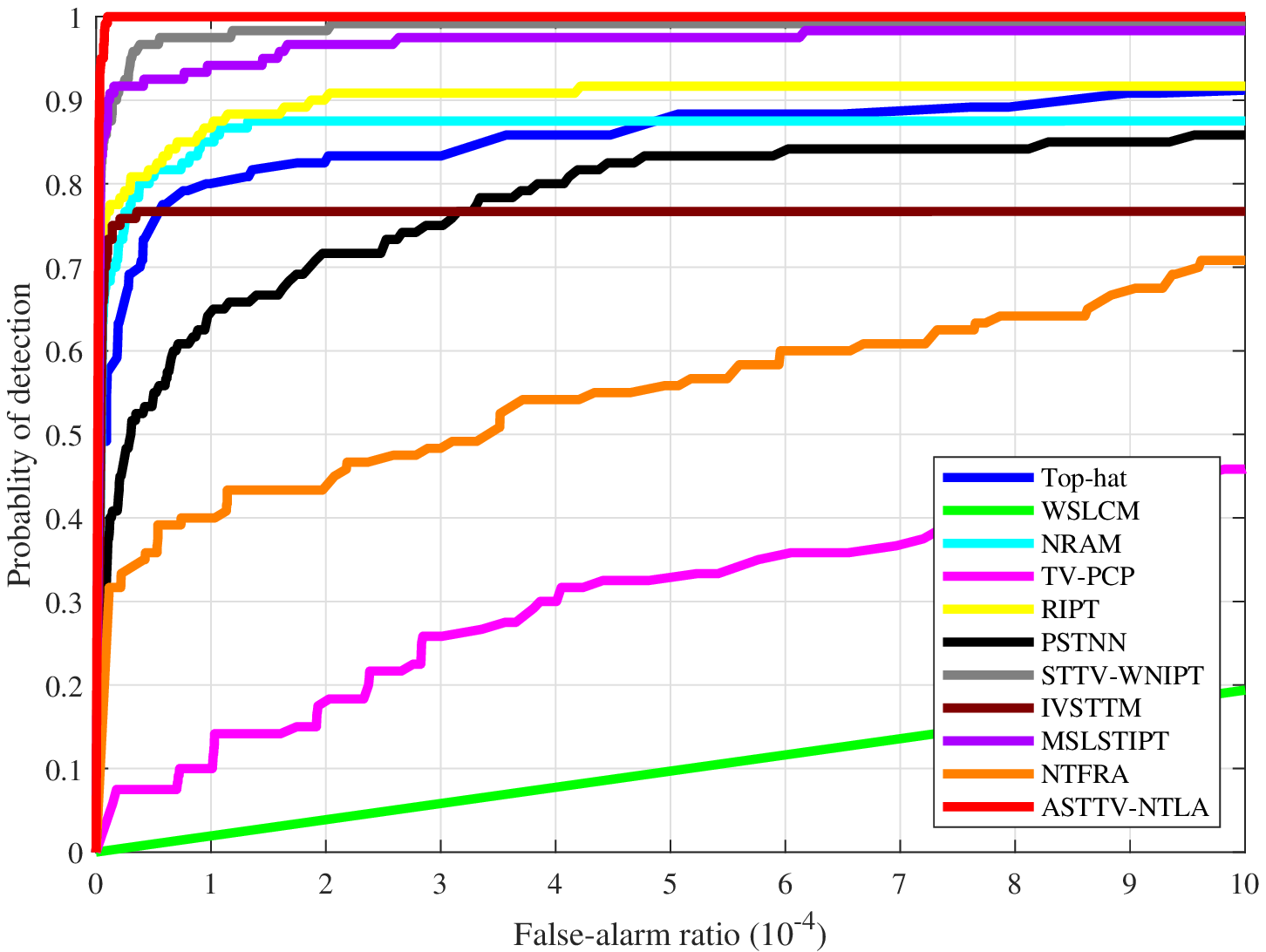}}\subfloat[Sequence 6]{
		\includegraphics[width=5.5cm]{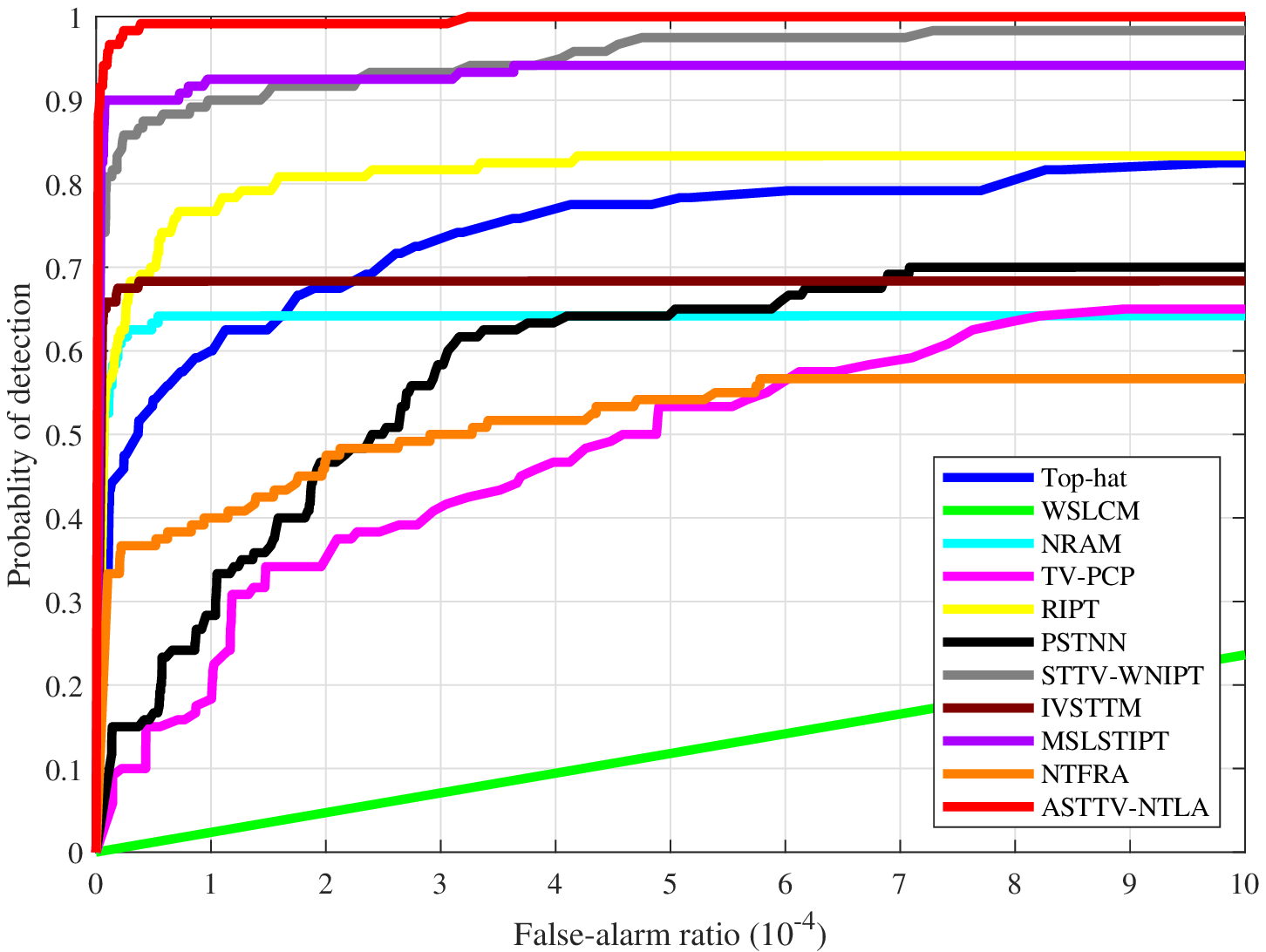}}
	\caption{ROC curves achieved by different methods.}
	\vspace{-0.2cm}
\end{figure*}

	\begin{table*}
		\centering
		\caption{QUANTITATIVE COMPARISON OF DIFFERENT METHODS ON SEQUENCES 1-3.}\label{results_table1}
		\setlength{\tabcolsep}{1mm}{
			\begin{tabular}{rccc|ccc|ccc}
				\midrule[0.75pt]
				\multirow{2}*{Method} & \multicolumn {3}{c}{60$ ^{th} $ frame of Sequence 1}  &  \multicolumn {3}{c}{100$ ^{th} $ frame of Sequence 2}  & \multicolumn {3}{c}{90$ ^{th} $ frame of Sequence 3}  \\
				\cmidrule(r){2-4} \cmidrule(r){5-7} \cmidrule(r){8-10} 
				&LSNRG & BSF&SCRG   &LSNRG & BSF&SCRG   &LSNRG & BSF&SCRG\\\midrule[0.75pt]
				\textbf{Top-hat} \cite{rivest1996detection} 
				& 0.92 & 0.63  & 1.42
				&0.54  &0.99  &0.43
				&0.48 &0.72  &0.24
				\\
				
				\textbf{WSLCM} \cite{han2020infrared}
				&0.90  &0.86  &4.76
				&0.93  &1.80  &4.47
				&0.77  &1.47 &6.18
				
				\\
				\textbf{NRAM} \cite{zhang2018infrared} 
				& 2.70  &7.68  &10.71
				& 1.19 &4.18  &2.95
				&1.28   &4.74 &4.36
				
				\\
				\textbf{TV-PCP} \cite{wang2017infrared1} 
				& 1.41  &4.42   &6.57
				&0.97   &4.70   &3.67
				&1.47  &7.46   &8.27
				
				\\
				\textbf{RIPT}  \cite{dai2017reweighted}
				& 2.55 &6.06  &8.25
				&1.22  &6.21  &3.53
				&3.16 &16.00 &14.73
				
				\\
				\textbf{PSTNN} \cite{zhang2019infrared1}
				&2.74  &7.10  &77.98
				&0.87  &3.76   &4.55 
				&1.52 &4.47   &6.54
				
				\\
				\textbf{STTV-WNIPT} \cite{sun2019infrared}
				&1.62 &2.58 &57.45
				&1.47 &4.27 &4.53
				&1.12 &2.86 &6.38 
				
				\\
				\textbf{IVSTTM} \cite{liu2020small}
				& 2.52 &7.18  &10.01
				&1.60 &8.18 &3.58
				&2.76&19.17 &13.73
				
				\\
				\textbf{MSLSTIPT} \cite{sun2020infrared}
				& 1.86 &2.17  &72.39
				&1.09 &1.44&1.29
				&1.05 &1.52 &2.50 
				
				\\
				\textbf{NTFRA} \cite{kong2021infrared}
				&$-$   &$-$  &$-$
				&1.05 &1.14 &1.13
				&$-$   &$-$ &$-$
				
				\\
				\textbf{ASTTV-NTLA (ours)} 
				&\textbf{3.05}  &\textbf{8.96} &\textbf{97.57}
				& \textbf{3.86} &\textbf{18.16}  &\textbf{14.24}
				& \textbf{3.70} &\textbf{21.30}  &\textbf{17.19}
				
				\\
				\midrule[0.75pt]
			\end{tabular}}
		\end{table*}
	
\subsection{Comparison to state-of-the-art Methods}

To demonstrate the advantages of the ASTTV-NTLA method, we compare it with other ten methods on six different real infrared image scenes. Figs. 12 and 13 show the comparative results of Sequences 1-6. As can be seen from Figs. 12 and 13, Top-hat method produces very rough detection results. Experimental results show that it not only enhances the target, but also enhances the noises and clutters. From the results of sequence 4 and sequence 5 in Fig. 13, it can be seen that the residuals of the reflective mountains and roads still remain in the target image. The main reason is that the size of the Top-hat filter is not suitable for the scenes with strong reflection clutter. As a top-performing HVS method, WSLCM can detect the target more accurately in simple background, but there are still clutter or missed detection in complex background. Compared with the BS and HVS method, matrix-based LRSD methods have less background residual or missed clutter in complex background, such as NRAM, TV-PCP methods. From the highlight scene Sequence 4 and the complex ground scene Sequences 5-6, it can be seen that NRAM method still has a little residual and clutter, but TV-PCP method achieves poor performance on these complex scenes. To handle these problems, tensor-based method is proposed. The results in Figs. 12 and 13 show that RIPT method is more effective in clutter suppression than matrix-based methods. Therefore, many improved methods are proposed, such as PSTNN and NTFRA methods. As can be seen from Fig. 12, these tensor-based methods can suppress clutter in complex background, but some non-target pixels still remain in their target image. The main reason is that there are many clutter in a single-frame complex image background, such as highlight background and ground background, which may seriously affect the detection of real targets. The information in a single-frame image is not enough to distinguish small targets. Based on this, many scholars have introduced temporal information and use spatial-temporal information to solve small target detection. As can be seen from Fig. 12-13, compared with those tensor methods that only use spatial information, the spatial-temporal information method (STTV-WNIPT, IVSTTM and MSLSTIPT) can better detect the target and suppress the background. It can be seen from Sequence 3 and Sequence 4 that IVSTTM method has background residue on the target image. The main reason is that NNM treats all singular values equally, which will lead to over-shrinkage problem. To solve this problem, the STTV-WNIPT method introduces WNNM and STTV regularization. It can be seen from Sequence 4 and Sequence 5 that there is background residue in the target image obtained by STTV-WNIPT method, which indicates that WNNM can only alleviate over-shrinkage problem. Further, MSLSTIPT method introduces WSNM to obtain more accurate background estimation. However, it can be seen from Sequence 2 and Sequence 3 that the background suppression effect of MSLSTIPT method is not good. In contrast, ASTTV-NTLA method can detect targets accurately under the premise of better suppression of background and noise. The results demonstrate the advantages of the ASTTV and non-convex tenor low-rank approximately property. Note that the dataset contains a variety of scenes, which shows the superiority and robustness of the ASTTV-NTLA method.

\begin{figure*}[htb]
	\vspace{-0.2cm}
	\centering
	
	\subfloat[Sequence 1]{
		\includegraphics[width=5.5cm]{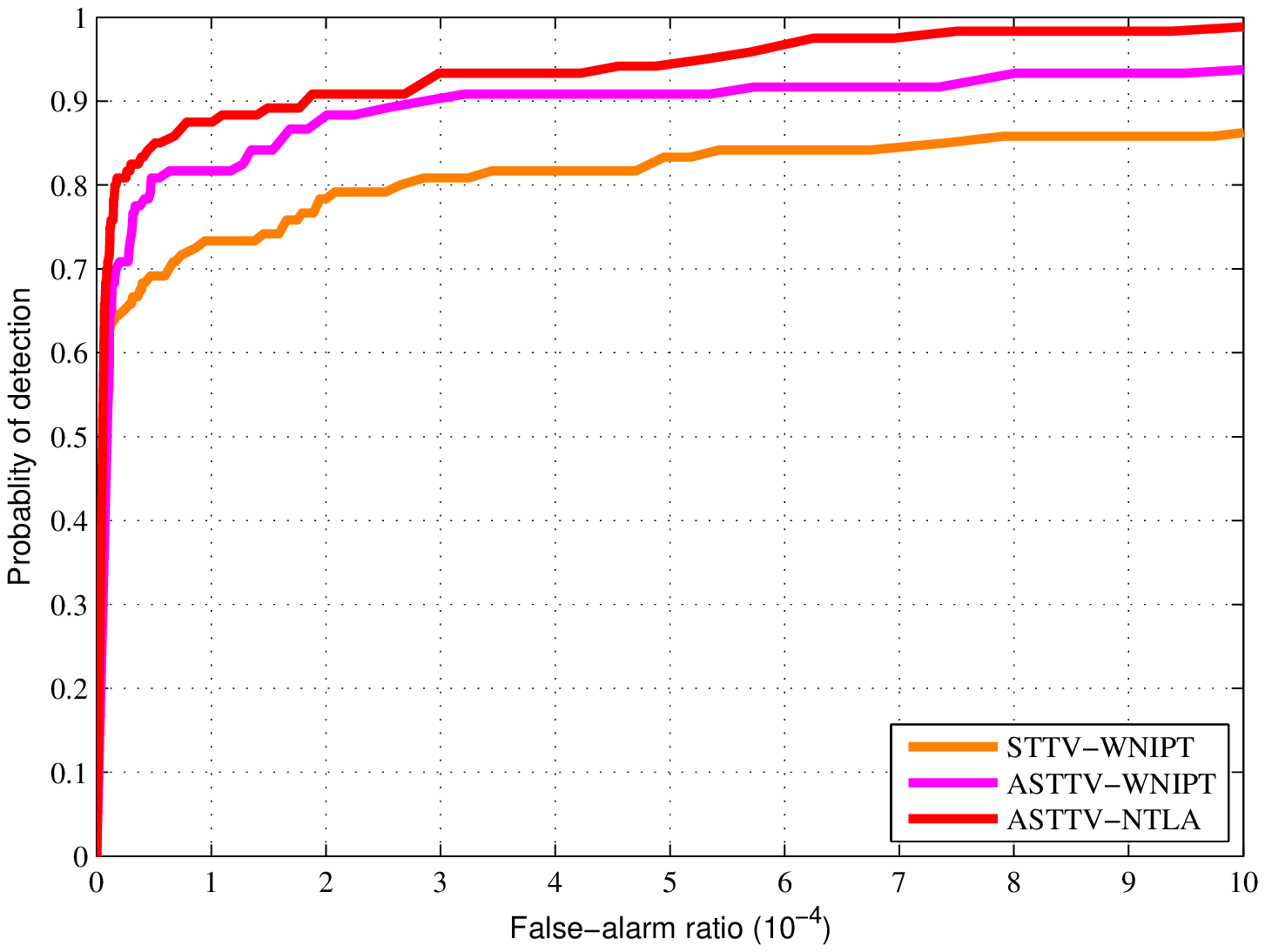}}\subfloat[Sequence 2]{
		\includegraphics[width=5.5cm]{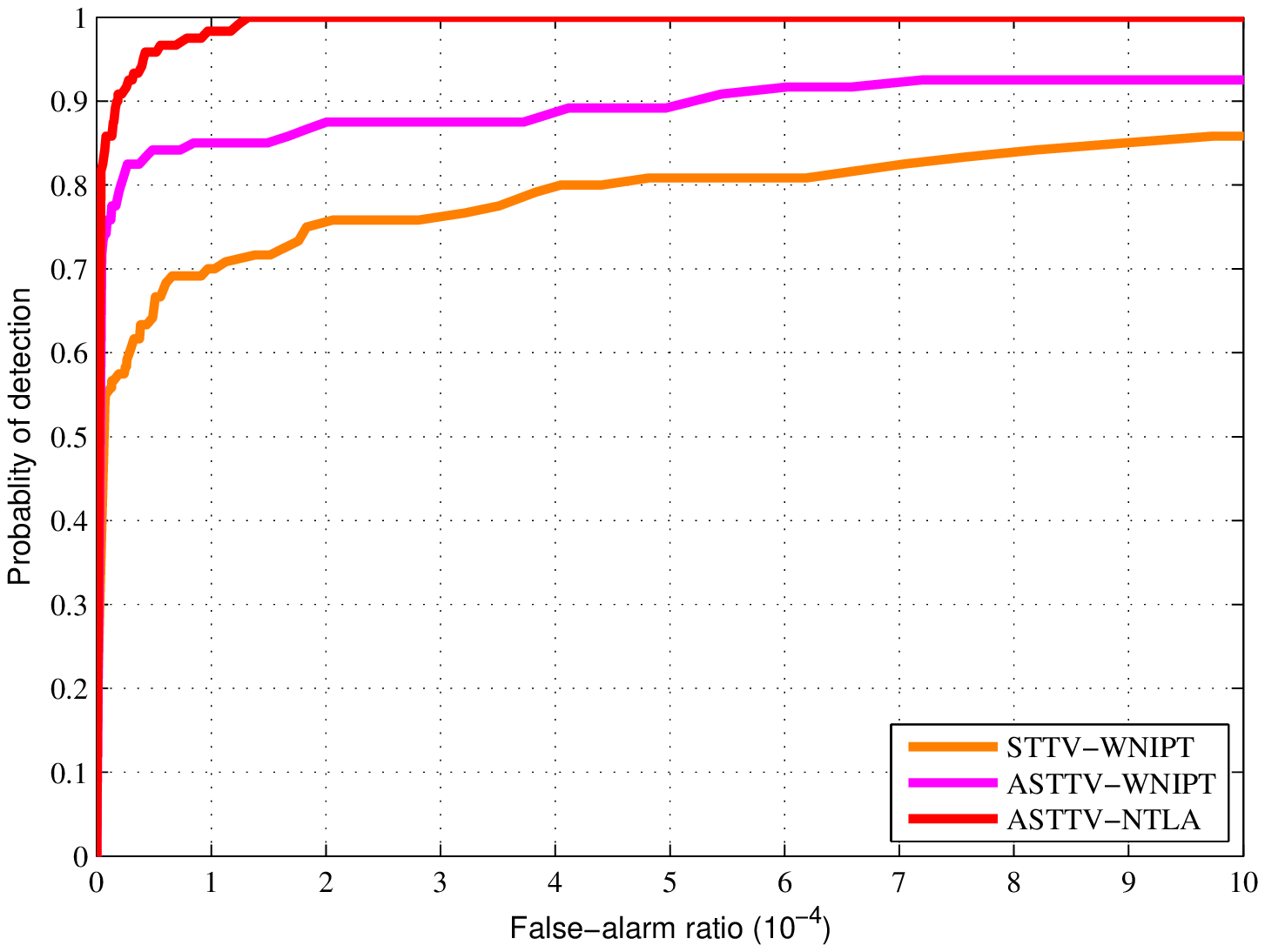}}\subfloat[Sequence 3]{
		\includegraphics[width=5.5cm]{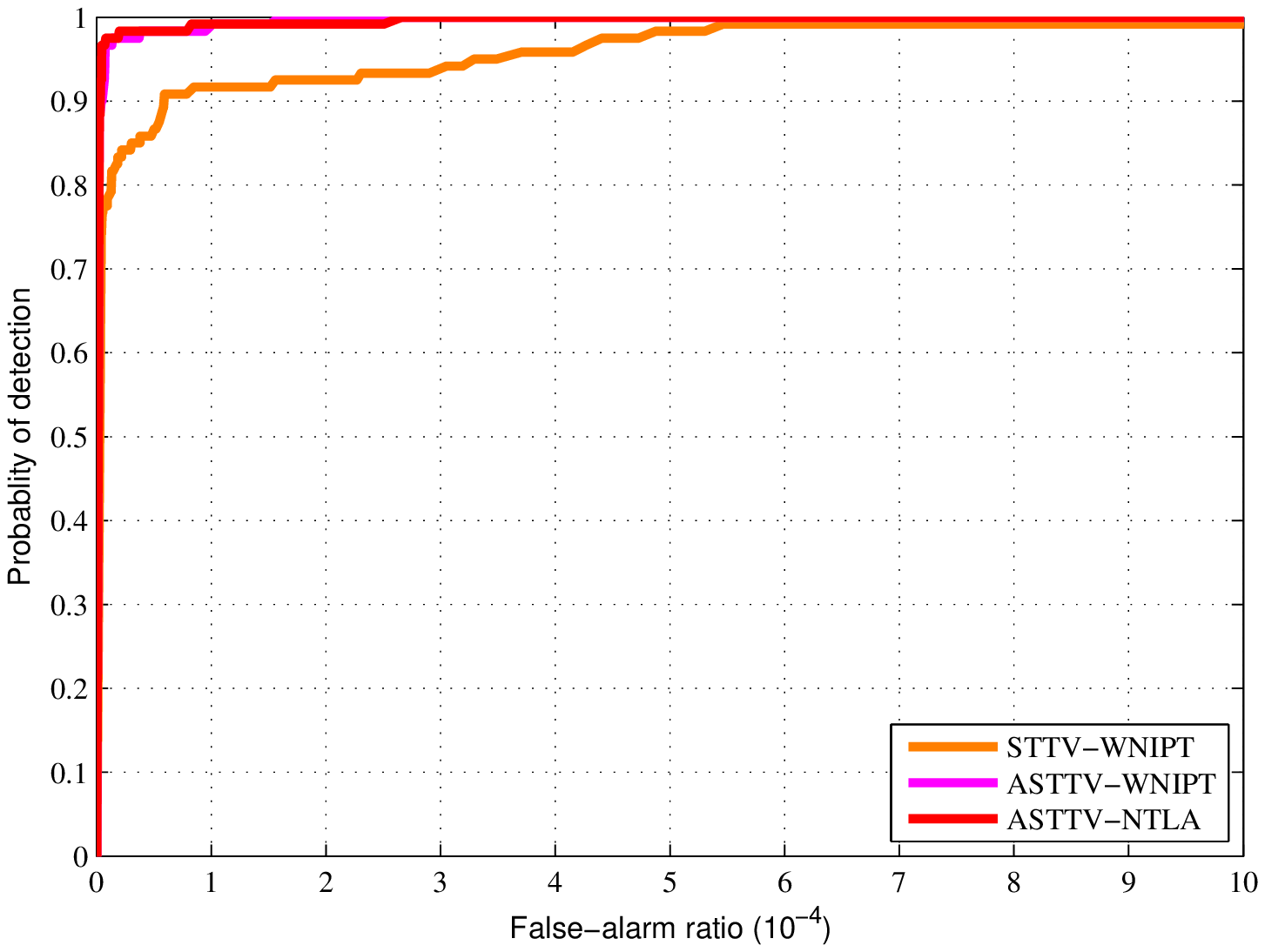}}
	\quad
	\subfloat[Sequence 4]{
		\includegraphics[width=5.5cm]{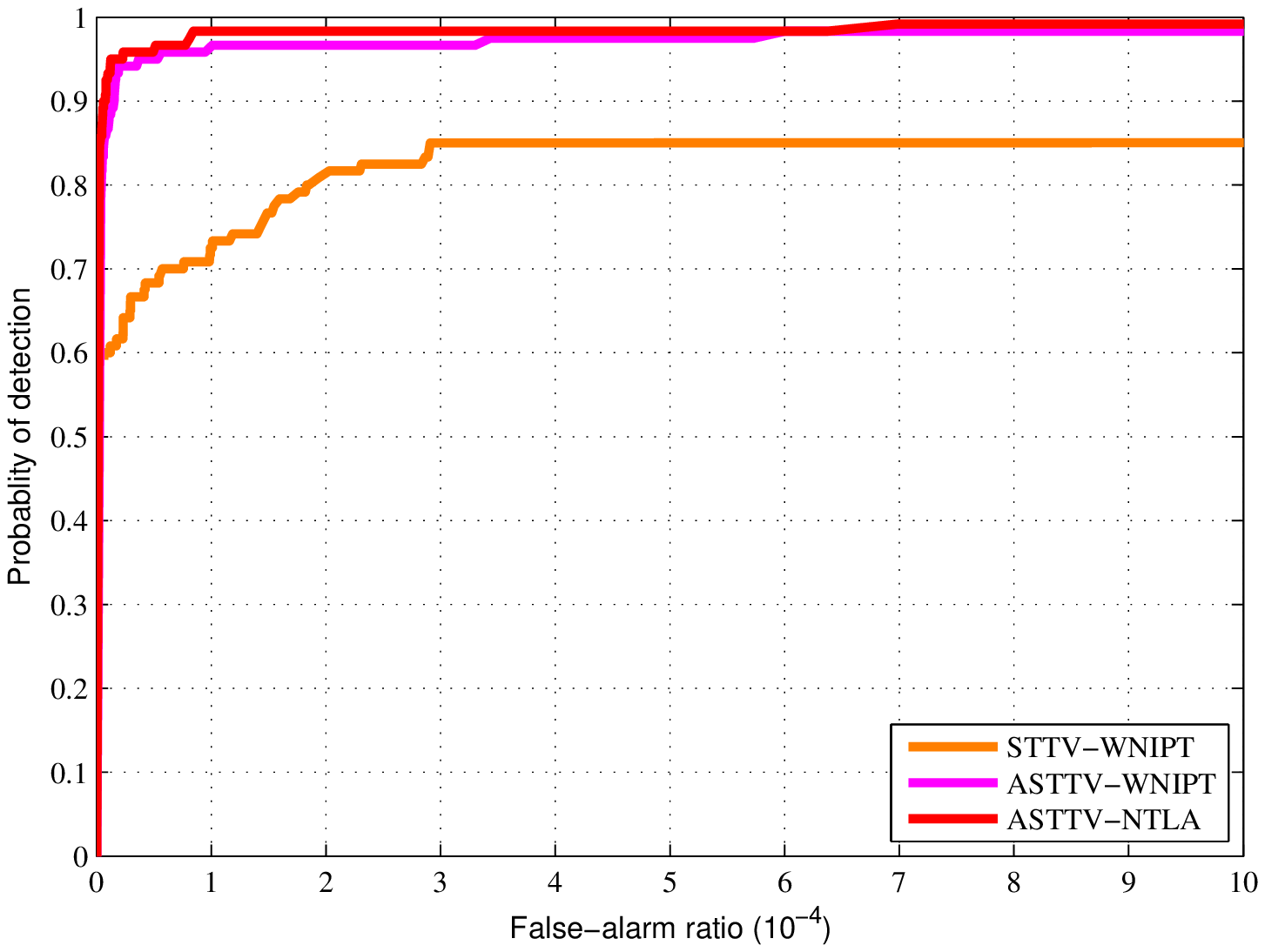}}\subfloat[Sequence 5]{
		\includegraphics[width=5.5cm]{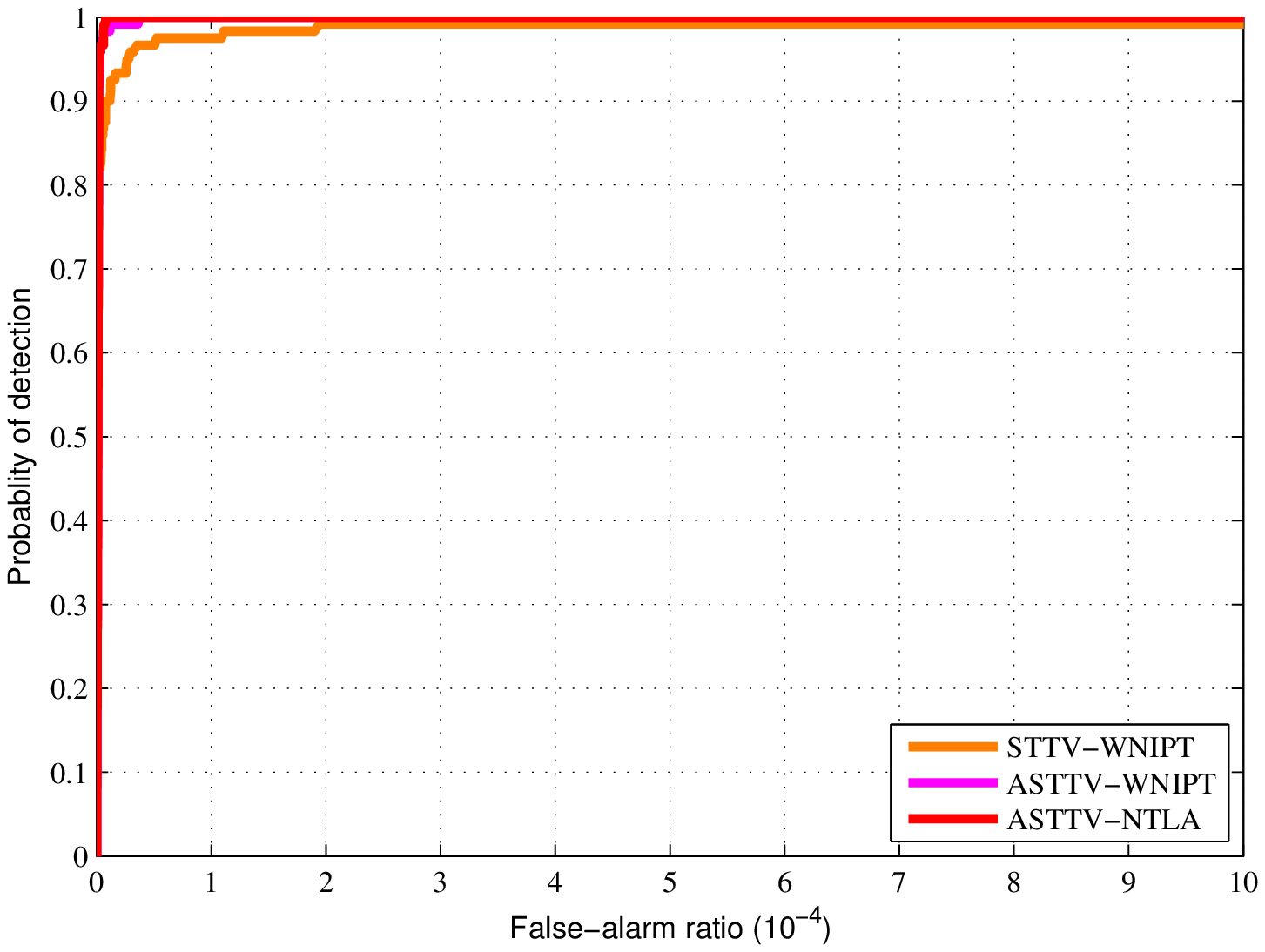}}\subfloat[Sequence 6]{
		\includegraphics[width=5.5cm]{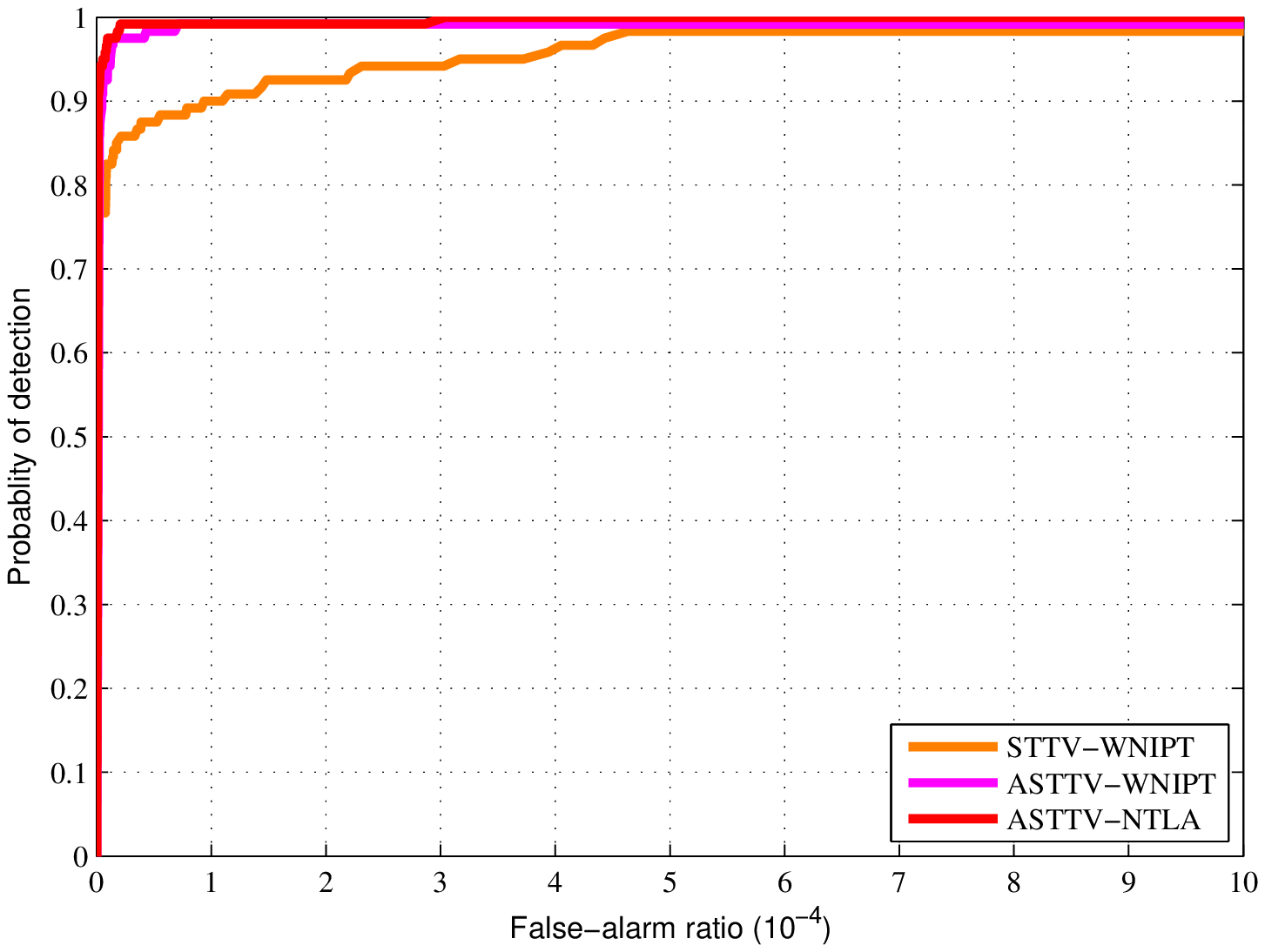}}
	\caption{ The ROC curves achieved by different variants of our method.}
	\vspace{-0.2cm}
\end{figure*}

		\begin{table*}
			\centering
			\caption{QUANTITATIVE COMPARISON OF DIFFERENT METHODS ON SEQUENCES 4-6.}\label{results_table1}
			\setlength{\tabcolsep}{1mm}{
				\begin{tabular}{rccc|ccc|ccc}
					\midrule[0.75pt]
					\multirow{2}*{Method} & \multicolumn {3}{c}{50$ ^{th} $ frame of Sequence 4}  &  \multicolumn {3}{c}{10$ ^{th} $ frame of Sequence 5}  & \multicolumn {3}{c}{90$ ^{th} $ frame of Sequence 6}  \\
					\cmidrule(r){2-4} \cmidrule(r){5-7} \cmidrule(r){8-10} 
					&LSNRG & BSF&SCRG   &LSNRG & BSF&SCRG   &LSNRG & BSF&SCRG\\\midrule[0.75pt]
					\textbf{Top-hat} \cite{rivest1996detection} 
					& 0.51 & 0.72  & 0.42
					&0.76  &1.76  &0.50
					&0.45 &1.88  &0.37
					\\
					
					\textbf{WSLCM} \cite{han2020infrared}
				    &$-$  &$-$  &$-$
					&0.12  &4.47  &24.87
					&0.77  &2.29 &5.84
					
					\\
					\textbf{NRAM} \cite{zhang2018infrared} 
					& 9.56  &23.74  &32.79
					& 1.33 &7.08  &21.12
					&1.34  &9.31 &9.54
					
					\\
					\textbf{TV-PCP} \cite{wang2017infrared1} 
					& 1.22  &3.98   &6.98
					&1.16  &9.23   &19.51
					&1.14  &6.70   &8.61
					
					\\
					\textbf{RIPT}  \cite{dai2017reweighted}
					& 1.27 &3.28  &5.74
					&1.40  &7.90  &25.19
					&1.53 &10.25 &8.70
					
					\\
					\textbf{PSTNN} \cite{zhang2019infrared1}
					&1.41  &3.86  &8.49
					&1.25  &4.14   &22.34
					&1.04 &4.14   &6.92
					
					\\
					\textbf{STTV-WNIPT} \cite{sun2019infrared}
					&1.17 &2.42 &6.63
					&1.34 &4.68 &21.15
					&1.03 &4.00 &7.42 
					
					\\
					\textbf{IVSTTM} \cite{liu2020small}
					& 1.38 &4.09  &7.40
					&0.78 &6.32 &6.11
					&1.69&12.23 &14.00
					
					\\
					\textbf{MSLSTIPT} \cite{sun2020infrared}
					& 1.20 &2.07  &6.55
					&1.40&4.23&23.17
					&1.00 &3.46 &7.21
					
					\\
					\textbf{NTFRA} \cite{kong2021infrared}
					&2.84  &24.69  &12.15
					&1.10 &7.01 &22.28
					&1.19 &5.94 &8.87
					
					\\
					\textbf{ASTTV-NTLA (ours)} 
					&\textbf{14.32}  &\textbf{33.84} &\textbf{253.34}
					& \textbf{26.70} &\textbf{110.82}  &\textbf{97.21}
					& \textbf{5.12} &\textbf{53.07}  &\textbf{34.04}
					
					\\
					\midrule[0.75pt]
				\end{tabular}}
			\end{table*}

In this paper, we adopt LSNRG, BSF and SCRG as metric for quantitative evaluation, which are widely used by the exsiting methods. Table III and Table IV show the results. The best results of the test method are highlighted in bold. The results in the Table show that ASTTV-NTLA method obtains the best results on the indicator values. It shows that the ASTTV-NTLA method not only can effectively suppress noise and clutters, but also can better highlight the targets. It is worth noting that we do not show the indicators values of NTFRA on the 60$ ^{th} $ frame of Sequence 1 and the 90$ ^{th} $ frame of Sequence 3 and WSLCM on the 50$ ^{th} $ frame of Sequence 4. The main reason is that the target is lost in these scenes, so it is meaningless to calculate the indicator value. To deal with the above problem, we use the approach in \cite{gao2018infrared} and introduce the CG metric to ensure the ability of expanding the gray value difference between the background and the target. Table V shows the average CG value of the entire sequence. The results in Table V show that ASTTV-NTLA method obtains the highest average CG value on all test Sequences. This metric demonstrate that the ASTTV-NTLA method has satisfactory background suppression capacity.

\begin{table*}[t]
	\vspace{-.05in}
	\scriptsize
	\centering
	\renewcommand\arraystretch{1.1}
	\caption{Average CG values achieved by different methods on sequences 1-6}\label{ablation1}
	\begin{tabular}{lccccccc}
		\hline
		Methods & Sequence 1 &  Sequence 2 &  Sequence 3 & Sequence 4 &  Sequence 5 &  Sequence 6  \\[0.05cm]
		\hline
		\textbf{Top-hat} \cite{rivest1996detection}    & 2.25 & 1.41  & 1.12  & 3.16 & 1.29  & 1.15\\
		\textbf{WSLCM} \cite{han2020infrared}    & 13.03 & 3.16  & 4.20  & 5.85 & 2.36  & 3.46\\
		\textbf{NRAM} \cite{zhang2018infrared}  & 13.16 & 1.03  & 1.80  & 6.38 & 1.57  & 1.92\\
		\textbf{TV-PCP} \cite{wang2017infrared1}  & 10.55 & 1.70  & 1.11  & 7.75 & 1.82  & 1.66\\
		\textbf{RIPT} \cite{dai2017reweighted}   & 15.43 & 1.69  & 1.92 & 6.72 & 1.58  & 3.11  \\
		\textbf{PSTNN} \cite{zhang2019infrared1}   & 19.22 & 1.80  & 5.46 & 8.20 & 1.16  & 2.99  \\
		\textbf{STTV-WNIPT } \cite{sun2019infrared} & 22.24 & 1.94 & 7.23  & 8.74 & 1.18  & 5.30 \\
		\textbf{IVSTTM} \cite{liu2020small}   &13.29 &2.44 & 1.81  & 7.90 & 1.70  & 1.72  \\
		\textbf{MSLSTIPT} \cite{sun2020infrared}   & 33.33 & 2.89  & 2.64  & 22.38 & 2.14 & 4.18 \\
		\textbf{NTFRA} \cite{kong2021infrared} & 1.38 & 1.02 & 1.52 & 2.56 & 1.42 & 1.64 \\
		\textbf{ASTTV-NTLA (ours)}    & $\mathbf{35.54}$ &  $\mathbf{3.57}$ & $\mathbf{5.64}$  & $\mathbf{27.49}$  & $\mathbf{5.88}$   & $\mathbf{8.30}$ \\

		\hline
	\end{tabular}
\end{table*}

\begin{table*}[t]
	\vspace{-.05in}
	\scriptsize
	\centering
	\renewcommand\arraystretch{1.1}
	\caption{Running Time of different methods}\label{ablation1}
	\begin{tabular}{lccccccc}
		\hline
		Methods & Sequence 1 &  Sequence 2 &  Sequence 3 & Sequence 4 &  Sequence 5 &  Sequence 6  \\[0.05cm]
		\hline
		\textbf{Top-hat} \cite{rivest1996detection}   & 32.52s & 40.46s  & 33.79s  & 32.73s & 32.42s  & 31.34s \\
		\textbf{WSLCM} \cite{han2020infrared}    & 1087.39s & 1701.05s  & 1411.60s  & 1450s & 1451.99s  & 1400.41s \\
		\textbf{NRAM} \cite{zhang2018infrared}  & 80.07s & 223.71s  & 185.28s  & 153.01s & 190.28s  & 157.94s\\
		\textbf{TV-PCP} \cite{wang2017infrared1}  & 7976.50s & 13314.75s  & 11144.09s  & 11339.91s & 11164.87s  & 11233.05s\\
		\textbf{RIPT} \cite{dai2017reweighted}   & 91.32s & 153.93s  & 138.49s  & 112.31s & 168.66s  & 131.04s  \\
		\textbf{PSTNN} \cite{zhang2019infrared1}    & 29.97s & 38.95s  & 34.86s  & 46.16s & 52.04s  & 45.42s  \\
		\textbf{STTV-WNIPT } \cite{sun2019infrared}  & 174.70s & 293.81s & 235.94s  & 243.41s & 245.79s  & 245.77s  \\
		\textbf{IVSTTM} \cite{liu2020small}   & 80.37s & 117.61s  & 106.35s  & 139.16s   & 102.02s  & 115.34s \\
		\textbf{MSLSTIPT} \cite{sun2020infrared}    & 176.13s  & 275.39s & 208.98s & 225.32s &181.82s   & 181.63s  \\
		\textbf{NTFRA} \cite{kong2021infrared}  & 129.19s   & 151.97s   & 156.08s   & 351.07s  & 320.56s  & 281.38s  \\
		\textbf{ASTTV-NTLA (ours)}    & 206.64s & 377.35s & 287.89s  & 284.78s  & 280.17s   & 283.33s \\

		\hline
	\end{tabular}
\end{table*}

Fig. 14 shows the ROC curves of all comparison methods in this paper. The results in Fig. 14 show that ASTTV-NTLA method has best performance. Top-hat method is greatly affected by bright clutters and noise. WSLCM and TLLCM method achieve better detection performance on simple background scences such as Sequence 1-3, and worse detection performance on highlighted and complex ground background scences such as Sequence 4-6. The ROC curves in Fig. 14 (d)-(f) show that the performance of NRAM method and TV-PCP method is slightly better than WSLCM and Top-hat method in complex scenes. However, it can be seen from Fig. 14 (b) and Figs. 14 (d)-(f) that the detection results of NRAM and TV-PCP methods are relatively poor. Therefore, tensor-based method (RIPT, PSTNN, NTFRA) are proposed. As can be seen from (d)-(e) in Fig. 14, the detection results of PSTNN in complex scenes are relatively poor. The main reason is that PSTNN method assigns same weights for all singular values. As can be seen from Fig. 14 (d)-(f), the detection results of tensor-based method in complex scenes are alleviated to a certain extent. Compared with the methods (NRAM, TV-PCP, RIPT, PSTNN, NTFRA) that only uses spatial information, the STTV-WNIPT, IVSTTM and MSLSTIPT methods achieve better performance in various scenes because they incorporate spatial-temporal information. It indicatess the effectiveness of spatial-temporal information. It can be seen from Fig. 14 that among all the test methods, the Pd of the ASTTV-NTLA method can reach 1 the fastest. This shows the necessity of considering temporal correlation and the effectiveness of integrating NTLA regularization and ASTTV regularization. In summary, all the above evaluation metrics show that ASTTV-NTLA method can achieve satisfactory robust performance in different real scenes, particularly for complex ground scenes and highly heterogeneous scenes.

\subsection{Ablation Experiments}
To show the advantages of the ASTTV-NTLA method that integrating the ASTTV regularization and NTLA regularization, we evaluate the contributions of the ASTTV and NTLA, respectively. The ASTTV-NTLA method mainly composes of three parts: the spatial-temporal tensor structure, ASTTV regularization and NTLA regularization. Therefore, the performance of the three versions of ASTTV-NTLA method is compared on Sequences 1-6, including: 1) Adopting the STTV regularization constraint weighted IPT model (STTV-WNIPT); 2) Adopting the ASTTV regularization constraint the weighted IPT model (ASTTV-WNIPT); 3) Adopting the ASTTV regularization constraint the NTLA regularization (ASTTV-NTLA). Fig. 15 shows the ROC curves of STTV-WNIPT, ASTTV-WNIPT and ASTTV-NTLA methods on Sequences 1-6. Moreover, the performance of ASTTV-WNIPT is better than that of STTV-WNIPT, which demonstrates that ASTTV regularzation can improve the detection ability of the model to a certain extent by using different smoothness strength for temporal TV and spatial TV. The conclusion drawn from Fig. 15 is that the detection results of  ASTTV-NTLA is better than that of ASTTV-WNIPT. Compared with WNIPT regularization, NTLA regularization is a better substitute for tensor rank, which can obtain more accurate background estimation and further improve the ability of target detection. In summary, the above experimental results demonstrate that integrating Laplace norm with ASTTV regularization can achieve better target detection performance.

\subsection{Running time}
In addition to the above four evaluation metrics, running time is also a crucial factor. However, it is difficult to balance these two factors. Therefore, we compare the efficiency of the comparison method on Sequences 1-6. Table VI shows the running time of all comparison methods in various scenes. Top-hat method is the fastest among all comparison methods, but its detection performance is relatively rough. In constrast, the speed of LRSD methods are relatively slow. Because the matrix-based LRSD method needs many SVD operations. Among them, TV-PCP needs relatively long running time. The main reason is that they adopt accelerated proximal gradient (APG) method for optimization. The methods optimized by ADMM are more efficient, such as  NRAM, RIPT, PSTNN, STTV-WNIPT, IVSTTM, MSLSTIPT, NTFRA and the proposed method. Among these methods, the speed of PSTNN is only slower than Top-hat. The main reason is that with the help of the additional stop criterion and t-SVD, the computing time and complexity of the algorithm are greatly reduced. The three experimental results based on the TV regularization method in Table VI show that the running time of the ASTTV-NTLA method is slightly slower than the STTV-WNIPT method, but much faster than TV-PCP method. Considering that the ASTTV-NTLA method achieves good performance in various complex scenes, the running time can be slightly sacrificed. As can be seen from Table VI, our method is faster than some matrix-based SVD methods (e.g., TV-PCP ), but slower than some t-SVD based methods (e.g., PSTNN, IVSTTM). That is because, ASTTV regularization has high computational complexity. In the future work, we can refer to the strategies in recent works  \cite{pang2020infrared, pang2021facet, zhang2019infrared1} to enhance the real-time performance of our method. 
\section{Conclusion}
         
To improve the capacity of target detection and background suppression in complex noise and strong clutters scenes, an asymmetric spatial-temporal total variation regularized non-convex low-rank tensor approximation method is proposed. The NTLA regularization is used to adaptively assign different weights to all singular values, which helps reconstruct the background image more accurately. In addition, ASTTV regularization can fully utilize the structure prior information to detect targets. Furthermore, ASTTV regularization can effectively exploit spatial-temporal information to suppress the background noise and detect the targets in non-uniform and non-smooth scenes. Extensive experimantal results show the promising detection performance of the ASTTV-NTLA method. 

However, in our model, we only consider white Gaussian noise, which deviates from the complicated real-world noise model. Therefore, the performance of the ASTTV-NTLA method in complex noisy scenes is limited. Recently, deep convolutional neural networks (CNNs) have achieved impressive success in image denoising and can handle more realistic noise. Therefore, in the future, we will consider to use a learning-based CNN denoiser to handle the denoising subproblem for better detection performance.

\section{acknowledgments}
This work was supported in part by the National Natural Science Foundation of China under
Grant 61972435, Grant 61401474, and Grant 61921001.

\bibliographystyle{IEEEtran}
\bibliography{SPL}

\begin{IEEEbiography}[{\includegraphics[width=1in,height=1.25in,clip,keepaspectratio]{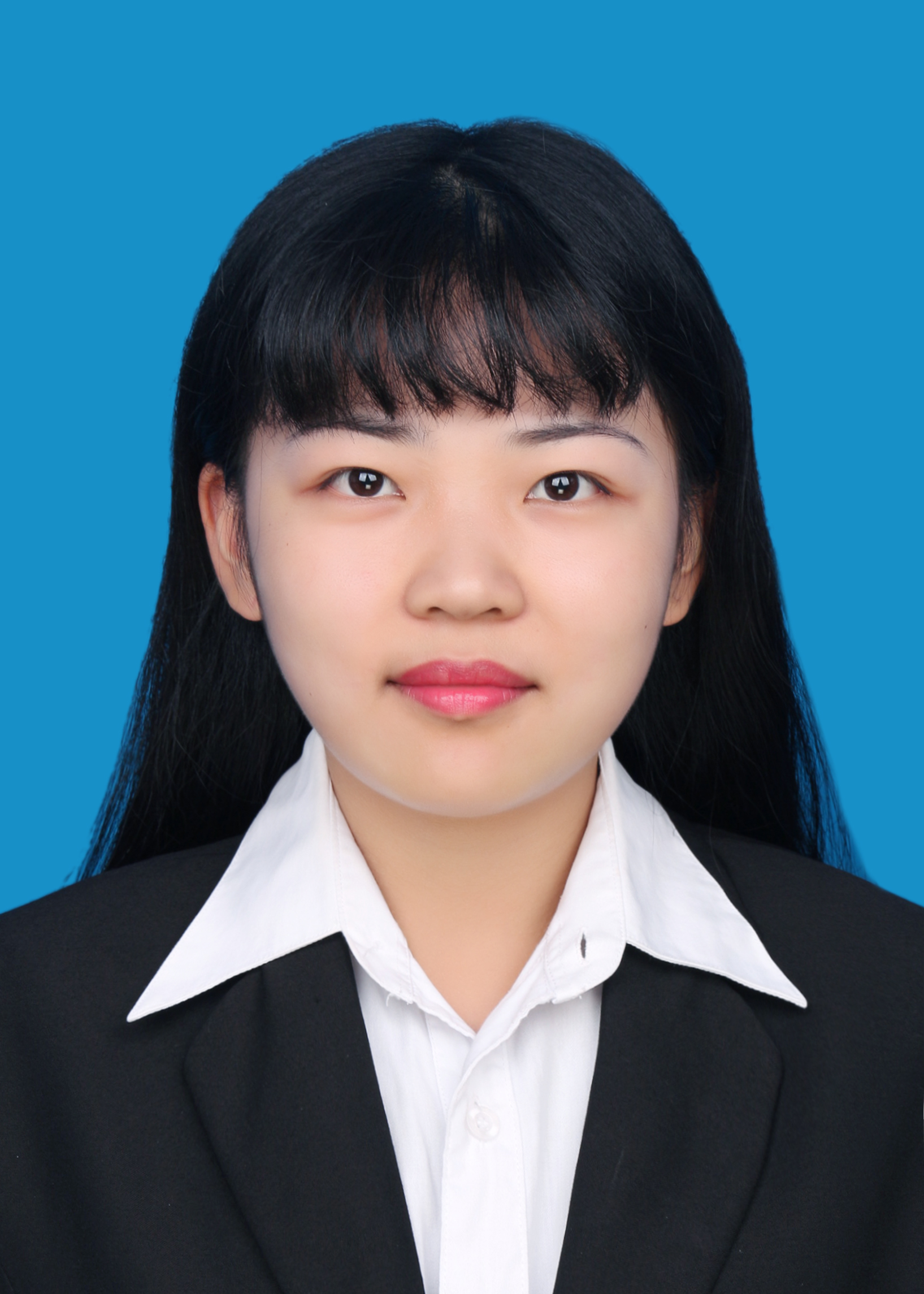}}] {Ting Liu} received the B.E. degree in electrical engineering and automation from Hunan Institute of Engineering , Xiangtan, China, in 2017, and the M.E. degree in control engineering from Xiangtan University (XTU), Xiangtan, China, in 2020. She is currently pursuing the Ph.D. degree with the College of Electronic Science in NUDT, Changsha, China. She research interests focus on signal processing, target detection and image processing.
\end{IEEEbiography}

\begin{IEEEbiography}[{\includegraphics[width=1in,height=1.25in,clip,keepaspectratio]{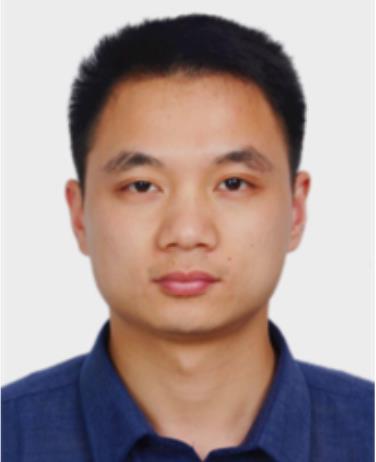}}]{Jungang Yang} received the B.E. and Ph.D. degrees from National University of Defense Technology (NUDT), in 2007 and 2013 respectively. He was a visiting Ph.D. student with the University of Edinburgh, Edinburgh from 2011 to 2012. He is currently an associate professor with the College of Electronic Science, NUDT. His research  interests include computational  imaging, image processing, compressive sensing and sparse representation. Dr. Yang received the New Scholar Award of Chinese Ministry of Education in 2012, the Youth Innovation Award and the Youth Outstanding Talent of NUDT in 2016.
\end{IEEEbiography}

\begin{IEEEbiography}[{\includegraphics[width=1in,height=1.25in,clip,keepaspectratio]{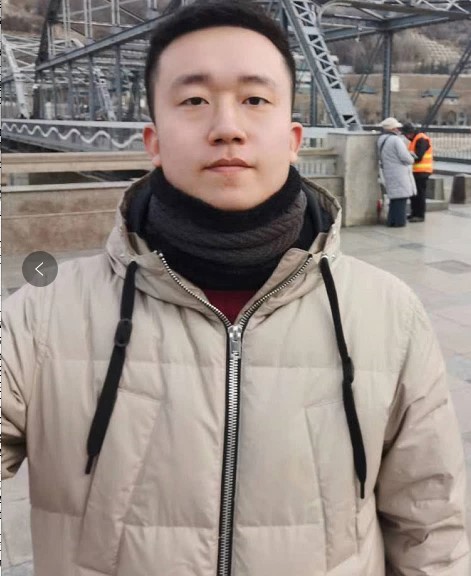}}]{Boyang Li}
received the B.E. degree in Mechanical Design manufacture and Automation from the Tianjin University, China, in 2017 and M.S. degree in biomedical engineering from National Innovation Institute of Defense Technology, Academy of Military Sciences, Beijing, China, in 2020. He is currently working toward the PhD degree in information and communication engineering from National University of Defense Technology (NUDT), Changsha, China. His research interests focus on VHR remote sensing image classification, infrared small target detection and deep learning.
\end{IEEEbiography}

\begin{IEEEbiography}[{\includegraphics[width=1in,height=1.25in,clip,keepaspectratio]{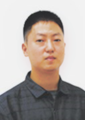}}]{Chao Xiao} received the BE degree in the communication engineering and the ME degree in information and communication engineering from the National University of Defense Technology (NUDT), Changsha, China in 2016 and 2018, respectively. He is currently working toward the Ph.D. degree with the College of Electronic Science in NUDT, Changsha, China. His research interests include deep learning, small object detection and object tracking.
\end{IEEEbiography}

\begin{IEEEbiography}[{\includegraphics[width=1in,height=1.25in,clip,keepaspectratio]{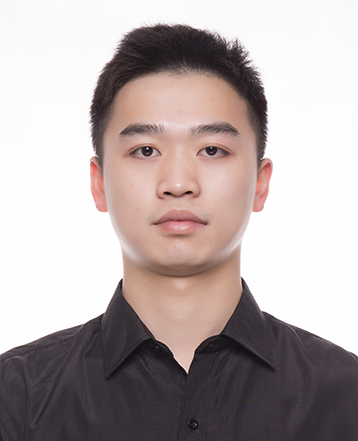}}]{Yang Sun} received the Ph.D. degree in information and communication engineering from the National University of Defense Technology (NUDT), Changsha, China, in 2020. He is currently a postdoctoral with the College of Electronic Science, NUDT. His research interests are image processing and infrared target detection..
\end{IEEEbiography}

\begin{IEEEbiography}[{\includegraphics[width=1in,height=1.25in,clip,keepaspectratio]{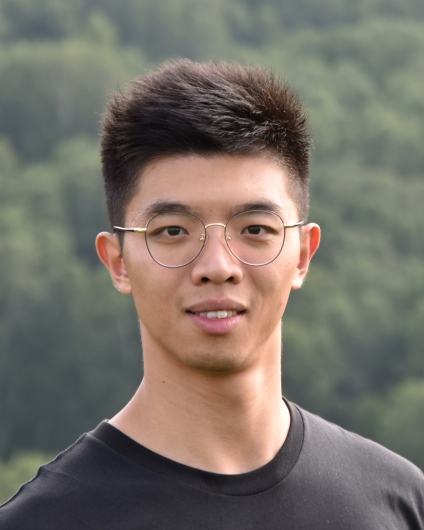}}]{Yingqian Wang} received the B.E. degree in electrical engineering from Shandong University (SDU), Jinan, China, in 2016, and the M.E. degree in information and communication engineering from National University of Defense Technology (NUDT), Changsha, China, in 2018. He is currently pursuing the Ph.D. degree with the College of Electronic Science and Technology, NUDT. His research interests focus on low-level vision, particularly on light field imaging and image super-resolution.
\end{IEEEbiography}

\begin{IEEEbiography}[{\includegraphics[width=1in,height=1.25in,clip,keepaspectratio]{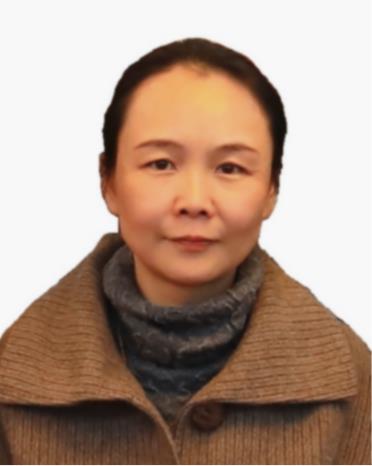}}]{Wei An} received the Ph.D. degree from the National University of Defense Technology (NUDT), Changsha, China, in 1999. She was a Senior Visiting Scholar with the University of Southampton, Southampton, U.K., in 2016. She is currently a Professor with the College of Electronic Science and Technology, NUDT. She has authored or co-authored over 100 journal and conference publications. Her current research interests include signal processing and image processing.
\end{IEEEbiography}

\end{document}